\documentclass[runningheads]{llncs}

\usepackage{eccv}

\usepackage{eccvabbrv}

\usepackage{graphicx}
\usepackage{booktabs}
\usepackage{algorithm}
\usepackage{algpseudocode}
\usepackage{amsmath}

\usepackage[accsupp]{axessibility}  

\usepackage{hyperref}

\usepackage{orcidlink}

\newcommand{\mytitle}{\textcolor{black}{RegionDrag}}
\newcommand{\jingyi}[1]{\textcolor{black}{#1}}
\newcommand{\xh}[1]{\textcolor{black}{#1}}

\makeatletter
\def\@fnsymbol#1{%
   \ifcase#1\or
   \TextOrMath \textdagger \dagger\or
   \TextOrMath \textdaggerdbl \ddagger \or
   \TextOrMath \textsection  \mathsection\or
   \TextOrMath \textparagraph \mathparagraph\or
   \TextOrMath \textbardbl \|\or
   \TextOrMath {\textdagger\textdagger}{\dagger\dagger}\or
   \TextOrMath {\textdaggerdbl\textdaggerdbl}{\ddagger\ddagger}\else
   \@ctrerr \fi
}
\makeatother

\begin{document}


\title{\mytitle: Fast Region-Based Image Editing with Diffusion Models}


\author{Jingyi Lu\inst{1}\orcidlink{0009-0005-1373-231X} \and
Xinghui Li\inst{2}\orcidlink{0000-0003-3797-5082} \and
Kai Han\inst{1}\thanks{Corresponding author.}\orcidlink{0000-0002-7995-9999} }

\authorrunning{Lu et al.}

\institute{$^{1}$The University of Hong Kong \quad $^{2}$University of Oxford \\
\email{lujingyi@connect.hku.hk, xinghui@robots.ox.ac.uk, kaihanx@hku.hk}
}

\maketitle

\vspace{0.8em}
\begin{figure*}[ht]
    \vspace{-12mm}
    \centering
    \includegraphics[width=.9\linewidth]{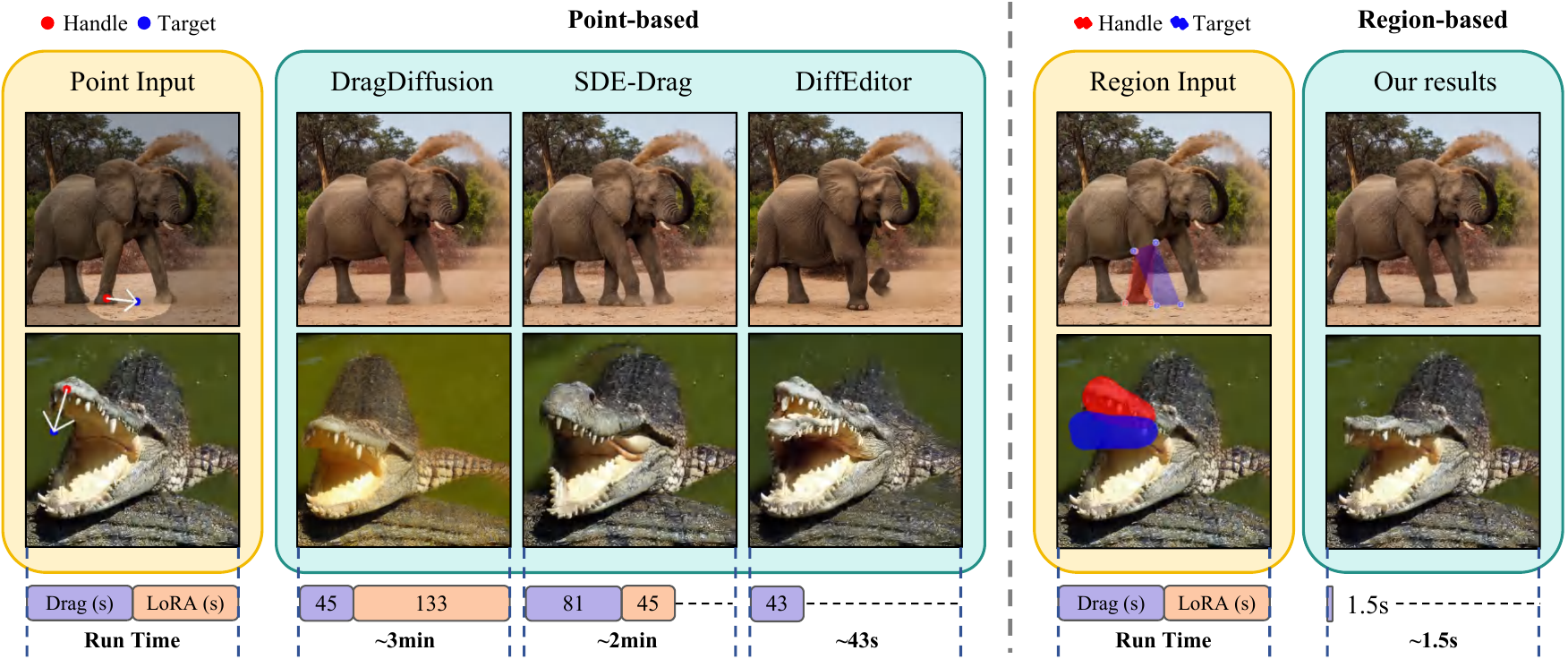}

    \caption{
    Comparison of editing results and latency between point-drag-based methods and our region-drag-based method. Our gradient-free, region-based framework reduces editing time from approximately one minute to about 1.5 seconds for 512\(\times\)512 resolution images, while producing results that better align with users' intentions.
    }
    \label{fig:teaser}

\end{figure*}

\vspace{-3em}

\begin{abstract}
Point-drag-based image editing methods, like DragDiffusion, have attracted significant attention. 
However, point-drag-based approach\-es suffer from computational overhead and misinterpretation of user intentions, due to the sparsity of point-based editing instructions.
In this paper, we propose a region-based copy-and-paste \jingyi{dragging method, \mytitle,} to overcome these limitations. 
\mytitle\ allows users to express their editing instructions in the form of handle and target regions, enabling more precise control and alleviating ambiguity. 
In addition, region-based operations complete editing in one iteration and are much faster than point-drag-based
methods.
We also incorporate the attention-swapping technique for enhanced stability during editing. 
To validate our approach, we extend existing point-drag-based datasets with region-based \jingyi{dragging} instructions.
Experimental results demonstrate that \mytitle\ outperforms existing point-drag-based approaches in terms of speed, accuracy, and alignment with user intentions.
Remarkably, \mytitle\ completes the edit on an image with a resolution of 512\(\times\)512 in less than 2 seconds, which is more than 100\(\times\) faster than DragDiffusion, while achieving better performance.
Project page: \url{https://visual-ai.github.io/regiondrag}.


\end{abstract}

\section{Introduction}
\label{sec:intro}

\jingyi{Stable Diffusion (SD) \cite{rombach2022high} is a widely adopted text-to-image generative model known for its efficacy in producing high-fidelity images.}
As it is trained with \jingyi{millions} of images, the vast amount of knowledge learned about images enables the model to edit existing images in a zero-shot manner.
\jingyi{
A recent line of work that has gained attention in the community is point-drag-based image editing using SD \cite{shi2023dragdiffusion, luo2024rotationdrag, nie2023blessing}.}
Point-drag-based methods first allow users to designate several points and drag them to desired positions. 
\xh{They} then invert the \xh{latent representation} of the \jingyi{input} image to a particular timestep in the diffusion process and edit the image by enforcing similarities between \xh{local latent representations} at the initial and final positions of the points. 
This is done either through optimization \cite{shi2023dragdiffusion, mou2023dragondiffusion, luo2024rotationdrag, mou2024diffeditor} or direct copy-and-paste \cite{nie2023blessing}.
The \xh{edited latent representation} is finally denoised and decoded to the edited image.

Although point-drag-based methods demonstrate encouraging results, they exhibit several limitations. First, as the editing solely relies on dragging sparse points, they have to interpolate dozens of intermediate points along the dragging directions and edit the image iteratively to avoid editing failures. This significantly slows down the speed of editing. Second, dragging points cannot always faithfully reflect the desired effect. This form of instruction is prone to being misinterpreted \jingyi{by the model}, so the editing results may not fully align with users' \jingyi{actual} intentions. For example, moving an object to the left and expanding the object to the left can equally be represented by dragging points to the left. 

Therefore, we propose an alternative form of \jingyi{dragging} to address the above issues. Instead of relying on points, we propose a region-based copy-and-paste \jingyi{dragging} method, \mytitle. \jingyi{Specifically, we first invert \xh{the latent representation} of the unedited image to a specific time step using DDPM inversion \cite{wu2022ddpminversion}},
\xh{an inverse process to the stochastic DDPM sampling~\cite{song2020denoising} which benefits the image editing~\cite{nie2023blessing}.}
\jingyi{Users then} draw a handle region and a target region \jingyi{(see \cref{fig:teaser})}, where \xh{the former} is the part that users \jingyi{intend} to \jingyi{drag} and \xh{the latter} illustrates desired positions that users \jingyi{intend} to achieve. \xh{After that, we establish} a dense mapping between the two regions using our proposed Region-to-Point Mapping algorithm, and the \xh{latent representation covered by} the handle region is mapped to the target region according to the dense mapping. Finally, the \xh{edited latent representation} goes through the DDPM sampling process, and the mapping operation is repeated at multiple time steps. In the meantime, we also employ the attention-swapping technique proposed in~\cite{cao2023masactrl} to stabilize the editing. To evaluate \mytitle, we expand two existing point-drag-based image editing datasets \cite{shi2023dragdiffusion, nie2023blessing} by adding equivalent region-based \jingyi{dragging} instructions to each image. \xh{Experiments demonstrate that }\mytitle\ is significantly faster than point-drag-based methods as \jingyi{it completes} editing in \jingyi{a single} iteration. In addition, the paste region offers clearer constraints than point inputs and reduces \jingyi{chances of misinterpretation}, leading to more \xh{faithful editing} results. 

Our contributions can be summarized as follows:
    (1) We introduce a region-based image \xh{editing} method to overcome the limitations of point-drag-based approaches, \jingyi{utilizing richer input context to better align the editing results with the users' intentions.}
    (2) By employing a gradient-free copy-paste operation, 
    our region-based image \xh{editing} becomes significantly faster than existing methods (see~\cref{fig:teaser}), \xh{completing the dragging in one single iteration.}
    (3) We extend two point-drag-based datasets with region-based \jingyi{dragging} instructions to validate \mytitle's effectiveness and \jingyi{benchmark region-based editing methods}.

\section{Related Work}
\noindent \textbf{Generative Models for Image Editing.} \
Early advancements in image generation have been driven by generative adversarial networks (GANs) \cite{goodfellow2014generative, karras2020analyzing, karras2021alias, kang2023scaling}. However, their practical application in real-world image editing is limited by the diversity of GAN training data and the effectiveness of GAN inversion techniques \cite{endo2022user, patashnik2021styleclip, xia2022gan, wang2022high, creswell2018inverting}. The emergence of text-to-image diffusion models leads to novel image editing techniques that utilize text prompts to modify high-level characteristics such as style, motion, or object categories \cite{cao2023masactrl, brooks2023instructpix2pix, epstein2024diffusion, kawar2023imagic}. 
Nevertheless, image editing methods based on user text prompts inherently struggle to manipulate images at the pixel level. 
Dragging methods \cite{endo2022user, pan2023drag, ling2023freedrag, shi2023dragdiffusion, luo2024rotationdrag, mou2023dragondiffusion, mou2024diffeditor} aim to address this limitation by controlling the overall posture and shape of objects through iterative movement and tracking of one or multiple key points. \mytitle\ introduces a region-based approach as a superior alternative, offering increased stability and efficiency for fine-grained image editing tasks.

\noindent \textbf{Image Editing by Dragging Points.} \
When employing drag-based methods for image editing, users can manipulate images by designating pairs of handle points and target points. The point-drag-based methods are expected to produce an image that meets two criteria: (1) features at the handle points are relocated to the target points, and (2) the original identity of the edited object is maintained. 
DragGAN \cite{pan2023drag} first enables editing on GAN-generated images involving multiple dragging point pairs. Specifically, DragGAN decomposes the editing process into several iterations, alternating between motion supervision and point tracking. In the motion supervision phase, DragGAN optimizes the StyleGAN's latent code \cite{karras2020analyzing, karras2021alias} using the distance loss between the initial and target positions of the handle points. Following each motion supervision iteration, DragGAN updates the positions of the handle points by employing a point-tracking technique. FreeDrag \cite{ling2023freedrag} switches the dragging from pixel space to feature space, enabling more stable and precise manipulation.
DragDiffusion \cite{shi2023dragdiffusion} extends the supervise-and-track framework of DragGAN \cite{pan2023drag} to diffusion models. It inverts the latent \jingyi{representation} of the unedited image to a partially noisy status at a selected time step, drags point features by optimizing the latent \jingyi{representation}, and denoises it. RotationDrag \cite{luo2024rotationdrag} further refines the DragDiffusion framework \cite{shi2023dragdiffusion} for cases of rotating objects. Another line of research, including DragonDiffusion \cite{mou2023dragondiffusion} and DiffEditor \cite{mou2024diffeditor}, draws inspiration from classifier-guided generation, extending feature dragging throughout the entire denoising process rather than confining it to a single timestep. A method similar to ours, named SDE-Drag \cite{nie2023blessing}, eliminates memory-intensive backpropagation found in \cite{mou2023dragondiffusion, mou2024diffeditor, shi2023dragdiffusion} by copying and pasting points within the diffusion latent space. Although SDE-Drag demonstrates promising results, it remains time-consuming due to its reliance on interpolating intermediate points along the dragging path to stabilize the editing process. In this paper, we introduce a simple yet effective region-based framework that utilizes the rich context inherent in region pairs to complete all edits in just one editing step.

\noindent \textbf{Appearance Consistency for Drag Editing.}
SD-based image editing shifts the image's latent distribution to incorporate user-specified structural, layout, and shape changes. However, this process can also introduce undesirable changes and artifacts. Mitigating these side effects remains challenging. \jingyi{Existing methods \cite{shi2023dragdiffusion, luo2024rotationdrag, nie2023blessing} often require training the SD model with LoRA (Low-Rank Adaptation) \cite{hu2021lora} for each unedited image to maintain style and appearance consistency. While LoRA is parameter-efficient through the use of low-rank matrix decomposition, it requires extensive preparation time and may overly constrain the image's appearance, limiting further edits.} Additionally, point-drag-based methods often necessitate masking the editing area to reduce input ambiguity and enhance image consistency.
In contrast, in this paper, we incorporate a training-free approach called mutual self-attention control \cite{cao2023masactrl}, which preserves the image's identity by leveraging keys and values in the self-attention blocks of the model. This approach is seamlessly integrated into the editing pipeline, eliminating the need for an additional module to maintain the image's identity. Furthermore, unlike point-drag-based methods that rely on a mask to confine the editing area, \mytitle\ eliminates this step to provide a more convenient editing experience.

\section{\mytitle}

\mytitle\ enables users to input handle and target region pairs, which are then used for editing through two primary steps: (1) copying \xh{latent representations covered by handle regions} and \xh{storing} self-attention features during inversion, and (2) pasting \xh{copied latent representations} to target positions and \xh{inserting stored} self-attention features during denoising. 

This section begins by reviewing the diffusion-based image editing pipeline and point-drag-based methods in \cref{sec:prelim}. We then discuss the limitations of point-drag-based methods and introduce \xh{how our region-based input addresses these limitations in \cref{sec:point&region}}. Finally, \cref{sec:drag} presents our editing pipeline used to process region-based inputs.

\subsection{Preliminary}
\label{sec:prelim}

\textbf{Diffusion-based image editing} involves two main stages: inversion and denoising. \jingyi{Unedited} images are first gradually inverted to a specific timestep in the diffusion process. The editing is then performed on the images at this timestep and the images are finally denoised back to their original image space. The transition between timesteps in the diffusion process is governed by a sampling scheduler. A commonly used one is the DDIM scheduler~\cite{song2020denoising}. When an image latent $z_s$ at timestep $s$ transitions to $z_t$ at timestep $t$, it goes through:

\begin{equation}
\begin{aligned}
z_t &= f_{s\rightarrow t}(z_s) \\
&= \sqrt{\alpha_t} \left( \frac{z_s - \sqrt{1-\alpha_s} \epsilon_\theta(z_s, s, C)}{\sqrt{\alpha_s}} \right) + \sqrt{1-\alpha_t - \sigma_s^2} \epsilon_\theta(z_s, s, C) + \sigma_s w_s, \\
\end{aligned}
\label{eq:1}
\end{equation}

\noindent where $\epsilon_\theta(z_s, s, C)$ is the noise predicted by the diffusion model, \(\alpha_t\) is a monotonically decreasing noise scheduling function dependent on \(t\), \(w_s\) is Gaussian noise, and \( \sigma_s = \eta\sqrt{(1 - \alpha_t)/(1 - \alpha_s)}\sqrt{1 - \alpha_s/\alpha_t}\) with $\eta=0$. Since $\eta$ is set to 0, the transition is a deterministic process and it is widely used in many methods \cite{shi2023dragdiffusion, mou2023dragondiffusion, rombach2022high, mou2024diffeditor}. If $\eta> 0$, the sampling becomes a stochastic process known as the DDPM sampling~\cite{ho2020denoising}. \jingyi{We choose DDPM over DDIM as \cite{nie2023blessing} shows that DDPM's randomness reduces divergence between unedited and edited image distributions.}

\noindent\textbf{Point-drag-based methods}, such as DragDiffusion and SDE-Drag \cite{shi2023dragdiffusion, nie2023blessing}, edit an image by relocating \xh{its latent representations} at user-designated \( n \) handle points \( h_{1:n} \) to \xh{corresponding} target positions \( t_{1:n} \). Due to the limited context available in sparse points, point-drag-based methods divide drag editing into \( K \) sub-steps. At sub-step \( k \), the algorithm drags \xh{latent representations} from handle points \( h_{k,1:n} \) to target positions \( t_{k,1:n} \), which are found by direct interpolation or an extra point matching step. DragDiffusion optimizes \( z_t^k \) by minimizing the \jingyi{\(\ell_1\)-distance} between UNet upblock features \( F(z^k_t) \) at handle \( h_{k,1:n} \) and target \( t_{k,1:n} \) points, \jingyi{respectively denoted as \( F(z^k_t)[h_{k,1:n}] \) and \( F(z^k_t)[t_{k,1:n}] \):}

\begin{equation}
\label{eq:4}
z_t^{k+1} = z_t^k - \eta \cdot \nabla_{z_t^k}\|F(z^k_t)[h_{k,1:n}] - F(z^k_t)[t_{k,1:n}]\|_1.
\end{equation}

\noindent DragonDiffusion \cite{mou2023dragondiffusion} and DiffEditor \cite{mou2024diffeditor} are DragDiffusion variants with \( K=1 \) and extend optimization across multiple diffusion timesteps. On the other hand, SDE-Drag copies the latent representations from handle points to target points using the copy-paste function \(\text{CP}\) (\cref{eq:5}), where \( z[h] \) and \( z[t] \) represent the latent code at positions \( h \) and \( t \). This is then followed by a denoising and inversion cycle (\cref{eq:6}).

\begin{equation}
\label{eq:5}
\text{CP}(z_1, z_2, h, t) = (z_2[t] \gets z_1[h]),
\end{equation}

\begin{equation}
\label{eq:6}
z_t^{k+1} = f_{0\rightarrow t}\left(f_{t\rightarrow 0}\left(\text{CP}(z_t^k, z_t^k, h_{k,1:n}, t_{k,1:n})\right)\right).
\end{equation}

\subsection{From Point-Based to Region-Based \jingyi{Dragging}}
\begin{figure*}[t]
    \centering
    \includegraphics[width=0.9\linewidth]{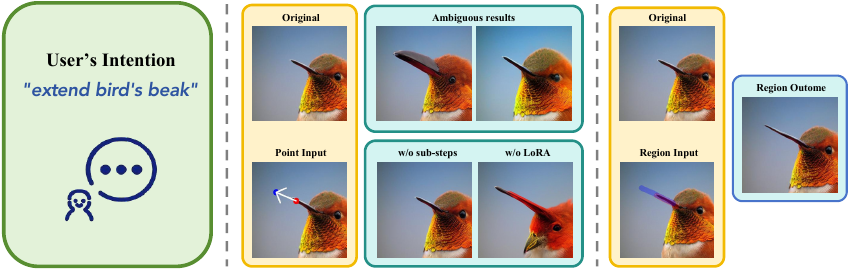}
    \caption{Overall comparison of point-based editing and region-based editing, exemplified by manipulating a bird's beak. The region-based approach is shown to provide a more user-friendly and less ambiguous editing experience.}
    \label{fig:edit}
\end{figure*}
\label{sec:point&region}

Although point-drag-based methods offer an intuitive means of user input, the limited information derived from sparse points poses challenges for models during the editing process.
Specifically, point instructions can result in two primary issues: input ambiguity and slow inference. \textbf{First, point instructions are inherently ambiguous.} One dragging action could correspond to multiple plausible editing effects. Consider a user attempting to elongate a bird's beak in an image, as depicted in \cref{fig:edit}. The user selects a point on the beak and drags it towards the upper-left corner. Point-drag-based methods, however, might misinterpret the user's goal as enlarging the beak or moving the entire bird to the left, rather than extending the beak as intended, leading to a misalignment between the user's intention and the model's output.

\textbf{Second, the complexity involved in point-drag-based editing requires considerable computational overhead.} Point-drag-based editing is challenging because the model must deduce changes across the entire image from the motion of a single or a few points. To carry out this complex drag operation while preserving the object's identity, point-drag-based methods heavily rely on two computationally intensive steps: training a unique LoRA~\cite{hu2021lora} for each image and breaking down the dragging process into a series of sub-steps. Particularly, LoRA helps the model maintain the original image's identity and step-by-step \jingyi{dragging} boosts the chance of achieving desired editing effects. Otherwise, the editing results may suffer from significant identity distortion or void editing, as shown in \cref{fig:edit}.
The root of this problem lies in that sparse points do not impose sufficient constraints on editing, so the model has to rely on LoRA to prevent distortion and iterative editing to provide some degree of additional supervision along the path of the dragging. Consequently, most point-drag-based methods require several minutes to edit one image, rendering them impractical for real-world applications.

\jingyi{The simplest solution to these problems is to encourage users to provide a sufficient number of points.} However, such an approach would result in users spending too much time on designating and dragging points.
Therefore, we design a form of editing that is not only user-friendly \jingyi{but also provides more informative context to the model, thus avoiding instruction ambiguity, slow inference, and exhaustive efforts from users.} Instead of relying on dragging points, we propose to use region-based operations, where users assign a handle region \(H\) to indicate the area they wish to \jingyi{drag} and a target region \(T\) to illustrate the desired \jingyi{position} they would like to achieve. We then establish a dense mapping between the two regions using our Region-to-Point Mapping algorithm and complete the editing by directly copying the \xh{latent representation covered by} the handle region to the target region in one inversion and denoising cycle. Despite the simplicity of this operation, it addresses ambiguity and overhead from two perspectives: (1) Region-based operation is more expressive and accurate than dragging points and would significantly alleviate the ambiguity. As demonstrated in \cref{fig:edit}, we express extending the bird's beak by simply drawing a longer beak, \jingyi{hence reducing ambiguity} presented in point-drag-based inputs. (2) Each region corresponds to \jingyi{a large number of} points after dense mapping, so it provides stronger constraints on editing results than sparse points. As a result, we do not have to interpolate intermediate points along the dragging path and crave extra supervision, and editing can be completed in one editing step. Moreover, the handle and target regions can vary in size and take arbitrary shapes allowing users to define them conveniently.

\subsection{Editing Pipeline}
\label{sec:drag}
\begin{figure*}[t]
    \centering
    \includegraphics[width=.9\linewidth]{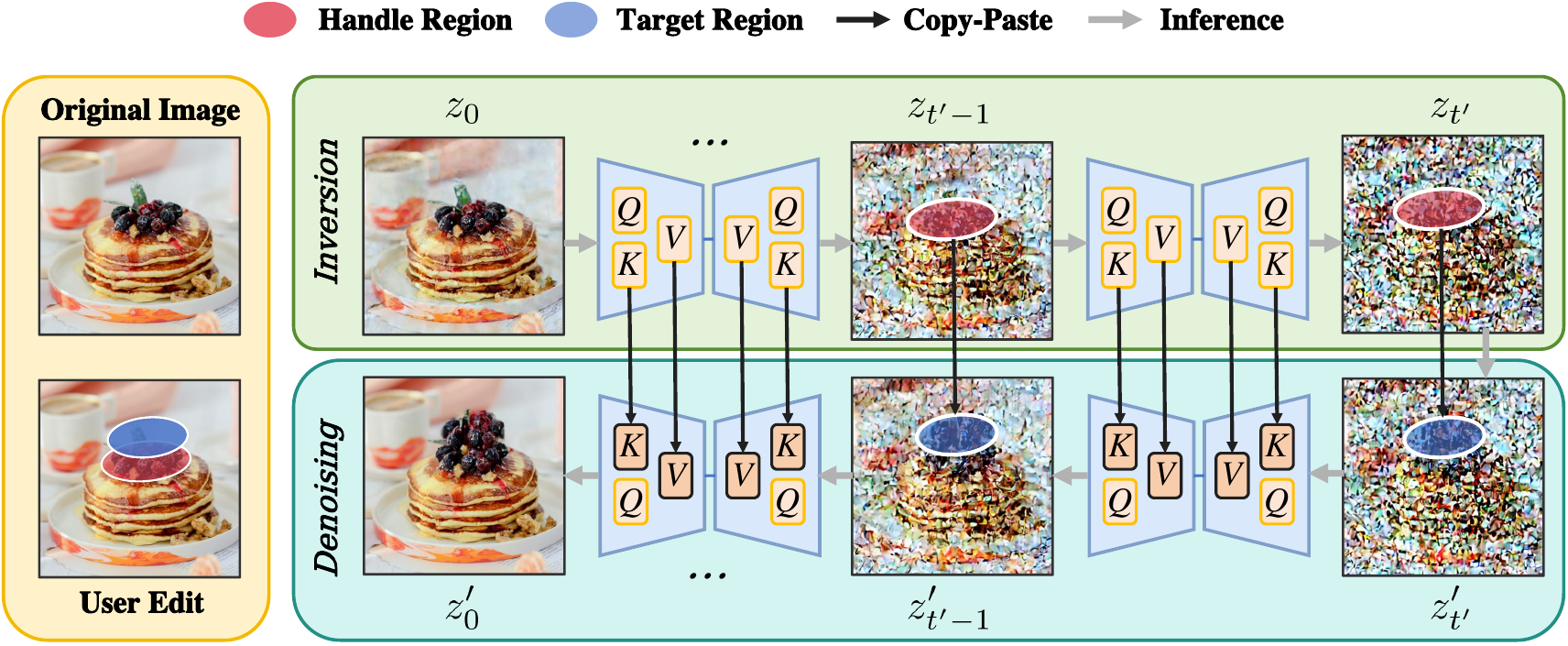}
    \caption{General pipeline of our method. Rich context provided by the region pairs enables users to complete accurate edits in one inversion and denoising cycle. The latent representation of the handle region is copied and pasted onto the target region throughout multiple timesteps for drag editing. Keys and values within the self-attention blocks are reused to ensure image consistency.}
    \label{fig:pipe}
\end{figure*}

We elaborate on our editing pipeline in this section. We first introduce the \jingyi{our region-based user input}, followed by our Region-to-Point Mapping algorithm, and finally the main working pipeline.

\noindent\textbf{\jingyi{User Input.}} \ The handle and target regions can be defined in two ways: \jingyi{(1) by entering vertices to form a polygon (e.g., a triangle or quadrilateral)}, or (2) by brushing out a flexible region using a brush tool. The choice of input form largely depends on the \jingyi{user's preferences.} Vertices are ideal for editing well-defined shapes, like moving a window on a building; a brush tool is better suited for irregular shapes, like a curved road or human hair.




\begin{algorithm}[t]
\caption{Region-to-Point Mapping}
\label{algo:1}
\begin{algorithmic}[1]
\Statex \textbf{Input:} Handle region $H$ with $N_H$ handle points $\{(x_i^H, y_i^H)\}_{i=1}^{N_H}$, target region $T$ with $N_T$ target points $\{(x_i^T, y_i^T)\}_{i=1}^{N_T}$.
\Statex \textbf{Output:} List of mapped point pairs $P$.
\Statex \textbf{Require:} Initialize empty list $P$.

\For{$(x, y) \in \{(x_i^T, y_i^T)\}_{i=1}^{N_T}$}
    \State \textbf{Horizontal scaling:}
    \State $x' \gets (x - \min(x_{1:N_T}^T)) / (\max(x_{1:N_T}^T) - \min(x_{1:N_T}^T))$
    \State $x' \gets \lfloor x' \cdot (\max(x_{1:N_H}^H) - \min(x_{1:N_H}^H)) + \min(x_{1:N_H}^H) \rfloor$
    \State \textbf{Column-by-column vertical scaling:}
    \State $y' \gets (y - \min(y_i^T \mid x_i^T = x)) / (\max(y_i^T \mid x_i^T = x) - \min(y_i^T \mid x_i^T = x))$
    \State $y' \gets \lfloor y' \cdot (\max(y_i^H \mid x_i^H = x') - \min(y_i^H \mid x_i^H = x')) + \min(y_i^H \mid x_i^H = x') \rfloor$
    \State Add point pair $((x', y'), (x, y))$ to $P$
\EndFor

\State \Return $P$
\end{algorithmic}
\end{algorithm}
\noindent\textbf{Region-to-Point Mapping.}  \ To preserve the original spatial information when copying and pasting latent representation, we need to establish a dense mapping between the handle and target region. If the regions are confined to triangular or quadrilateral shapes, we can compute a transformation matrix using affine or perspective mappings. However, finding a similar transformation for brush-out regions with arbitrary shapes is challenging. To address this issue, we propose an algorithm to numerically find the mapping between two regions. 

\jingyi{Let \(\{(x_i^H,\ y_i^H)\}_{i=1}^{N_H}\) and \(\{(x_j^T,\ y_j^T)\}_{j=1}^{N_T}\)} be the sets of pixels in handle and target regions respectively. We begin by linearly scaling the target region's width to match the handle region's width. For each pixel \jingyi{ \((x,\ y)\in \{(x_j^T,\ y_j^T)\}_{j=1}^{N_T}\)} in the target region, \(x\) \jingyi{is adjusted by first subtracting} the x-coordinate lower boundary \(\min(x_{1:N_T}^T)\) and is normalized using the target region's x-coordinate range \(\max(x_{1:N_T}^T)-\min(x_{1:N_T}^T)\). It is then scaled to the handle region's x-coordinate range \(\max(x_{1:N_H}^H)-\min(x_{1:N_H}^H)\), resulting in \(x'\). 

This step ensures that both regions have the same number of \jingyi{columns of pixels}. Next, we map each target region's vertical pixel column to its corresponding handle region's column. Each point's \(y\) in the target region is \jingyi{adjusted by subtracting} the lower boundary of its vertical column \(\min(y_i^T \mid x_i^T = x)\) and is normalized by its column's y-coordinate range \(\max(y_i^T \mid x_i^T = x)-\min(y_i^T \mid x_i^T = x)\), followed by being scaled to the corresponding handle region's column's y-coordinate range \(\max(y_i^H \mid x_i^H = x')-\min(y_i^H \mid x_i^H = x')\). The notation \(f(\cdot \mid \cdot)\) represents a function \(f\) (e.g., \(\min\) or \(\max\)) that takes a variable and a condition, and returns the function's value for the variable among points satisfying the condition. \jingyi{The whole process is summarised in \cref{algo:1}.}

\noindent\textbf{Main Pipeline. } As illustrated in \cref{fig:pipe}, \mytitle\ utilizes the image editing pipeline mentioned in \cref{sec:prelim}. Initially, \xh{the latent representation} of the image \( z_0 \) is inverted to \( z_{t'} \), where \( t' \) is a selected timestep prior to maximum timestep \( T \). Each intermediate step \( z_0, z_1, ..., z_{t'} \) is cached during inversion for future use. We then duplicate \( z_{t'} \) and denote the copy as \( z'_{t'} \). The handle regions of \( z'_{t'} \) are blended with Gaussian noise $\varepsilon$ according to the blending function $r_{\alpha}(z, H)$ defined in \cref{eq:8}:
\begin{equation}
\label{eq:8}
    r_{\alpha}(z, H) = (1 - H) \cdot z + H \cdot (\sqrt{1-\alpha^2} \cdot z + \alpha \cdot \varepsilon), \quad \varepsilon \sim \mathcal{N}(0, I),
\end{equation}
where \( H \) is a binary mask, with the handle region assigned a value of 1 and \( \alpha \) is a blending coefficient that ranges from 0 to 1, which governs the strength of the blending effect. The noise \( \varepsilon \) is drawn from a Gaussian distribution with mean 0 and identity covariance matrix, denoted by \( \mathcal{N}(0, I) \). If \(\alpha\) is set to a lower value, less noise is added, which preserves more of the original image's features and details in the handle regions. 
After blending with Gaussian noise, \( z'_{t'} \) goes through the denoising process using the DDPM sampler, and the \jingyi{dragging} is conducted in a copy-paste manner. At a denoising timestep \( t \), we extract \xh{latent representation} within the handle region of \( z_t \) and map it to the target region of \( z'_t \) according to the dense mapping computed by either geometric transformation or \cref{algo:1}. In the meantime, we employ mutual self-attention control \cite{cao2023masactrl} to help maintain the identity of images. In brief, when passing through the self-attention module of the UNet, the key and value (\( k'_t, v'_t \)) used when denoising \( z'_t \) are replaced with those (\( k_t, v_t \)) from \( z_t \). This allows the edited image to keep the layout and identity of the original image, thereby stabilizing the editing process. After \( z'_t \) is gradually denoised to \( z'_0 \) \jingyi{(see \cref{fig:pipe})}, it is decoded to the edited image \(x'_0\). 
Although \mytitle\ is introduced with a single handle-target region pair, it supports multiple pairs of input \(\{ (H_i, T_i) \}_{i=1}^{n}\), allowing users to specify several modifications in a single editing session. Dense mapping is individually constructed for each region pair but collectively used in latent copy-paste operations. The entire pipeline is \jingyi{summarised} in \cref{algo:2}. 
\begin{algorithm}[htb]
\caption{Editing Pipeline}
\label{algo:2}
\begin{algorithmic}[1]
\Statex \textbf{Input:} Image \(x_0\), handle and target region pairs \(\{ (H_i, T_i) \}_{i=1}^{n}\).
\Statex \textbf{Output:} Edited image \(x'_0\).
\Statex \textbf{Require:} Diffusion function \(f_{s\rightarrow t}\) \cref{eq:1}, VAE encoder \(E\), VAE decoder \(D\), region-to-point mapping function \(m\), copy-paste function \(CP\) \cref{eq:5}, handle resampling function \(r_{\alpha}\) \cref{eq:8}, copy-paste time interval \((t', t'')\). 

\State \textbf{Preparation:}
\State \(\{ (h_i, t_i) \}_{i=1}^{N}\) \(\leftarrow m(\{ (H_i, T_i) \}_{i=1}^{n})\)
\State \(z_0 \leftarrow E(x_0)\)
\State \textbf{Inversion stage:}
\For{\(t = 0\) to \(t' - 1\)}
    \State \(z_{t+1}, k_{t+1}, v_{t+1} \leftarrow f_{t \rightarrow t+1}(z_t)\) 
\EndFor
\State \(z'_{t'} \leftarrow r_{\alpha}(z_{t'}, \cup_{i=1}^n H_i)\) 
\State \textbf{Denoising stage:}
\For{\(t = t'\) down to \(1\)}
    \If{\(t \geq t''\)}
        \State \(z'_{t} \leftarrow \text{CP}(z_t, h_{1:N}, z'_{t}, t_{1:N})\) 
    \EndIf
    \State \(z'_{t-1} \leftarrow f_{t \rightarrow t-1}(z'_t, k'_t\leftarrow k_{t}, v'_t\leftarrow v_{t})\)
\EndFor
\State \(x'_0 \leftarrow D(z'_0)\)
\State \Return \(x'_0\)
\end{algorithmic}
\end{algorithm}

\section{Experimental Results}

\subsection{Datasets}
SDE-Drag \cite{nie2023blessing} and DragDiffusion \cite{shi2023dragdiffusion} have each \jingyi{introduced} a dataset named DragBench to evaluate the performance of point-drag-based methods. For clarity, we refer to \jingyi{the one by} SDE-Drag as DragBench-S, which contains 100 samples, and to \jingyi{the one by} DragDiffusion as DragBench-D, which includes 205 samples. Each sample comprises an image, a descriptive prompt, an optional mask delineating the editing region, and at least one pair of points that reflects the user's intention. Empirically, DragBench-S offers relatively more detailed prompts and includes a mask in 56 of its samples, while DragBench-D is characterized by more complex intentions and features carefully crafted masks in each of its samples.

To evaluate region-based editing, we introduce two new benchmarks: DragB\-ench-SR and DragBench-DR (R is short for ‘Region’), which are modified versions of DragBench-S and DragBench-D, respectively. These benchmarks are consistent with their point-drag-based counterparts in terms of images, prompts, and masks but differ by reflecting the user's intention through regions instead of points. DragBench-SR and DragBench-DR consist of 8 and 23 polygon-annotated samples, respectively, with the remainders of each dataset using brushes. For a handle-target region pair, the median number of equivalent point pairs is 267 in DragBench-SR and 201 in DragBench-DR. We visualize the distribution of the number of transformed point pairs in \cref{fig:log_num_pts}.

\jingyi{During the annotation process, we ensure that the region pairs are aligned with the point pairs to achieve the same editing effect. To simulate how real users would typically draw handle-target region pairs, we do not meticulously draw each region. According to our observations, each region-based sample is annotated in approximately 10 seconds.} This approach allows for a more realistic evaluation of the region-based editing methods while maintaining consistency with the point-based datasets.

\subsection{Evaluation Metrics}
\noindent\textbf{LPIPS:} We follow \cite{shi2023dragdiffusion} and use Learned Perceptual Image Patch Similarity (LPIPS) v0.1 \cite{zhang2018perceptual} to measure the identity similarity between the edited image and the original image. LPIPS computes the AlexNet \cite{krizhevsky2012imagenet} feature distances between the image pairs. A high LPIPS score indicates that unexpected identity changes or artifacts occur due to editing. A lower LPIPS score suggests that the object's identity has been well-preserved during editing; however, it does not necessarily imply a better drag edit, as two identical images would yield an LPIPS score of 0.

\begin{figure}[t]
  \centering
  \begin{minipage}{0.45\textwidth}
    \includegraphics[width=0.9\linewidth]{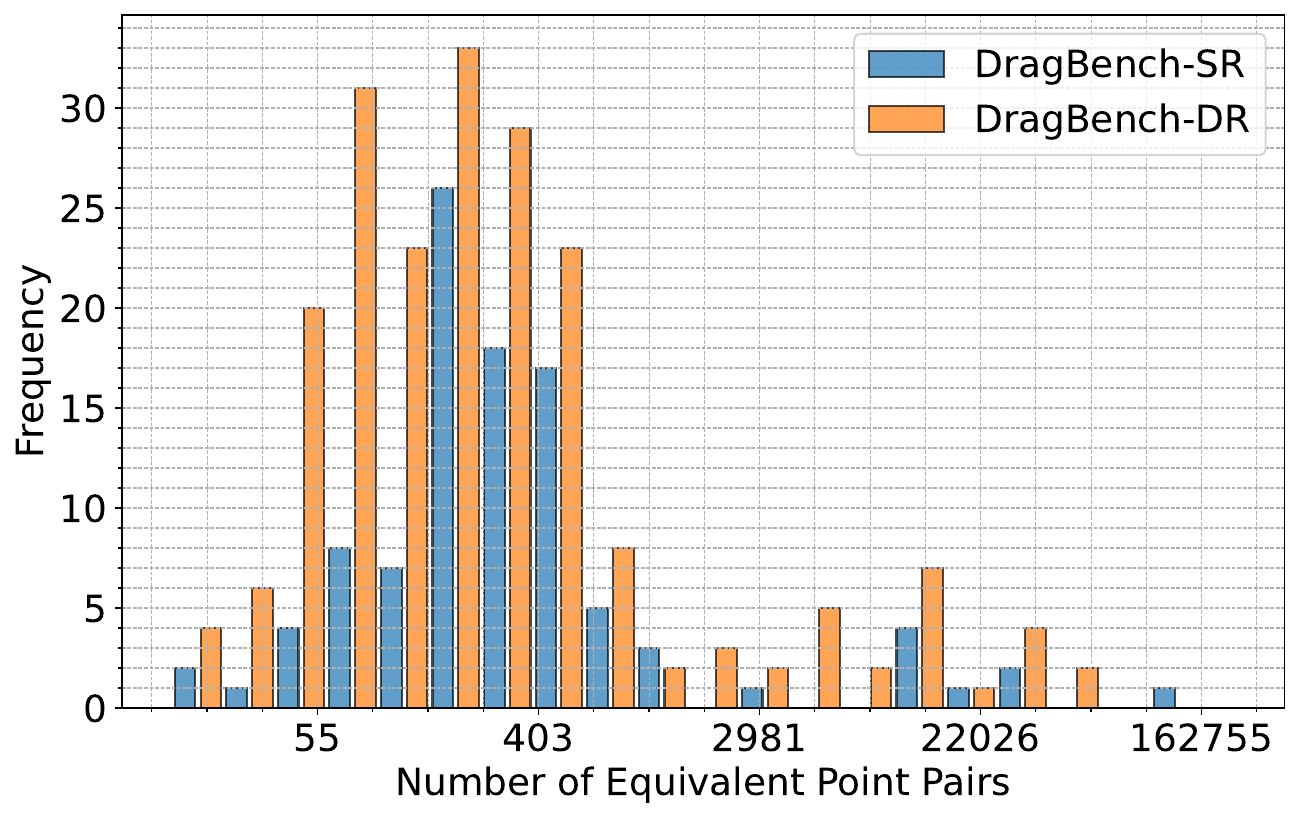}
    \caption{Log-transformed frequency distribution of equivalent point pair counts in DragBench-SR and DragBench-DR.}
    \label{fig:log_num_pts}
  \end{minipage}
  \hspace{0.01\textwidth}
  \begin{minipage}{0.46\textwidth}
    \vspace{-.3cm}
    \includegraphics[width=0.99\linewidth]{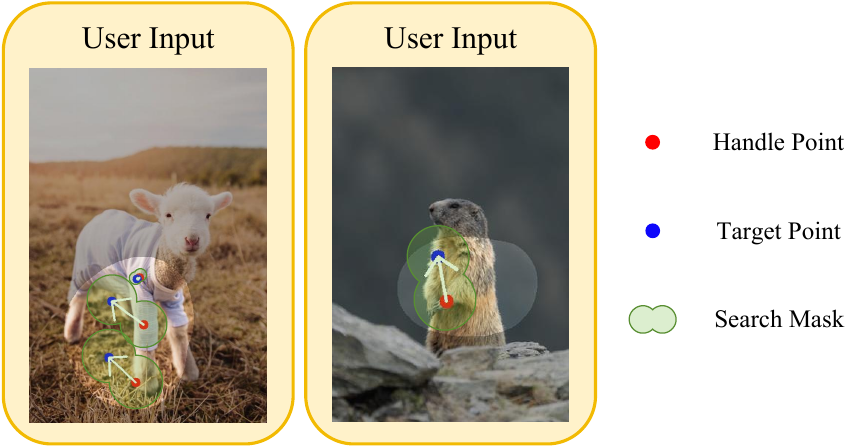}
    \vspace{-.2cm}
    \caption{Illustration of MD computation. Dragged handle points will be searched within the search mask.}
    \label{fig:mean_distance}
  \end{minipage}\hfill
\end{figure}

\noindent\textbf{Mean Distance (MD):} DragDiffusion \cite{shi2023dragdiffusion} introduces the \jingyi{MD} metric to assess how well an approach moves the handle points' content to the target points. 
To identify where the handle points have been moved to by the method, DragDiffusion employs DIFT \cite{tang2024emergent} to find the most similar points to the handle points \(h_{1:n}\) in the entire edited image and denotes them as \(h_{1:n}'\). It then uses the normalized Euclidean distance between DIFT-matched points and the true target positions \(t_{1:n}\) as the metric.
However, we believe that we only have to search the area around the handle points and their corresponding true target points for \(h_{1:n}'\) instead of the entire image, to avoid DIFT mistakenly identifying irrelevant points in the images as \(h'\) and excessively penalizing certain methods. 
Formally, given a dragging point pair \(\{h, t\}\), we define its searching mask as
\begin{equation}
    M(x) = 
      \begin{cases} 
       1 & \text{if } \min(d(x, h),d(x, t)) < \frac{\ d(h, t)}{\sqrt{2}} \\
       0 & \text{otherwise}
      \end{cases},
\end{equation}
where \(\text{d}(\cdot, \cdot)\) calculates the normalized Euclidean distance between two points \( (x_1, y_1) \) and \( (x_2, y_2) \) by
\begin{equation}
    d((x_1, y_1), (x_2, y_2)) = \sqrt{\left( \frac{x_2 - x_1}{W} \right)^2 + \left( \frac{y_2 - y_1}{H} \right)^2},
\end{equation}
where \(W\) and \(H\) are width and height of the image. We provide examples of searching masks in \cref{fig:mean_distance}. MD is then defined as the average of the normalized distances \(\bar{d}\) between the target points and the DIFT-matched points across all \( n \) points:
\begin{equation}
    \bar{d} = \frac{1}{n} \sum_{i=1}^{n} d(t_i, h_i').
\end{equation}

\subsection{Implementation Details}
\xh{Our method} is implemented in Python using the HuggingFace \cite{von-platen-etal-2022-diffusers} and PyTorch \cite{paszke2019pytorch} libraries. We employ Stable Diffusion v1-5 as our diffusion model \xh{with an image size of $512\times 512$}, being consistent with previous diffusion-based dragging methods. A DDPM sampler is utilized for both the inversion and denoising sampling processes, configured to use a total of 20 steps. The latent \jingyi{representation} is inverted to timestep \(t' = 500\) out of the total 1000 steps in SD1-5. Consequently, we perform 10 inversion steps and 10 denoising steps. Mutual self-attention control is enabled throughout all timesteps, and the latent copy-paste operation is terminated at \(t'' = 200\). The noise weight \(\alpha\) is set to 1. All \jingyi{experimental results are obtained} on an NVIDIA Tesla V100 GPU.

\subsection{Baselines}
We compare \mytitle\ with point-drag-based diffusion methods, including DragDiffusion \cite{shi2023dragdiffusion}, SDE-Drag \cite{nie2023blessing}, and DiffEditor \cite{mou2024diffeditor}. GAN-based methods are excluded due to their limitations in editing the diverse images in the DragBench-S and DragBench-D datasets, as they require domain-specific StyleGAN checkpoints. Diffusion-based methods, which outperform GAN-based methods in editing tasks, are better suited for our evaluation. \jingyi{Execution} times are averaged across both datasets, and all methods are tested on the same device using publicly released code.

\subsection{Quantitative Evaluation}
\label{Quantitative Evaluation}
\begin{figure*}[t]
\centering
\begin{minipage}[c]{.99\textwidth}
    \centering
    {\small
    \begin{tabular}{lcccccc}
    \toprule
    Method &  & \multicolumn{2}{c}{DragBench-S(R)} & \multicolumn{2}{c}{DragBench-D(R)} \\
           & Time (\(\downarrow\)) & MD (\(\downarrow\)) & LPIPS (\(\downarrow\)) & MD (\(\downarrow\)) & LPIPS (\(\downarrow\))  \\
    \midrule
    SDE-Drag\cite{nie2023blessing}             & 126.1 & 7.5 & 12.4 & 8.1 & 14.9 \\
    DragDiffusion\cite{shi2023dragdiffusion}  & 177.7 & 7.0 & 18.0 & 6.7 & 11.5 \\
    DiffEditor\cite{mou2024diffeditor}        & 43.1 & 23.6 & 17.6 & 22.1 & 10.9 \\
    Ours                                      & \textbf{1.5} & \textbf{6.4} & \textbf{9.9} & \textbf{6.6} & \textbf{9.2} \\
    \bottomrule
    \end{tabular}
    }
     \captionof{table}{Comparisons of our method with baseline methods using MD(\(\times\)100) and LPIPS(\(\times\)100) metrics on DragBench-S(R) and DragBench-D(R) datasets. The time is measured in seconds and averaged across both datasets. \xh{The image size is 512$\times$512.}}
    \label{tab:results}
\end{minipage}
\end{figure*}

To quantitatively evaluate the editing performance of the methods, we employ LPIPS and Mean Distance as metrics, multiplying both by 100 for illustration purposes. As demonstrated in Table \ref{tab:results}, \mytitle\ significantly outperforms those computationally expensive point-based methods on both DragBench-S(R) and DragBench-D(R) datasets. These results highlight \mytitle's superior performance in maintaining image consistency while achieving competitive editing results across different datasets. In addition to its effectiveness, \mytitle\ also excels in terms of efficiency. \mytitle\ achieves fast inference speed, requiring approximately 1.5 seconds to  edit \xh{a $512\times 512$ image}, which is 20 times faster than the second-fastest method and 100 times faster than DragDiffusion \cite{shi2023dragdiffusion}. The inference time for \mytitle\ is comparable to generating an image with 20 steps using SD1-5, given that the copy-paste operations introduce negligible computational overhead.

\subsection{Qualitative Results}
\begin{figure*}[t]
    \centering
    \includegraphics[width=1.\linewidth]{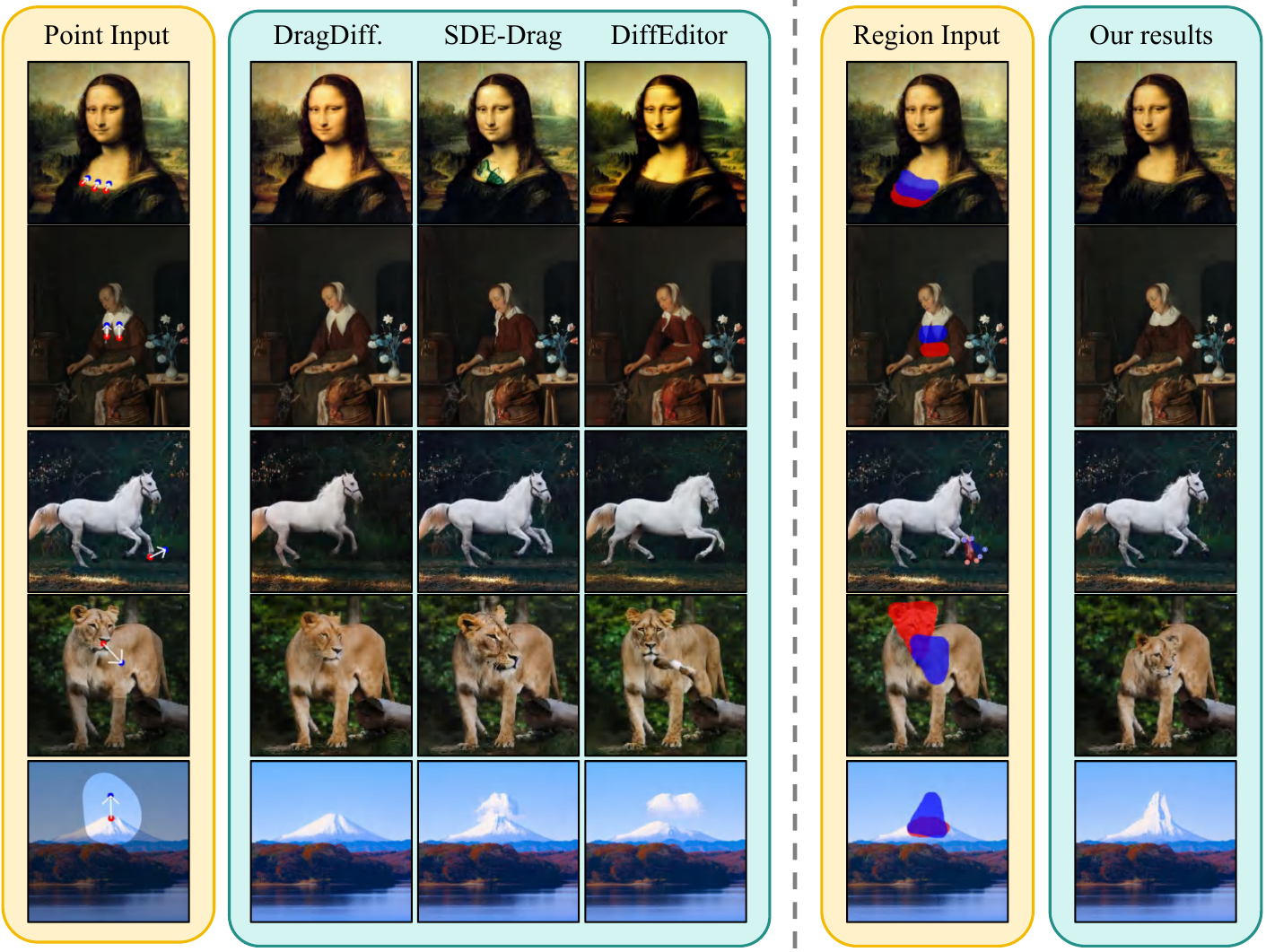}
    \caption{Qualitative comparisons with baseline methods. Handle regions and target regions are respectively denoted by red and blue masks.}
    \label{fig:qualitative_results} 
\end{figure*}

\Cref{fig:qualitative_results} compares examples of point-drag-based and region-based editing inputs and their corresponding results, demonstrating the effectiveness of \mytitle. Our region-based method utilizes the comprehensive context provided by annotated regions to target desired modifications while preserving the overall coherence of the image, outperforming point-drag-based editing methods.

\subsection{Ablation Study} 
\begin{figure*}[t]
    \centering
    \begin{minipage}[t]{.38\textwidth}
        \vspace{0pt} 
        \includegraphics[width=\linewidth]{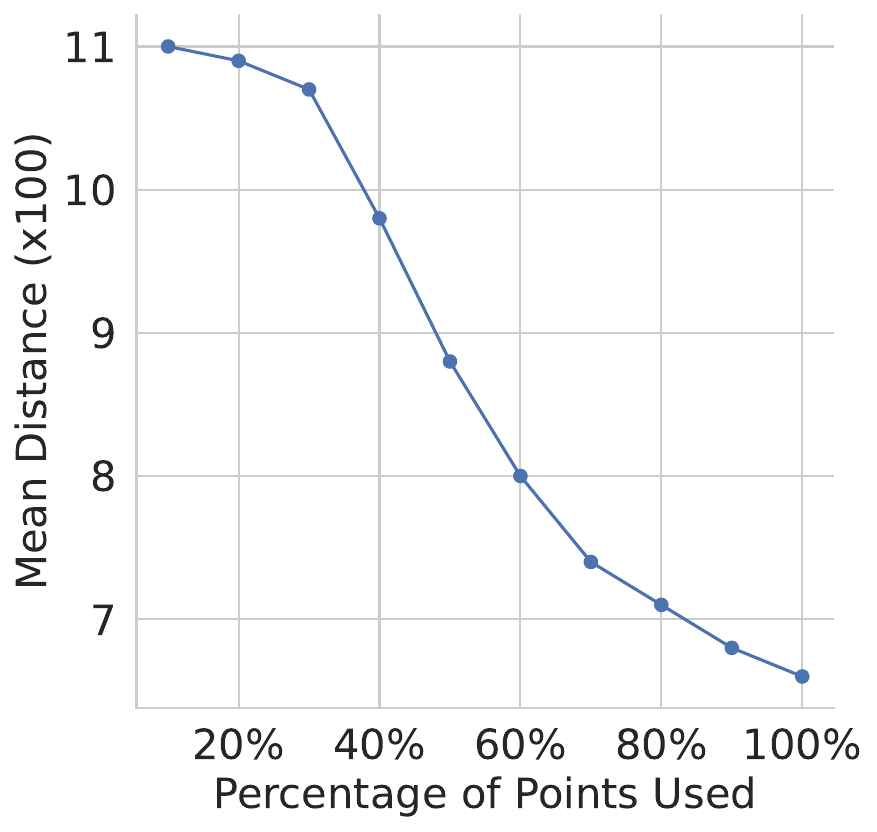}
        \captionof{figure}{Ablation study on the impact of the percentage of inputted transformed points.}
        \label{fig:point_ablation}
    \end{minipage}
    \hfill
    \begin{minipage}[t]{.57\textwidth}
        \vspace{0pt} 
        \centering
        \includegraphics[width=\linewidth]{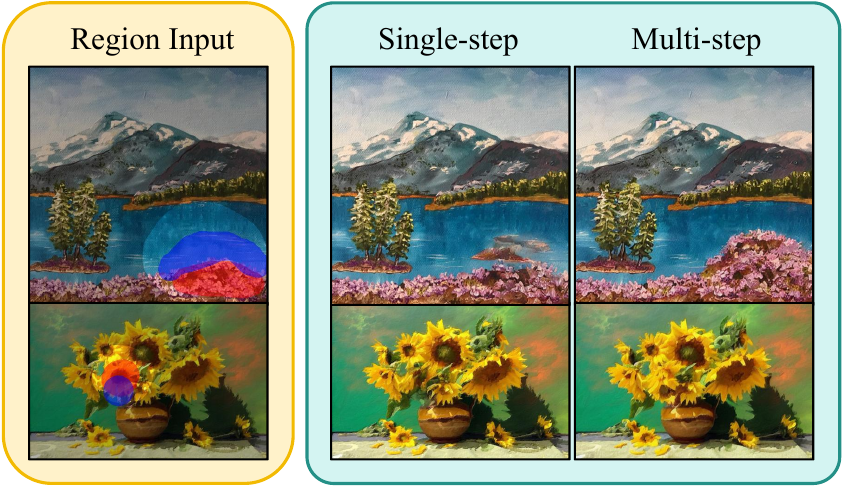}
        \vspace{.5mm}
        \caption{Qualitative examples illustrating the impact of multi-step copy-paste.}
        \label{fig:multi}
    \end{minipage}
\end{figure*}

We argue that the sparsity of point inputs leads to inferior editing results. To quantitatively demonstrate this, we conducted tests on the DragBench-DR dataset by randomly selecting subsets of the equivalent transformed points within each sample and performing inference using these subsets. We gradually reduced the percentage of selected points to observe the impact on the MD metric. As illustrated in \cref{fig:point_ablation}, the results exhibit a clear upward trend in MD as the percentage of utilized points decreases. This \jingyi{suggests} that sparse point inputs provide weaker constraints on the output compared to region-based inputs, leading to unsatisfactory editing results. It confirms the benefits of employing region-based inputs in \mytitle. 

We copy and paste image's latent representation over a time interval during denoising. To validate this design, we compare it to copy-paste only at the initial denoising timestep. \Cref{fig:multi} shows that editing just at the initial step can yield unpredictable results because the edits may be \jingyi{lost} in subsequent denoising phases. Multi-step copy-paste solves this by providing extra guidance at \jingyi{smaller} timesteps while preserving image fidelity.

\section{Conclusion}
In this paper, we have introduced an efficient and effective region-based editing framework, \mytitle, for high-fidelity image editing. 
Unlike existing approaches that utilize point-drag-based editing, \mytitle\ reconsiders the editing problem from the region perspective. \mytitle\ allows for editing in a single step through copying and pasting the latent representation and self-attention features of the image, which not only provides excellent efficiency but also achieves superior editing performance. 
Furthermore, we have introduced two new benchmarks, DragBench-SR and DragBench-DR based on existing datasets, for the evaluation of region-based editing. 
Experimental results have consistently demonstrated the superior efficiency and editing performance of our method.

\section*{Acknowledgments}
Kai Han and Jingyi Lu would like to acknowledge the support of the Hong Kong Research Grants Council - General Research Fund (Grant No.: $17211024$).

\clearpage
\bibliographystyle{splncs04}
\bibliography{main}

\begin{thebibliography}{10}
\providecommand{\url}[1]{\texttt{#1}}
\providecommand{\urlprefix}{URL }
\providecommand{\doi}[1]{https://doi.org/#1}

\bibitem{brooks2023instructpix2pix}
Brooks, T., Holynski, A., Efros, A.A.: Instructpix2pix: Learning to follow image editing instructions. In: CVPR (2023)

\bibitem{cao2023masactrl}
Cao, M., Wang, X., Qi, Z., Shan, Y., Qie, X., Zheng, Y.: Masactrl: Tuning-free mutual self-attention control for consistent image synthesis and editing. arXiv preprint arXiv:2304.08465  (2023)

\bibitem{creswell2018inverting}
Creswell, A., Bharath, A.A.: Inverting the generator of a generative adversarial network. IEEE TNNLS  (2018)

\bibitem{endo2022user}
Endo, Y.: User-controllable latent transformer for stylegan image layout editing. In: CGF (2022)

\bibitem{epstein2024diffusion}
Epstein, D., Jabri, A., Poole, B., Efros, A., Holynski, A.: Diffusion self-guidance for controllable image generation. In: NeurIPS (2024)

\bibitem{goodfellow2014generative}
Goodfellow, I., Pouget-Abadie, J., Mirza, M., Xu, B., Warde-Farley, D., Ozair, S., Courville, A., Bengio, Y.: Generative adversarial nets. In: NeurIPS (2014)

\bibitem{ho2020denoising}
Ho, J., Jain, A., Abbeel, P.: Denoising diffusion probabilistic models. In: NeurIPS (2020)

\bibitem{hu2021lora}
Hu, E.J., Shen, Y., Wallis, P., Allen-Zhu, Z., Li, Y., Wang, S., Wang, L., Chen, W.: Lora: Low-rank adaptation of large language models. arXiv preprint arXiv:2106.09685  (2021)

\bibitem{kang2023scaling}
Kang, M., Zhu, J.Y., Zhang, R., Park, J., Shechtman, E., Paris, S., Park, T.: Scaling up gans for text-to-image synthesis. In: CVPR (2023)

\bibitem{karras2021alias}
Karras, T., Aittala, M., Laine, S., H{\"a}rk{\"o}nen, E., Hellsten, J., Lehtinen, J., Aila, T.: Alias-free generative adversarial networks. In: NeurIPS (2021)

\bibitem{karras2020analyzing}
Karras, T., Laine, S., Aittala, M., Hellsten, J., Lehtinen, J., Aila, T.: Analyzing and improving the image quality of stylegan. In: CVPR (2020)

\bibitem{kawar2023imagic}
Kawar, B., Zada, S., Lang, O., Tov, O., Chang, H., Dekel, T., Mosseri, I., Irani, M.: Imagic: Text-based real image editing with diffusion models. In: CVPR (2023)

\bibitem{krizhevsky2012imagenet}
Krizhevsky, A., Sutskever, I., Hinton, G.E.: Imagenet classification with deep convolutional neural networks. NeurIPS  (2012)

\bibitem{ling2023freedrag}
Ling, P., Chen, L., Zhang, P., Chen, H., Jin, Y.: Freedrag: Point tracking is not you need for interactive point-based image editing. arXiv preprint arXiv:2307.04684  (2023)

\bibitem{luo2024rotationdrag}
Luo, M., Cheng, W., Yang, J.: Rotationdrag: Point-based image editing with rotated diffusion features. arXiv preprint arXiv:2401.06442  (2024)

\bibitem{mou2023dragondiffusion}
Mou, C., Wang, X., Song, J., Shan, Y., Zhang, J.: Dragondiffusion: Enabling drag-style manipulation on diffusion models. arXiv preprint arXiv:2307.02421  (2023)

\bibitem{mou2024diffeditor}
Mou, C., Wang, X., Song, J., Shan, Y., Zhang, J.: Diffeditor: Boosting accuracy and flexibility on diffusion-based image editing. arXiv preprint arXiv:2402.02583  (2024)

\bibitem{nie2023blessing}
Nie, S., Guo, H.A., Lu, C., Zhou, Y., Zheng, C., Li, C.: The blessing of randomness: Sde beats ode in general diffusion-based image editing. arXiv preprint arXiv:2311.01410  (2023)

\bibitem{pan2023drag}
Pan, X., Tewari, A., Leimk{\"u}hler, T., Liu, L., Meka, A., Theobalt, C.: Drag your gan: Interactive point-based manipulation on the generative image manifold. In: ACM SIGGRAPH (2023)

\bibitem{paszke2019pytorch}
Paszke, A., Gross, S., Massa, F., Lerer, A., Bradbury, J., Chanan, G., Killeen, T., Lin, Z., Gimelshein, N., Antiga, L., et~al.: Pytorch: An imperative style, high-performance deep learning library. In: NeurIPS (2019)

\bibitem{patashnik2021styleclip}
Patashnik, O., Wu, Z., Shechtman, E., Cohen-Or, D., Lischinski, D.: Styleclip: Text-driven manipulation of stylegan imagery. In: ICCV (2021)

\bibitem{von-platen-etal-2022-diffusers}
von Platen, P., Patil, S., Lozhkov, A., Cuenca, P., Lambert, N., Rasul, K., Davaadorj, M., Wolf, T.: Diffusers: State-of-the-art diffusion models. \url{https://github.com/huggingface/diffusers} (2022)

\bibitem{rombach2022high}
Rombach, R., Blattmann, A., Lorenz, D., Esser, P., Ommer, B.: High-resolution image synthesis with latent diffusion models. In: CVPR (2022)

\bibitem{shi2023dragdiffusion}
Shi, Y., Xue, C., Pan, J., Zhang, W., Tan, V.Y., Bai, S.: Dragdiffusion: Harnessing diffusion models for interactive point-based image editing. arXiv preprint arXiv:2306.14435  (2023)

\bibitem{song2020denoising}
Song, J., Meng, C., Ermon, S.: Denoising diffusion implicit models. arXiv preprint arXiv:2010.02502  (2020)

\bibitem{tang2024emergent}
Tang, L., Jia, M., Wang, Q., Phoo, C.P., Hariharan, B.: Emergent correspondence from image diffusion. In: NeurIPS (2024)

\bibitem{wang2022high}
Wang, T., Zhang, Y., Fan, Y., Wang, J., Chen, Q.: High-fidelity gan inversion for image attribute editing. In: CVPR (2022)

\bibitem{wu2022ddpminversion}
Wu, C.H., De~la Torre, F.: Unifying diffusion models' latent space, with applications to cyclediffusion and guidance. arXiv preprint arXiv:2210.05559  (2022)

\bibitem{xia2022gan}
Xia, W., Zhang, Y., Yang, Y., Xue, J.H., Zhou, B., Yang, M.H.: Gan inversion: A survey. IEEE TPAMI  (2022)

\bibitem{zhang2018perceptual}
Zhang, R., Isola, P., Efros, A.A., Shechtman, E., Wang, O.: The unreasonable effectiveness of deep features as a perceptual metric. In: CVPR (2018)

\end{thebibliography}
\newpage

\title{RegionDrag: Fast Region-Based Image Editing with Diffusion Models\\
\jingyi{\textit{--Supplementary Material--}}}
\titlerunning{Supplementary Material of RegionDrag}


\author{Jingyi Lu\inst{1}\orcidlink{0009-0005-1373-231X} \and
Xinghui Li\inst{2}\orcidlink{0000-0003-3797-5082} \and
Kai Han\inst{1}\thanks{Corresponding author.}\orcidlink{0000-0002-7995-9999} }

\authorrunning{Lu et al.}

\institute{$^{1}$The University of Hong Kong \quad $^{2}$University of Oxford \\
\email{lujingyi@connect.hku.hk, xinghui@robots.ox.ac.uk, kaihanx@hku.hk}
}

\maketitle

\section{Effectiveness of Region-Based Inputs}
\label{sec:supp_point_ablation}
\begin{figure*}[ht]

    \centering
    \includegraphics[width=.8\linewidth]{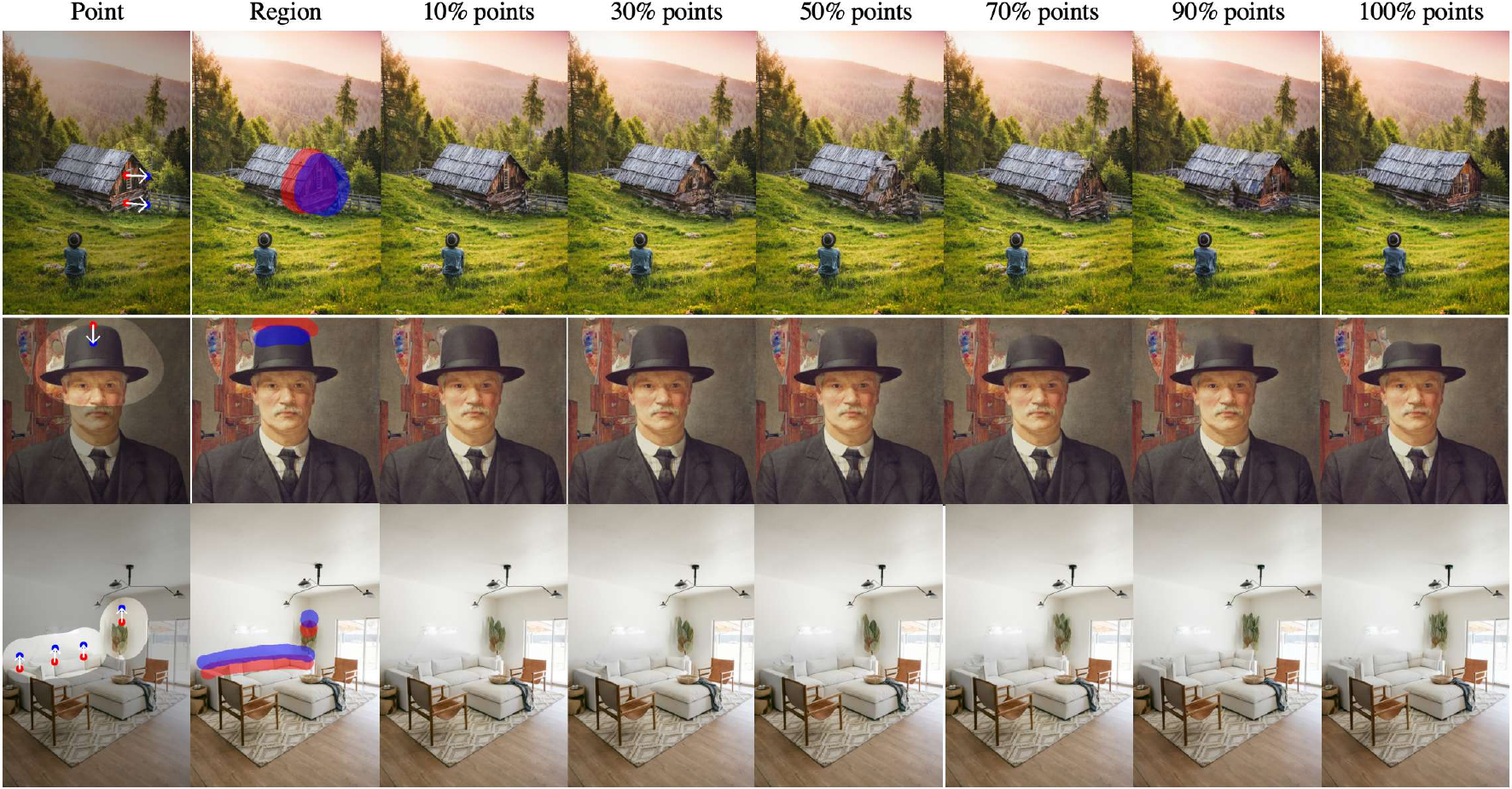}
    \caption{Improved editing quality with a higher percentage of transformed points.}
    \label{fig:points_supp}
\end{figure*}

In the main text, we quantitatively demonstrate the effectiveness of using more point pairs obtained from the region pairs. To complement our quantitative findings, we present qualitative examples in \cref{fig:points_supp}. The results suggest that increasing the percentage of transformed points used enhances the quality and stability of the editing outcomes.

\vspace{-1cm}
\section{Discussion of Noise Weight}
\label{sec:supp_noise}
\begin{figure*}[ht]
    \centering
    \includegraphics[width=.8\linewidth]{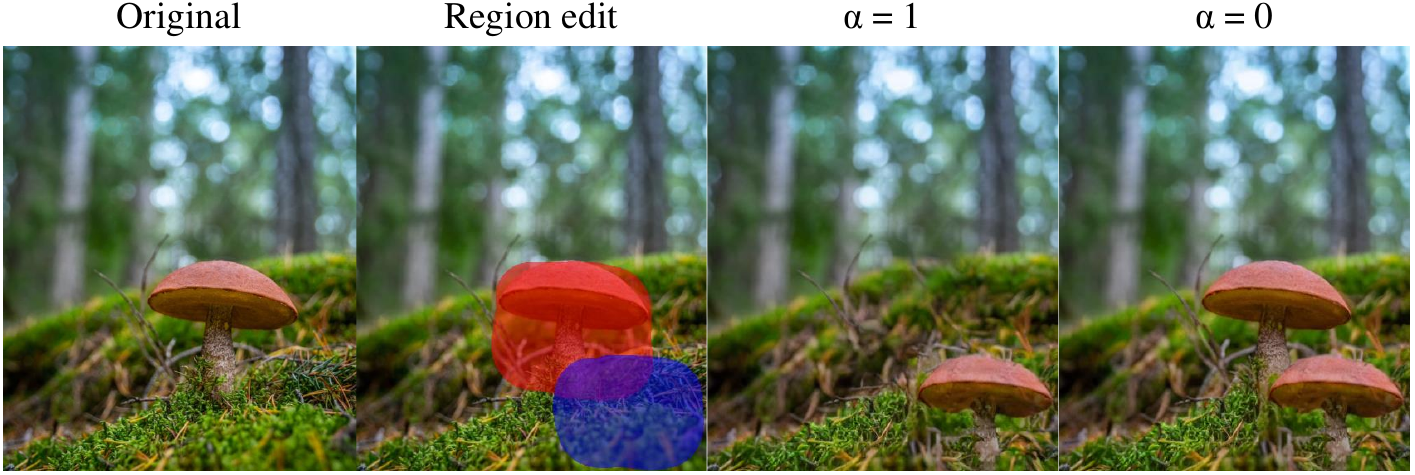}
    \caption{Results of applying different noise weight \(\alpha\).}
    \label{fig:alpha_supp}
\end{figure*}

In the initial denoising step, the latent representations of the handle regions are blended with random noise weighted by \(\alpha\), where \(\alpha\) ranges from 0 to 1 (Eq. (5) in main text). As illustrated in \cref{fig:alpha_supp}, a higher \(\alpha\) value retains less of the original content. Nonetheless, the object in the handle region is not guaranteed to be removed even when \(\alpha=1\), as the outcome is influenced by other factors including the denoising process, the attention swapping mechanism, and the text prompt.

\section{More Qualitative Results}
\begin{figure*}[ht]

    \centering
        \begin{subfigure}{.16\linewidth}
        \caption*{Original}
    \end{subfigure}%
    \begin{subfigure}{.16\linewidth}
        \caption*{Region input}
    \end{subfigure}%
    \begin{subfigure}{.16\linewidth}
        \caption*{Our results}
    \end{subfigure}%

    \begin{subfigure}{.16\linewidth}
        \centering
        \includegraphics[width=\linewidth]{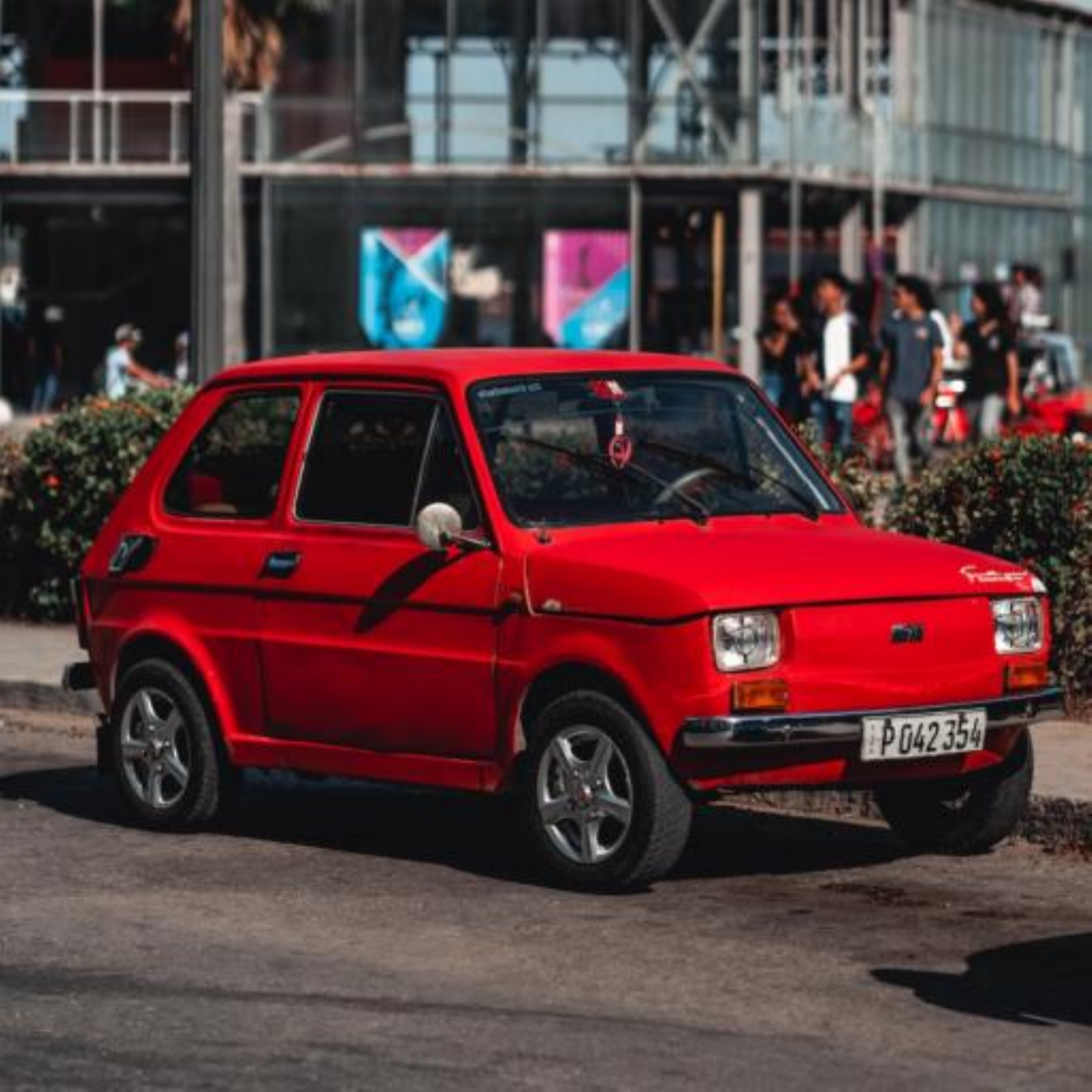}
    \end{subfigure}%
    \begin{subfigure}{.16\linewidth}
        \centering
        \includegraphics[width=\linewidth]{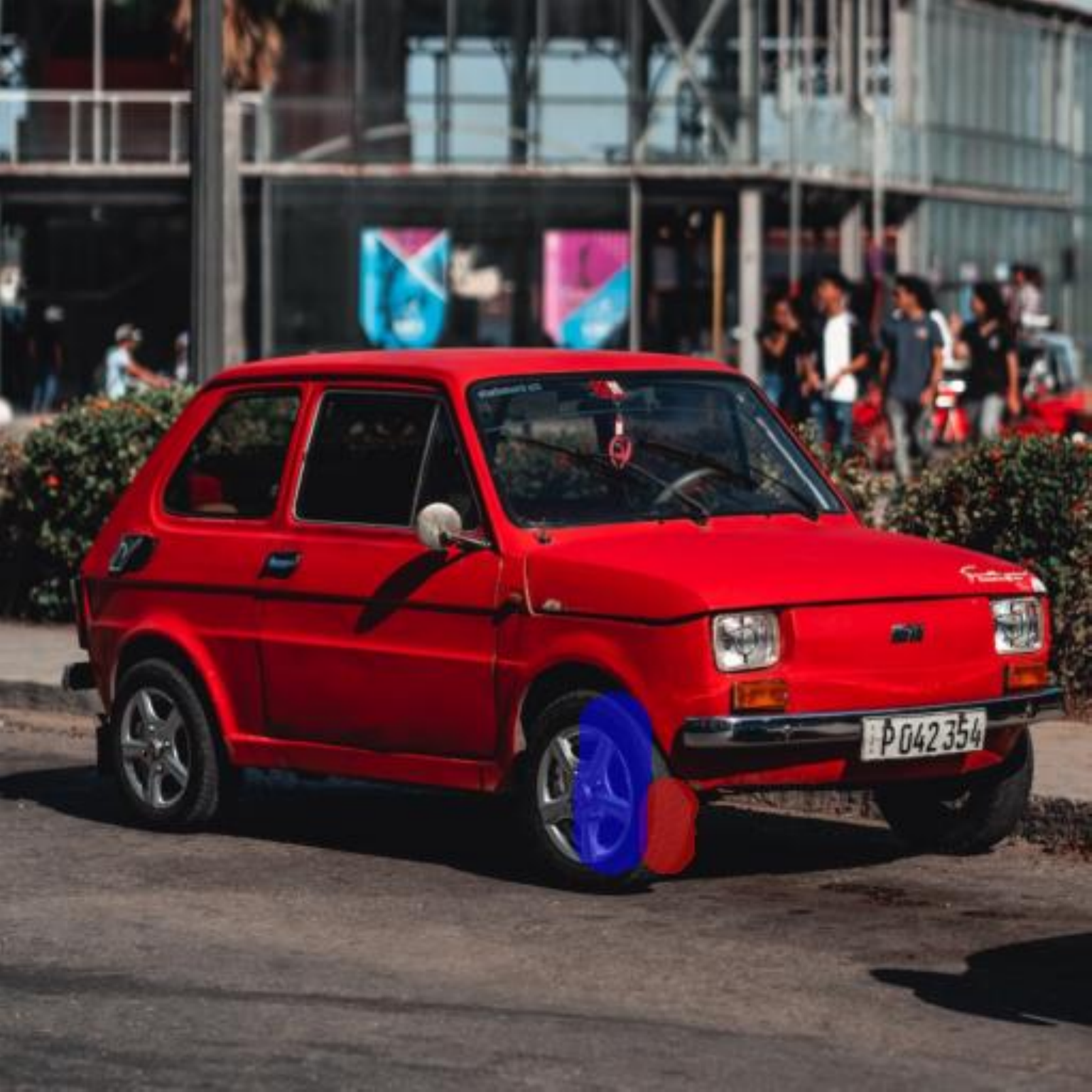}
    \end{subfigure}%
    \begin{subfigure}{.16\linewidth}
        \centering
        \includegraphics[width=\linewidth]{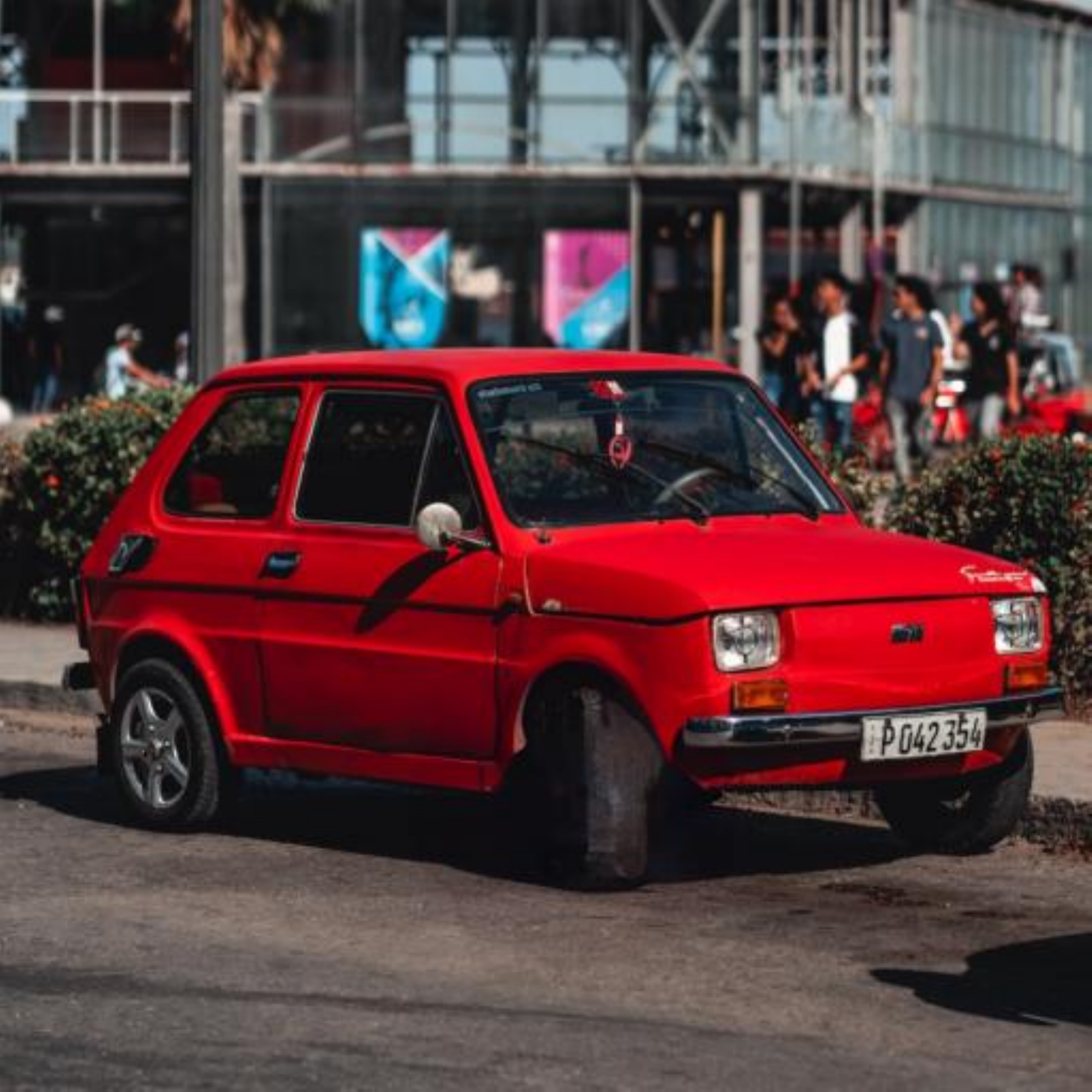}
    \end{subfigure}%


    \begin{subfigure}{.16\linewidth}
        \centering
        \includegraphics[width=\linewidth]{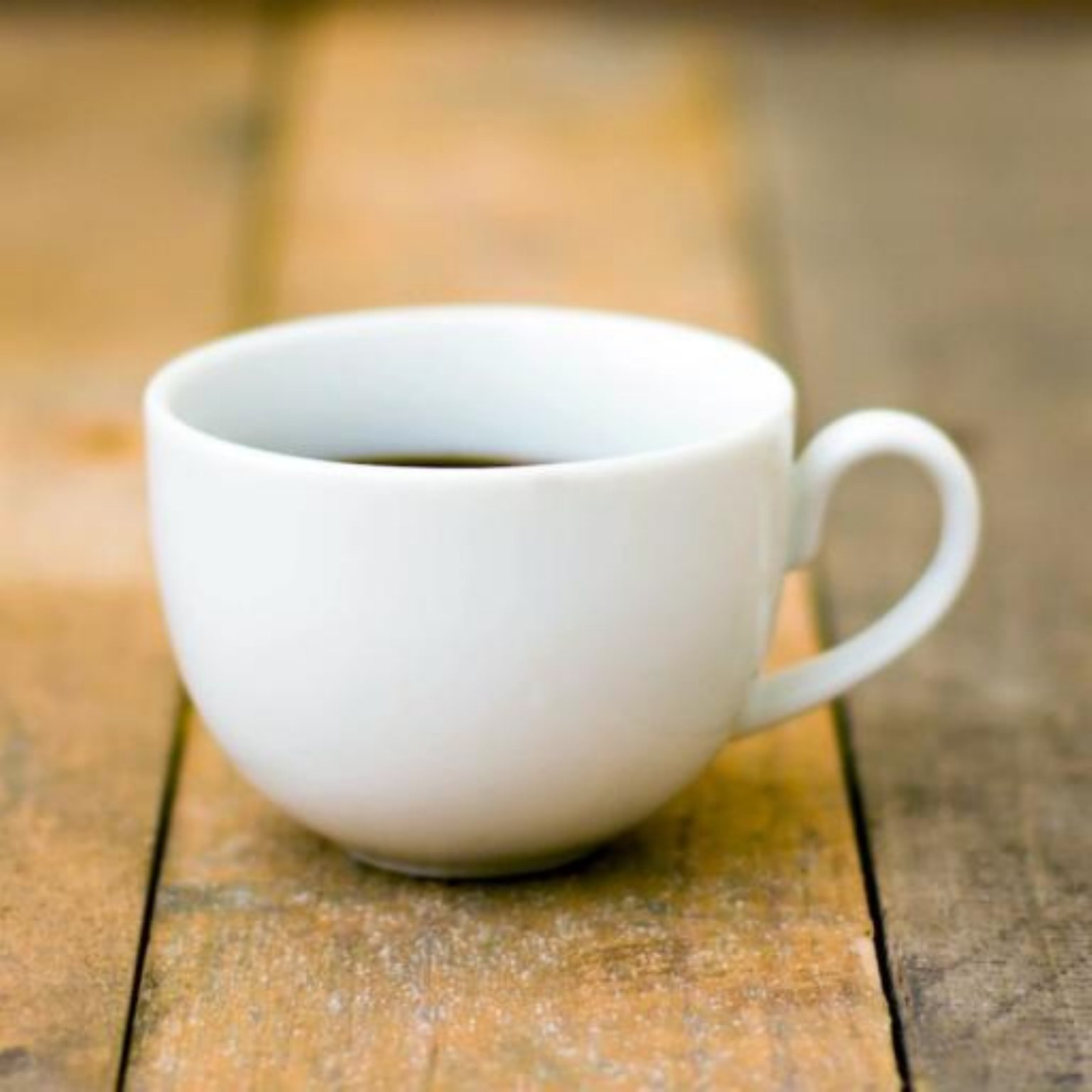}
    \end{subfigure}%
    \begin{subfigure}{.16\linewidth}
        \centering
        \includegraphics[width=\linewidth]{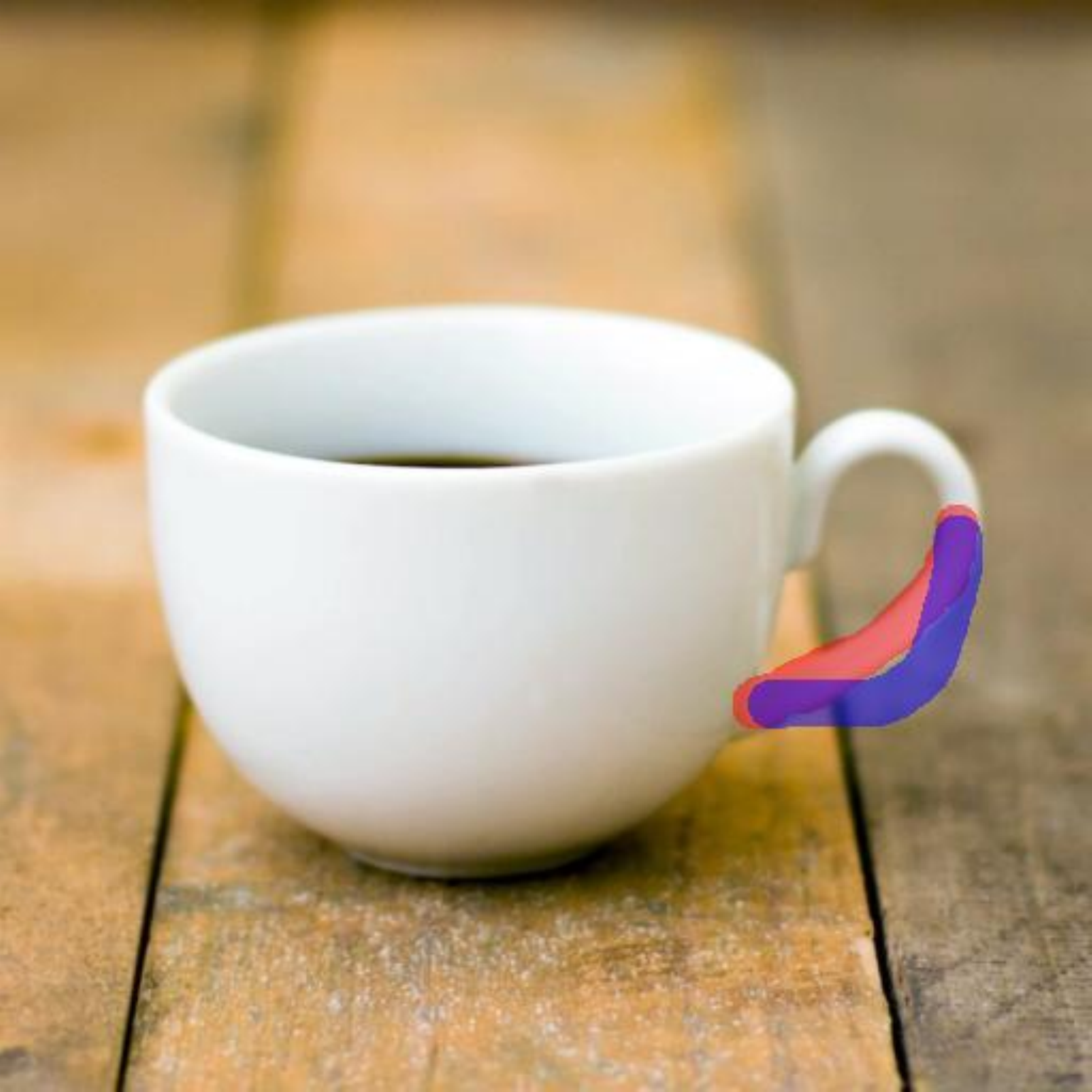}
    \end{subfigure}%
    \begin{subfigure}{.16\linewidth}
        \centering
        \includegraphics[width=\linewidth]{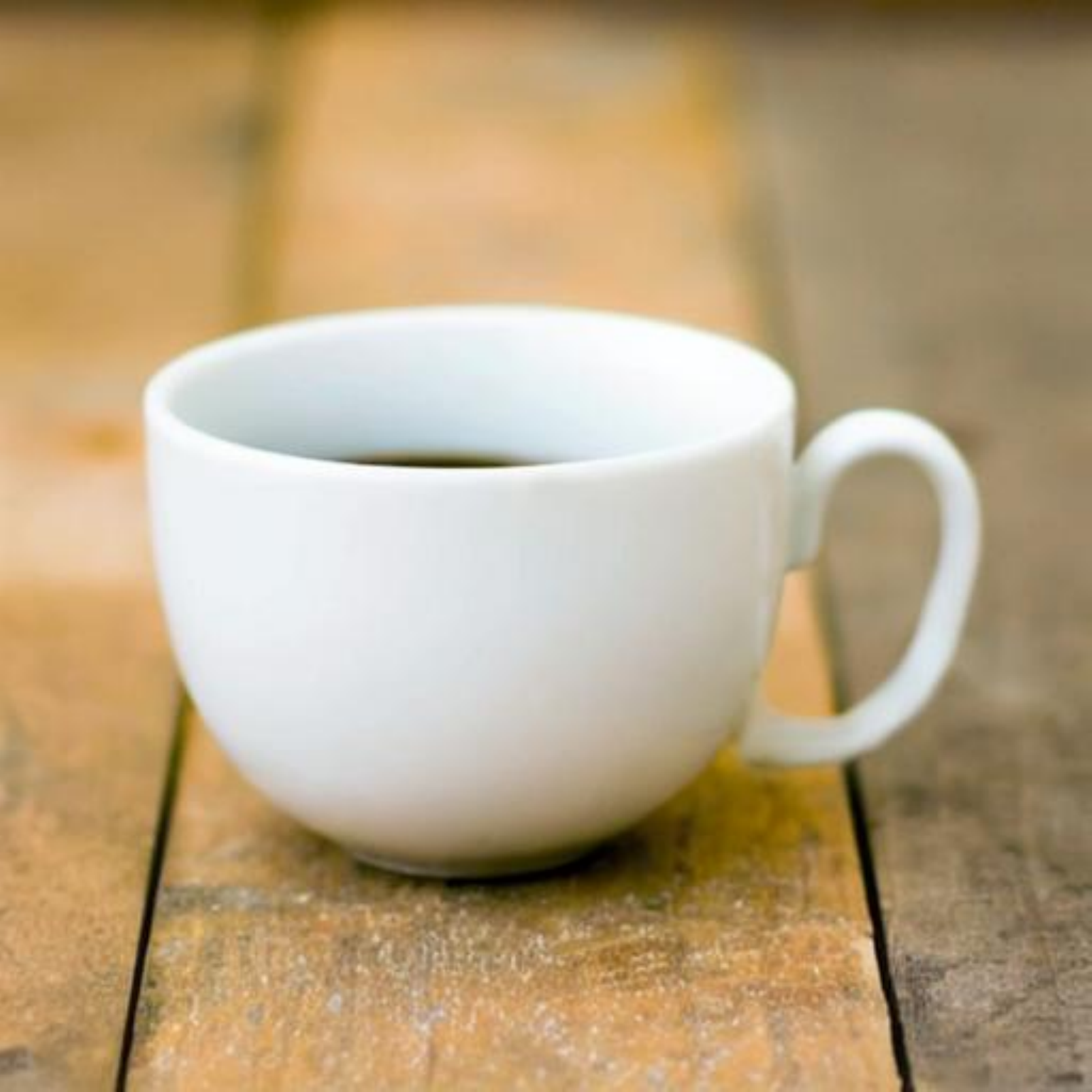}
    \end{subfigure}%

    \begin{subfigure}{.16\linewidth}
        \centering
        \includegraphics[width=\linewidth]{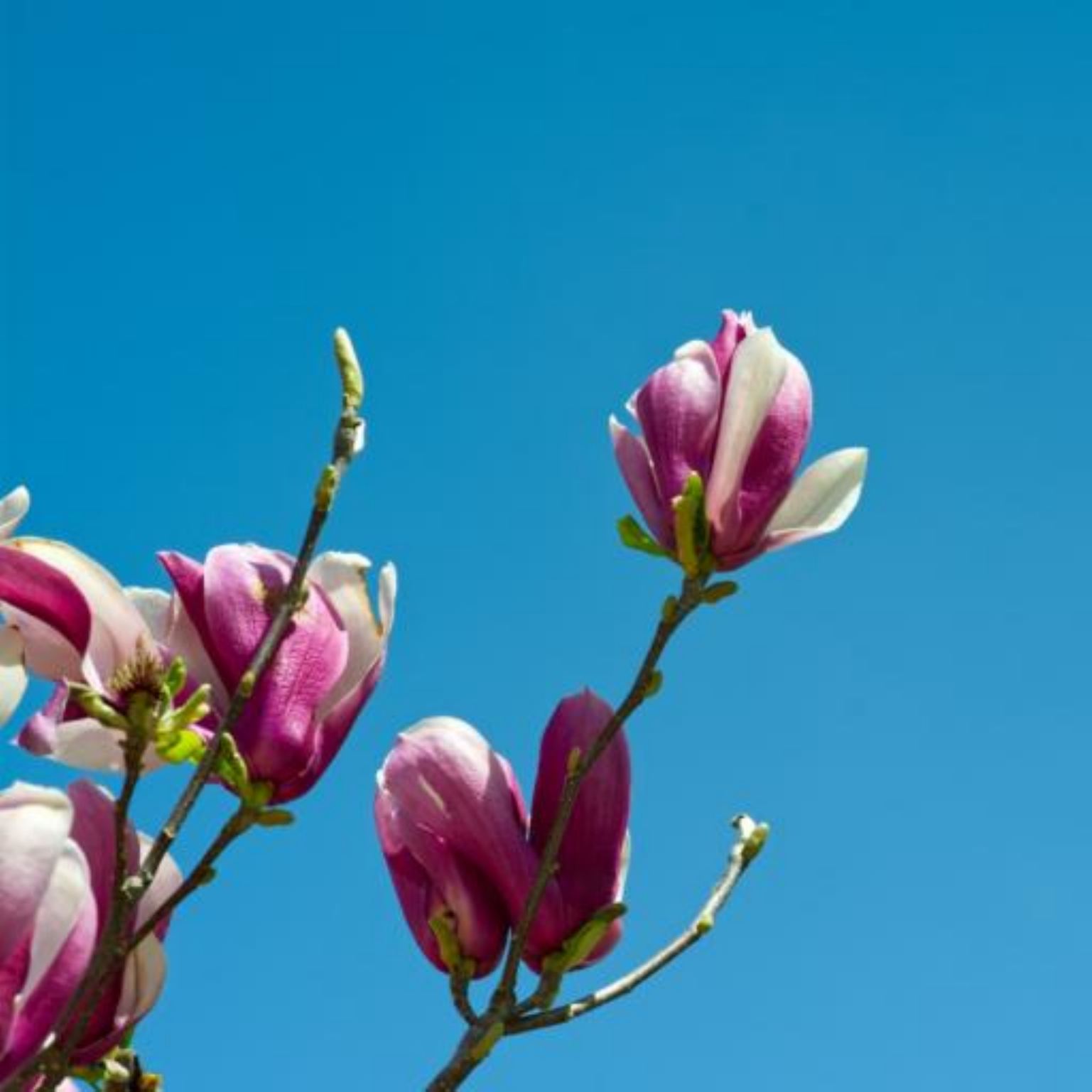}
    \end{subfigure}%
    \begin{subfigure}{.16\linewidth}
        \centering
        \includegraphics[width=\linewidth]{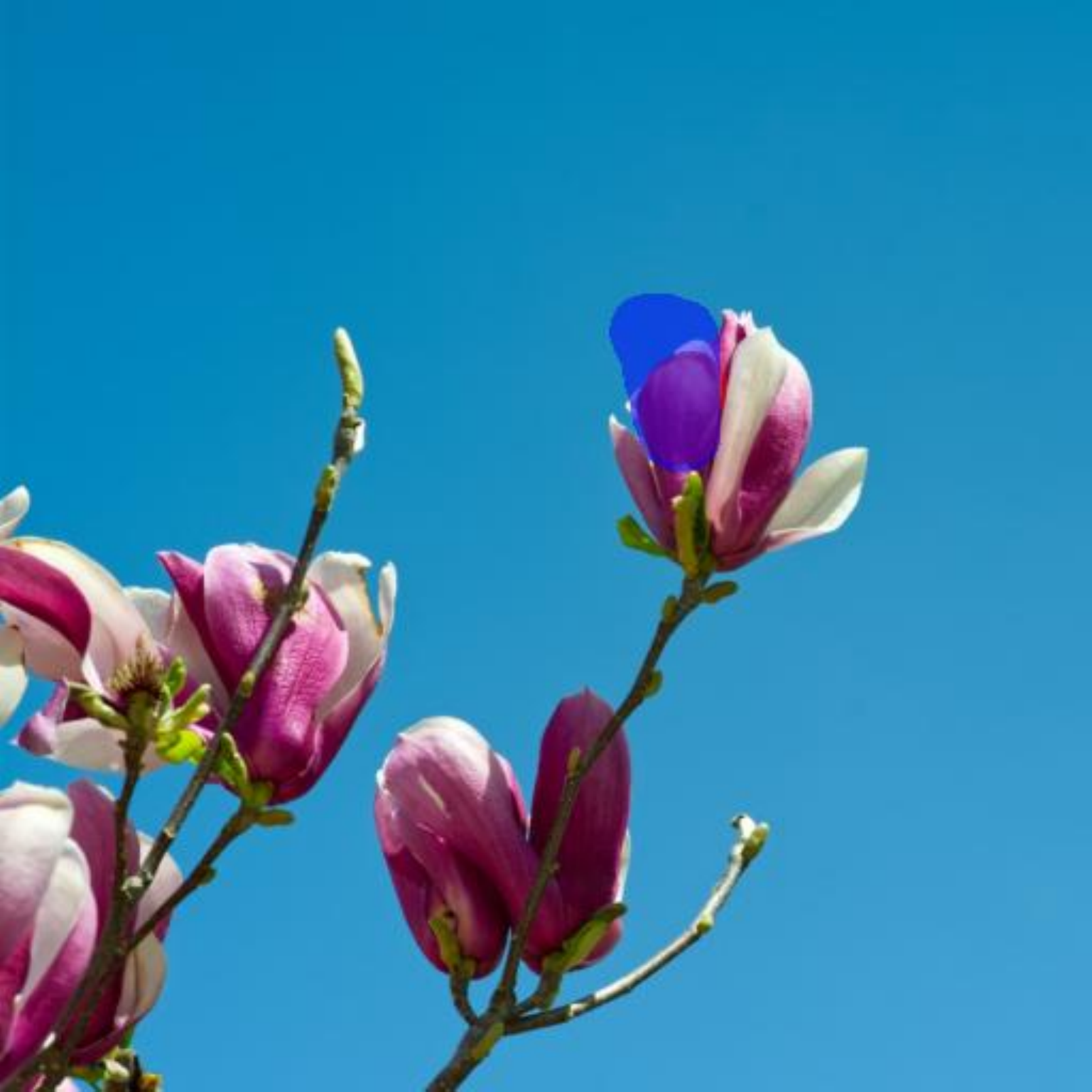}
    \end{subfigure}%
    \begin{subfigure}{.16\linewidth}
        \centering
        \includegraphics[width=\linewidth]{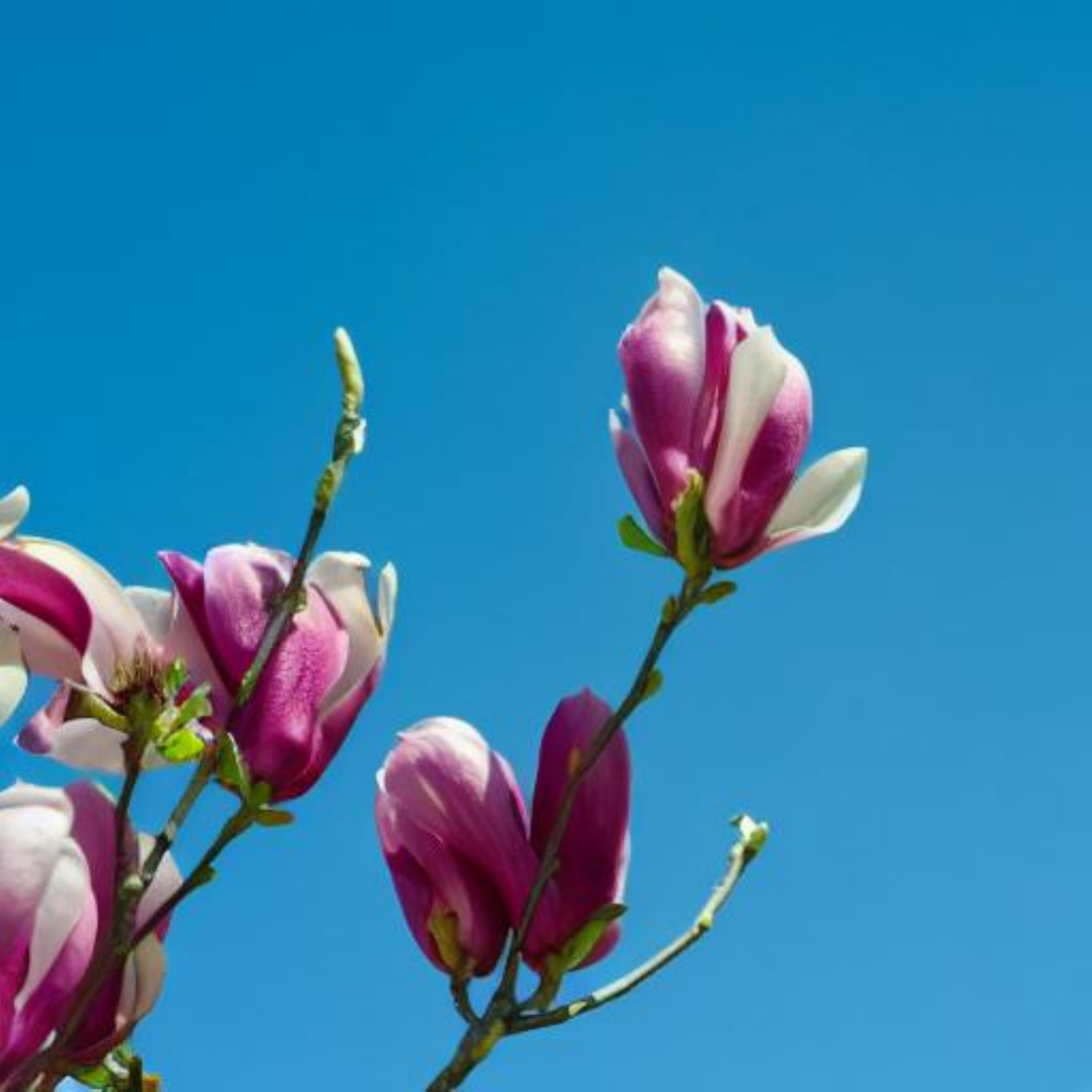}
    \end{subfigure}%


    \caption{More qualitative results.}
    \label{fig:results} 
\end{figure*}

In \cref{fig:results}, we showcase additional qualitative outcomes of RegionDrag as it processes objects from various domains.

\section{More Qualitative Comparisons}
\begin{figure*}[t]

    \centering
        \begin{subfigure}{.16\linewidth}
        \caption*{Point input}
    \end{subfigure}%
    \begin{subfigure}{.16\linewidth}
        \caption*{DragDiff.}
    \end{subfigure}%
    \hspace{-2mm}
    \begin{subfigure}{.16\linewidth}
        \caption*{SDE-Drag}
    \end{subfigure}%
    \begin{subfigure}{.16\linewidth}
        \caption*{DiffEditor}
    \end{subfigure}%
    \hspace{1mm}
    \begin{subfigure}{.16\linewidth}
        \caption*{Region input}
    \end{subfigure}%
    \begin{subfigure}{.16\linewidth}
        \caption*{Our results}
    \end{subfigure}

    \begin{subfigure}{.16\linewidth}
        \centering
        \includegraphics[width=\linewidth]{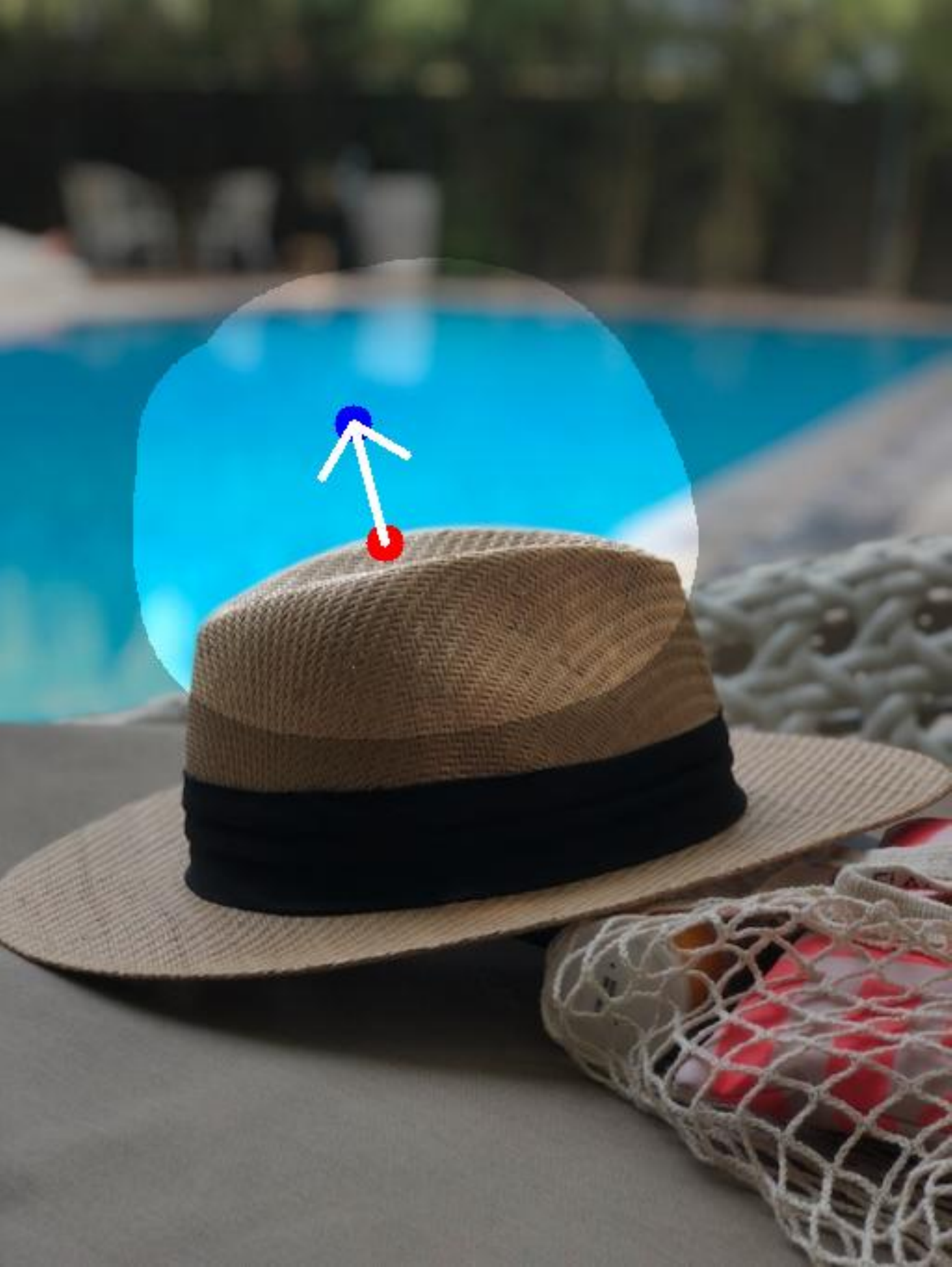}
    \end{subfigure}%
    \begin{subfigure}{.16\linewidth}
        \centering
        \includegraphics[width=\linewidth]{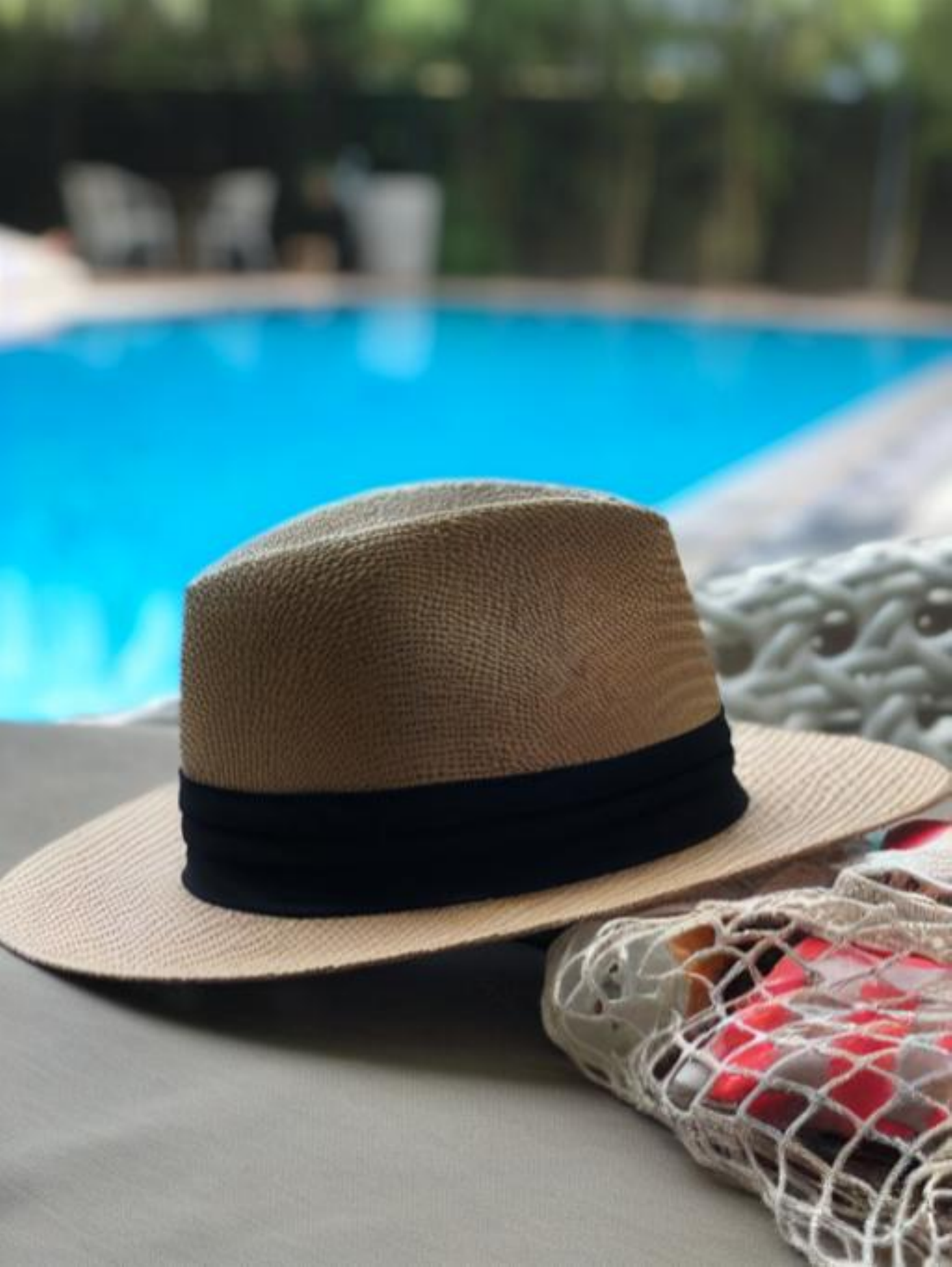}
    \end{subfigure}%
    \begin{subfigure}{.16\linewidth}
        \centering
        \includegraphics[width=\linewidth]{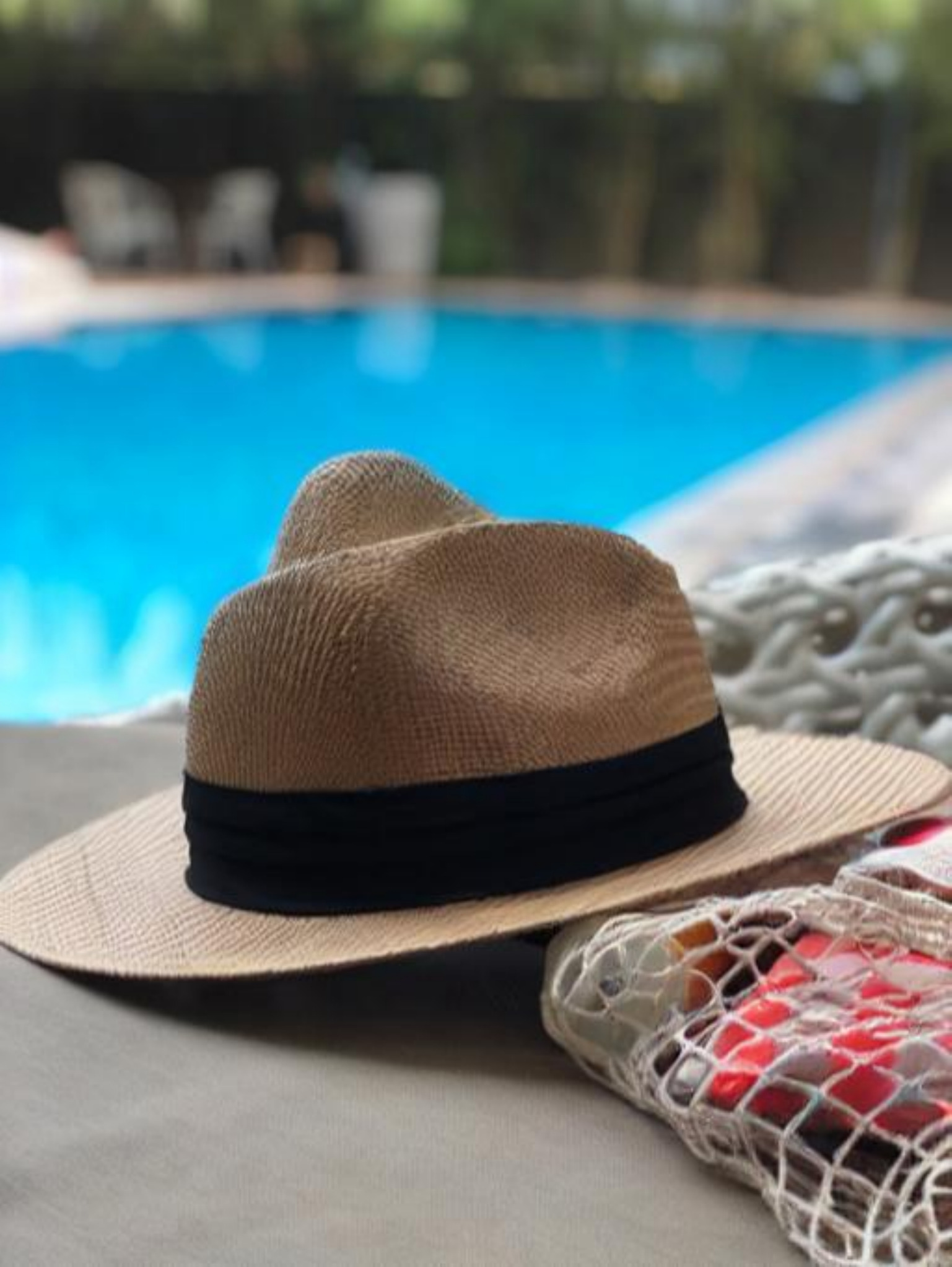}
    \end{subfigure}%
    \begin{subfigure}{.16\linewidth}
        \centering
        \includegraphics[width=\linewidth]{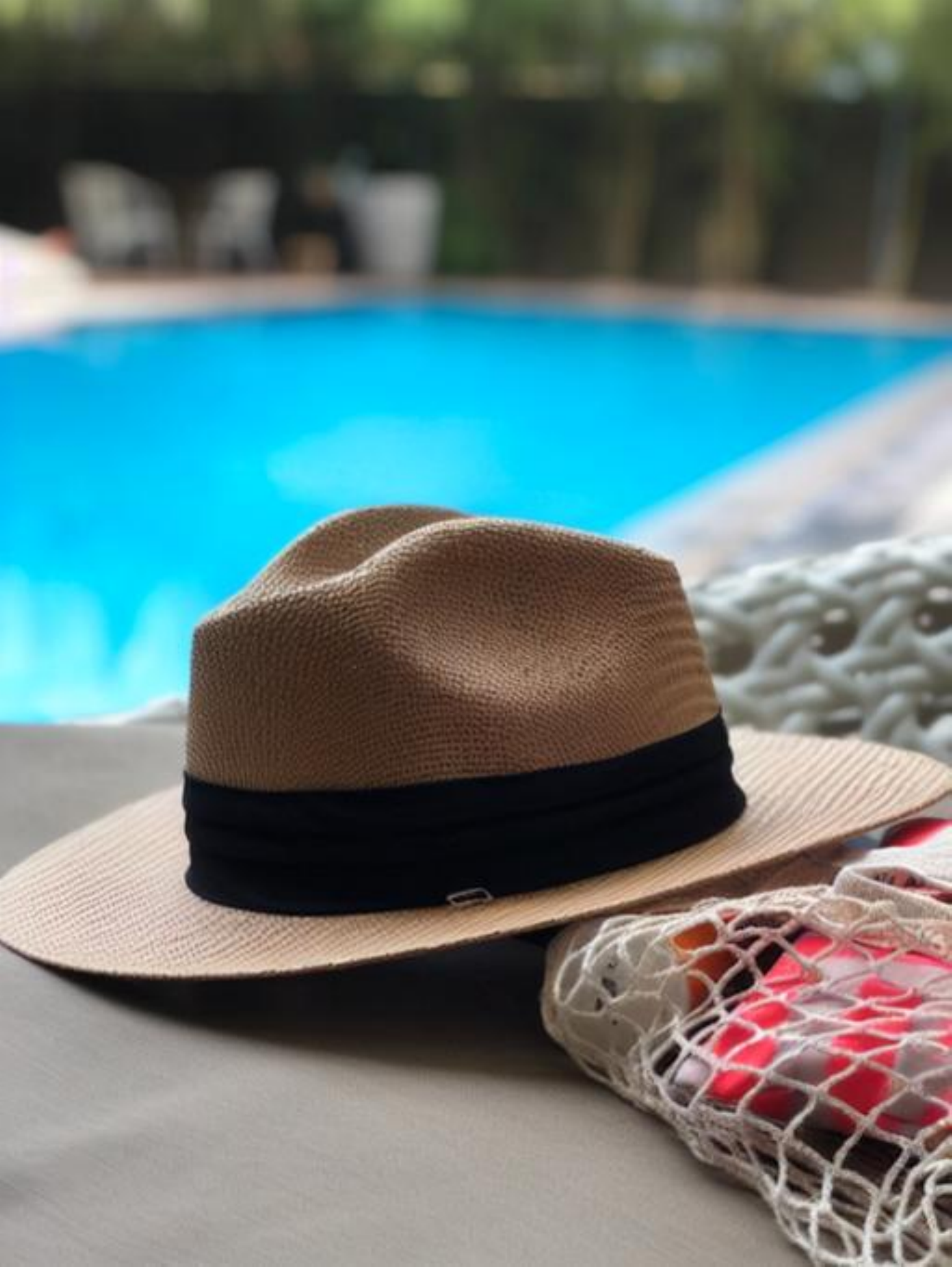}
    \end{subfigure}%
    \hspace{1mm}
    \begin{subfigure}{.16\linewidth}
        \centering
        \includegraphics[width=\linewidth]{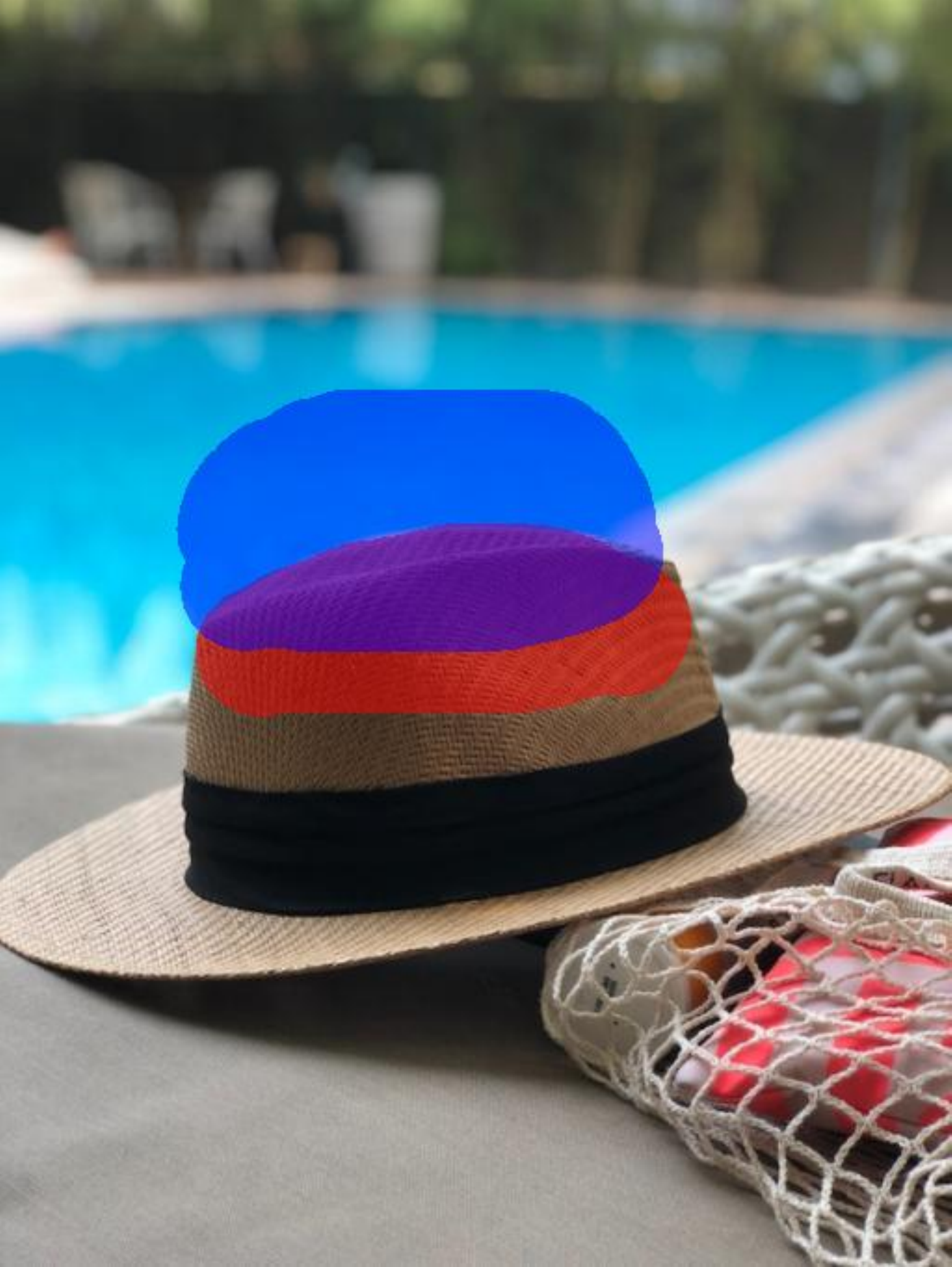}
    \end{subfigure}%
    \begin{subfigure}{.16\linewidth}
        \centering
        \includegraphics[width=\linewidth]{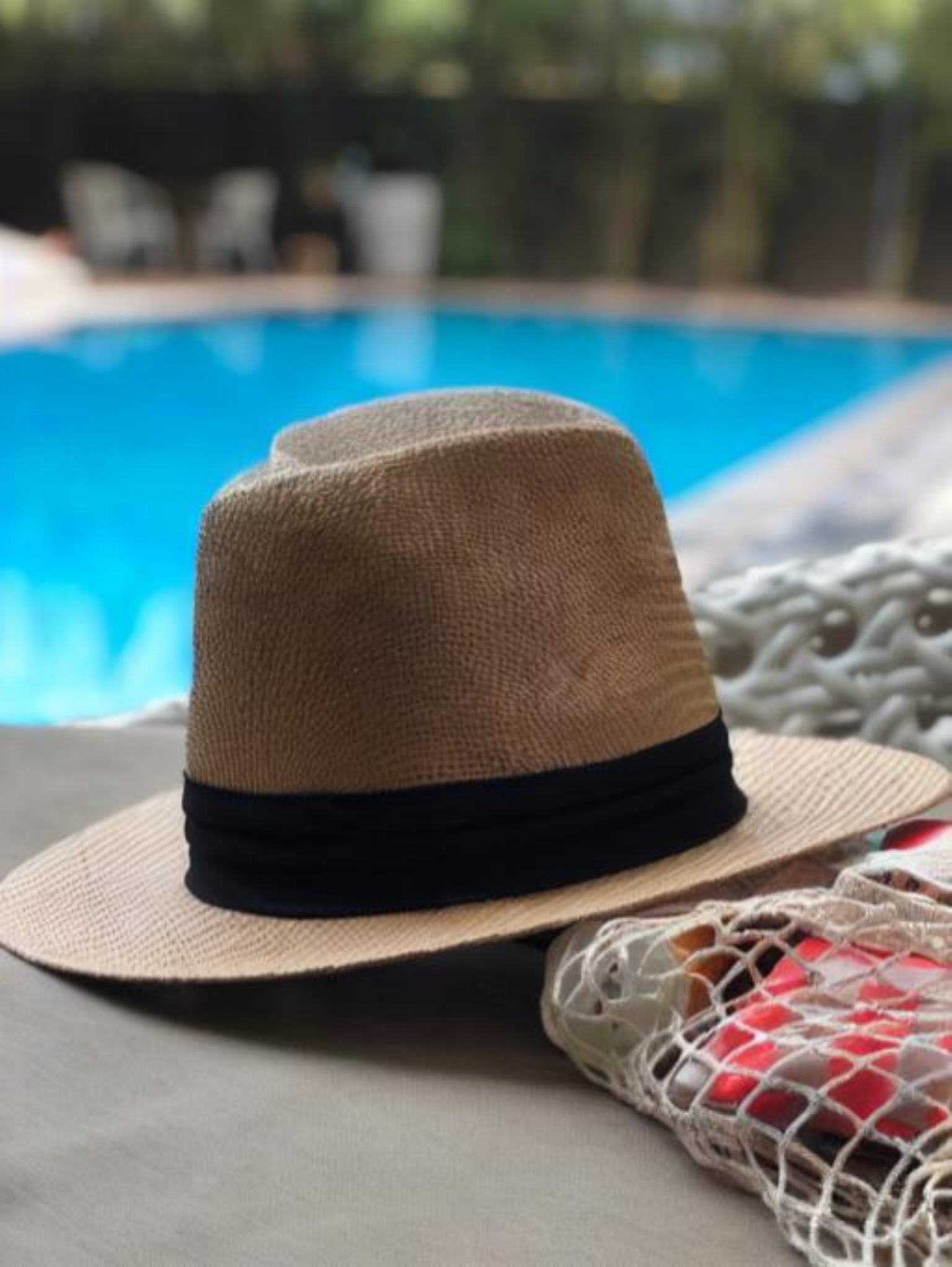}
    \end{subfigure}%
    
    \begin{subfigure}{.16\linewidth}
        \centering
        \includegraphics[width=\linewidth]{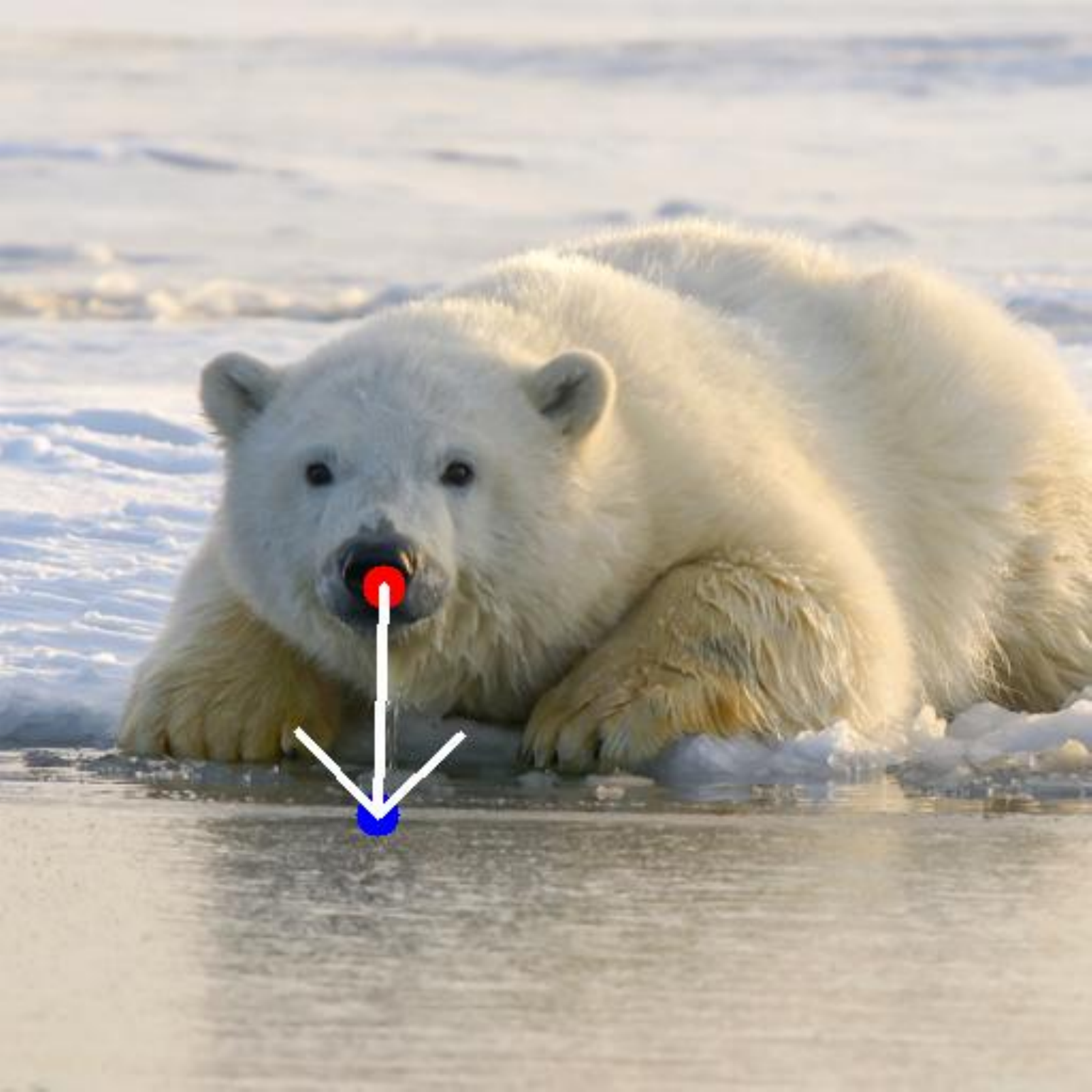}
    \end{subfigure}%
    \begin{subfigure}{.16\linewidth}
        \centering
        \includegraphics[width=\linewidth]{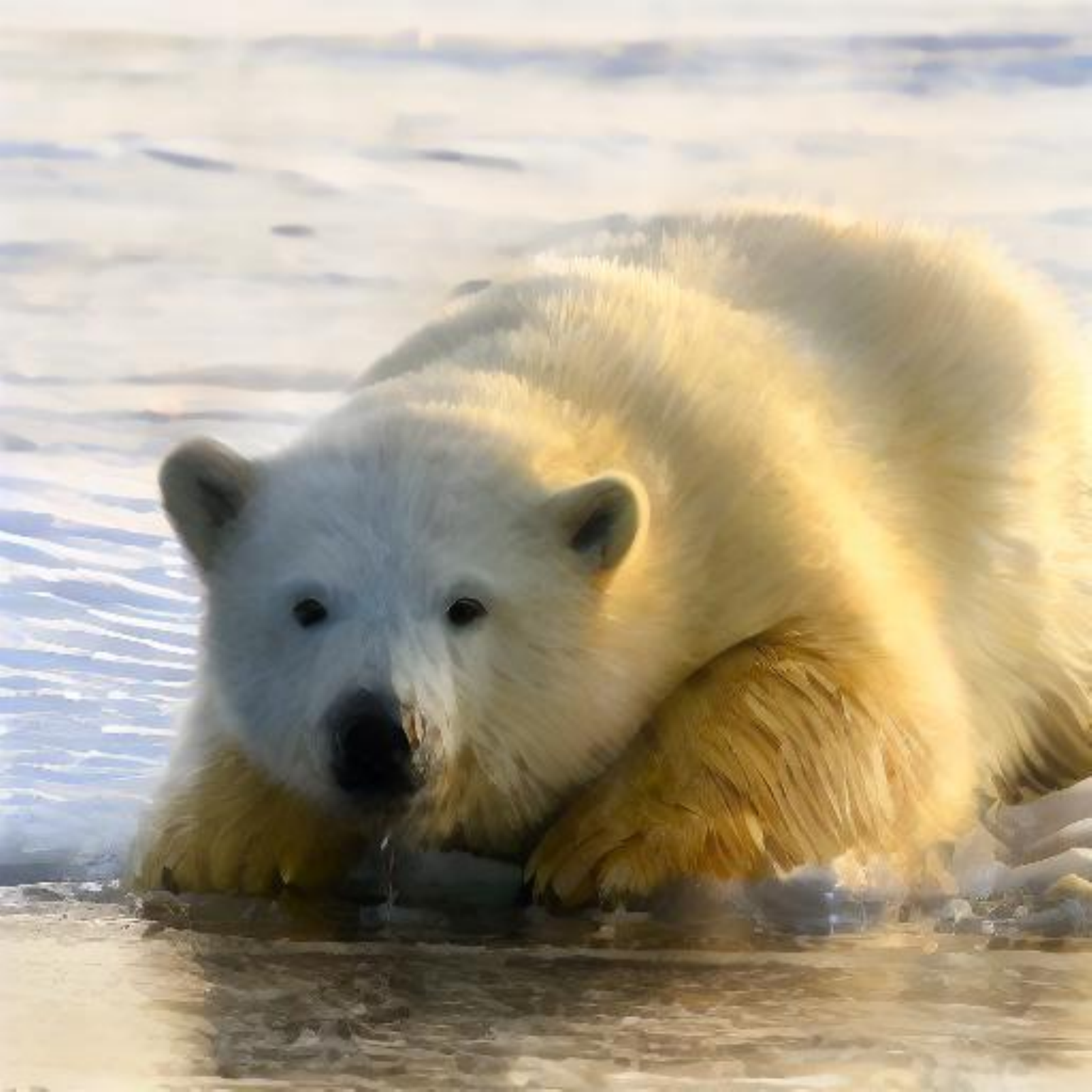}
    \end{subfigure}%
    \begin{subfigure}{.16\linewidth}
        \centering
        \includegraphics[width=\linewidth]{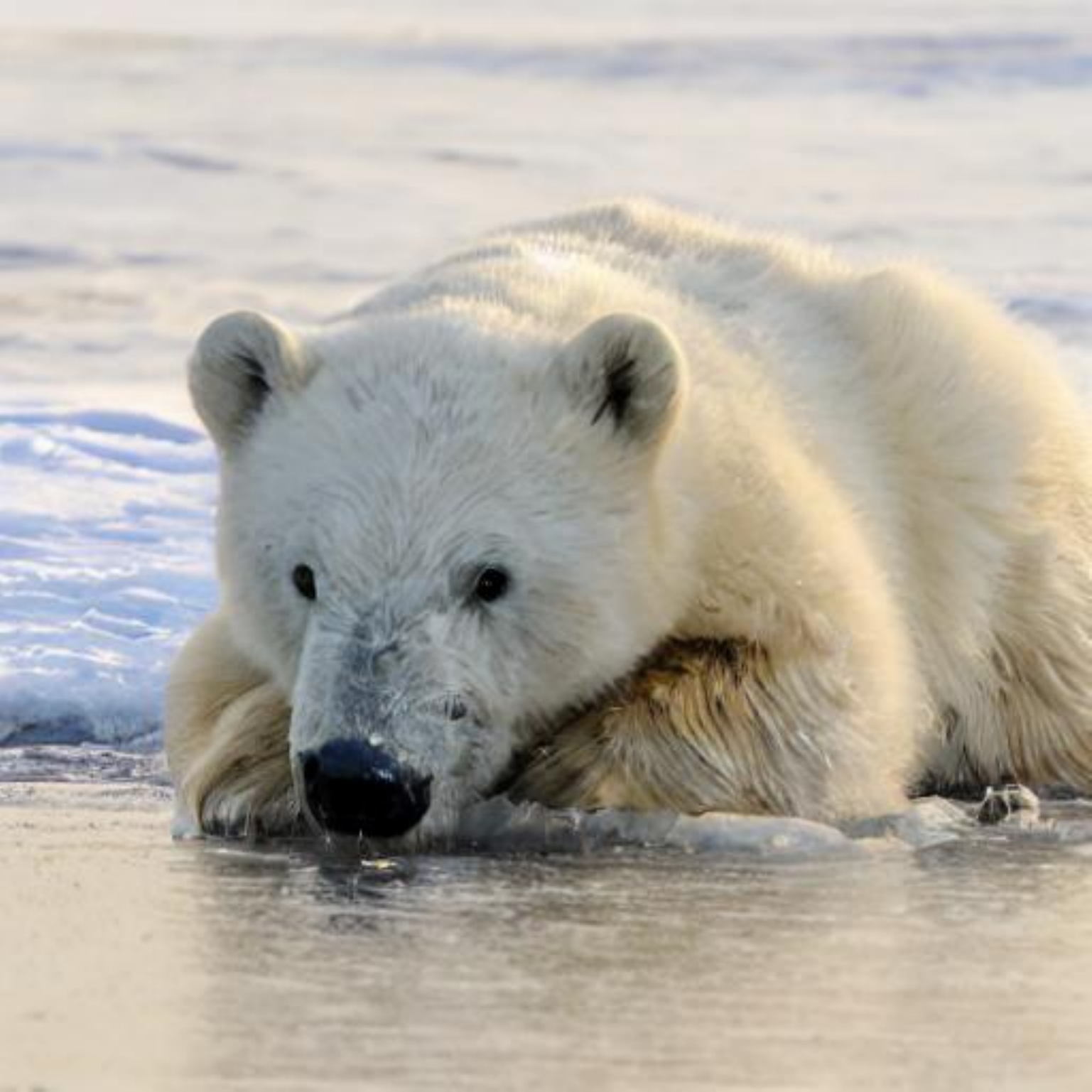}
    \end{subfigure}%
    \begin{subfigure}{.16\linewidth}
        \centering
        \includegraphics[width=\linewidth]{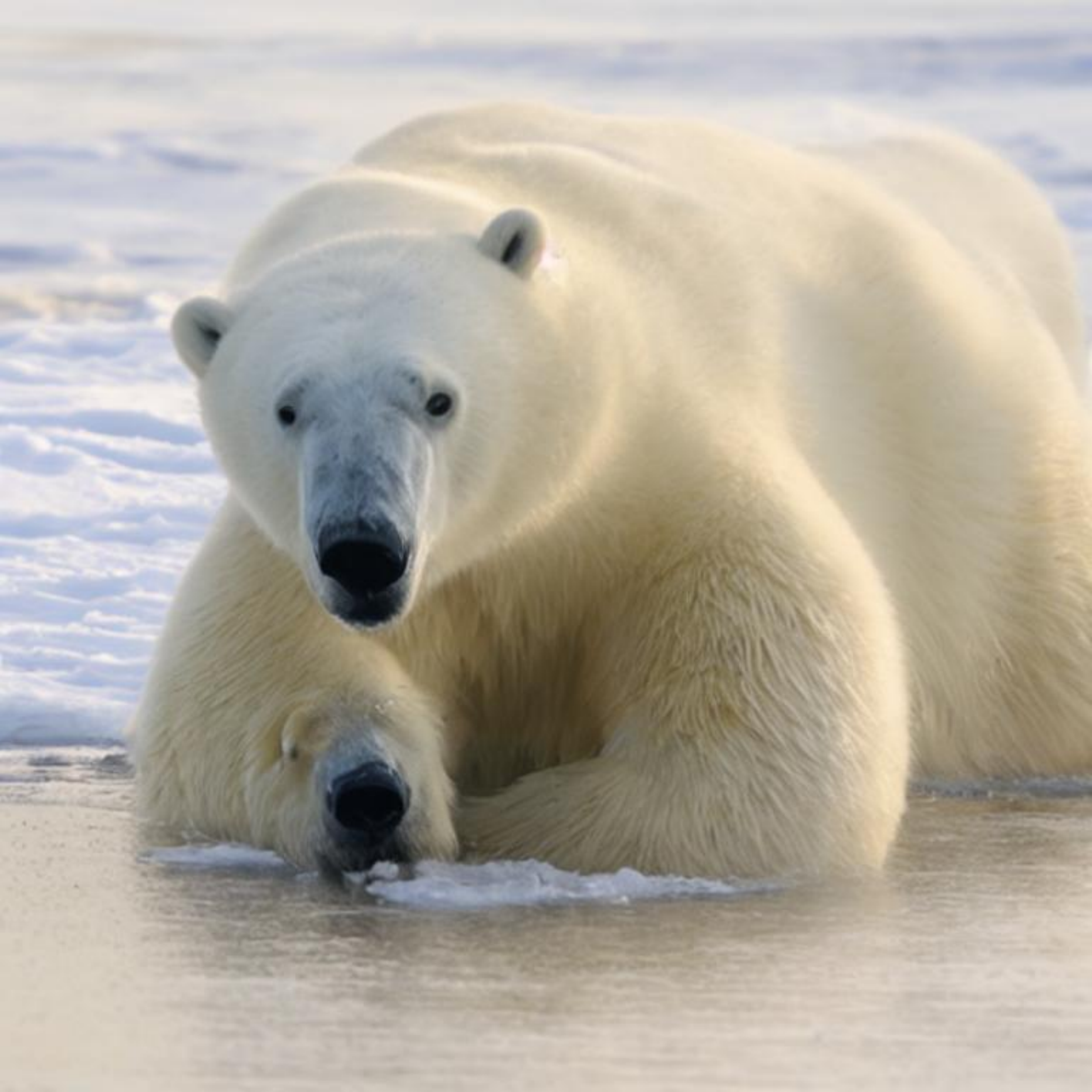}
    \end{subfigure}%
    \hspace{1mm}
    \begin{subfigure}{.16\linewidth}
        \centering
        \includegraphics[width=\linewidth]{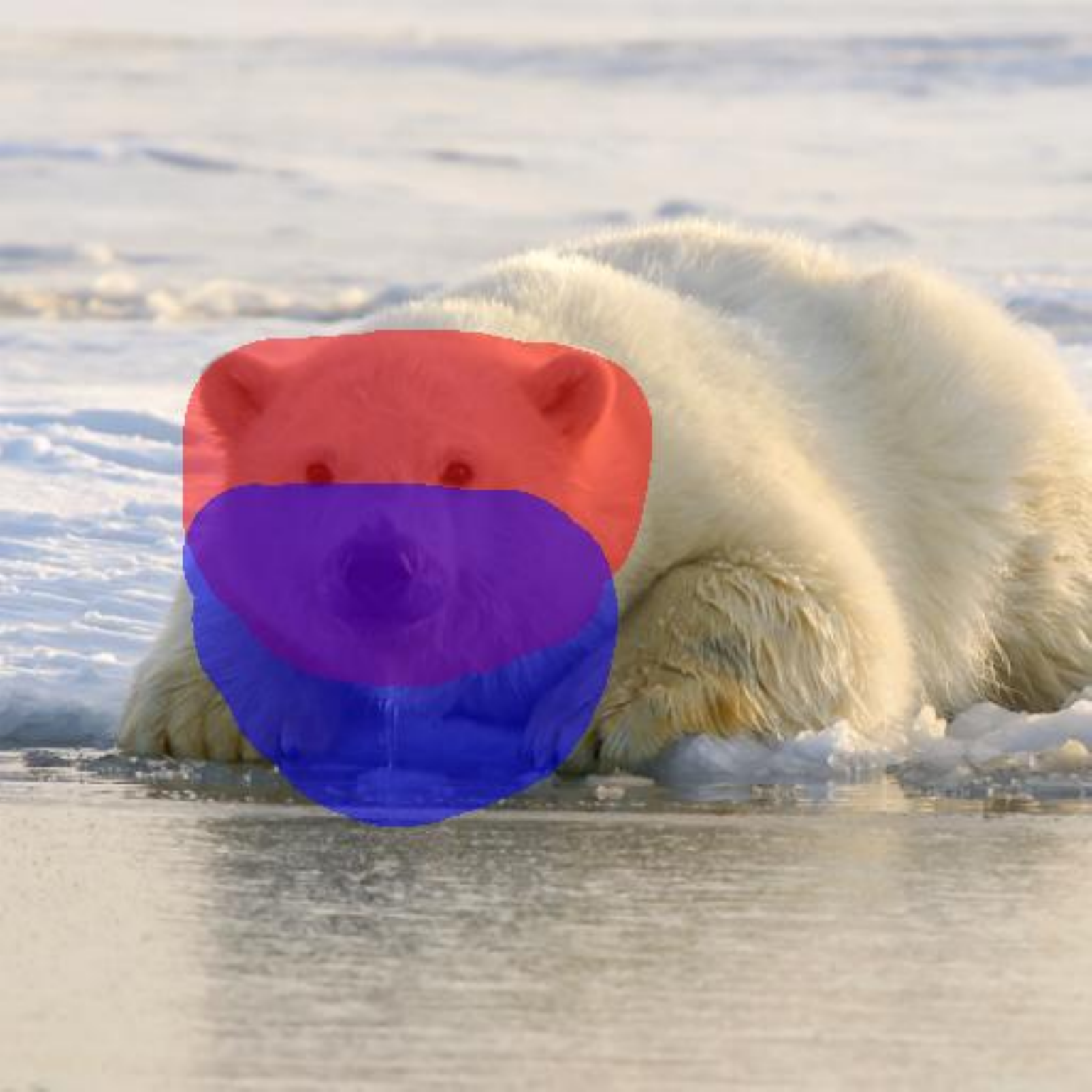}
    \end{subfigure}%
    \begin{subfigure}{.16\linewidth}
        \centering
        \includegraphics[width=\linewidth]{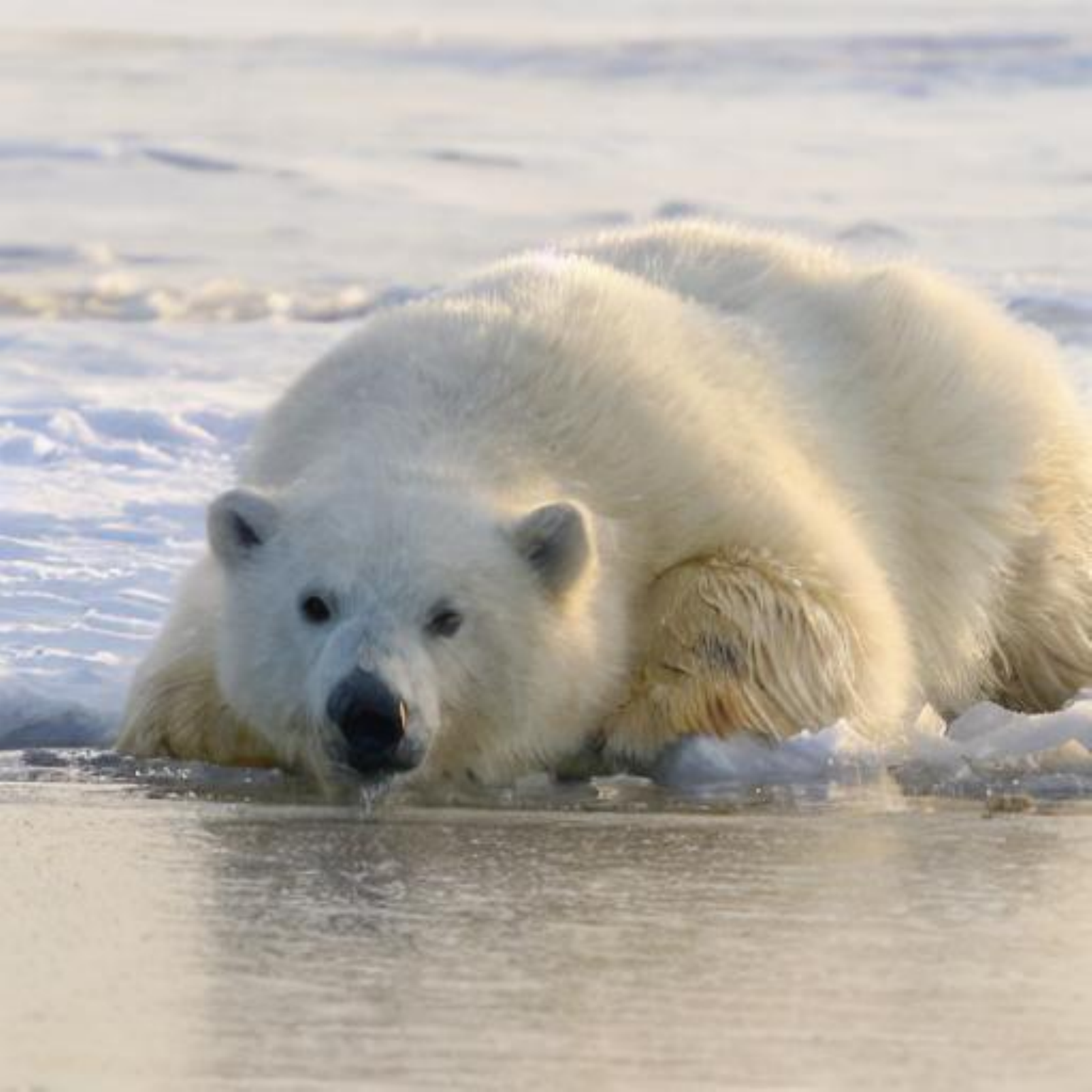}
    \end{subfigure}%

    \begin{subfigure}{.16\linewidth}
        \centering
        \includegraphics[width=\linewidth]{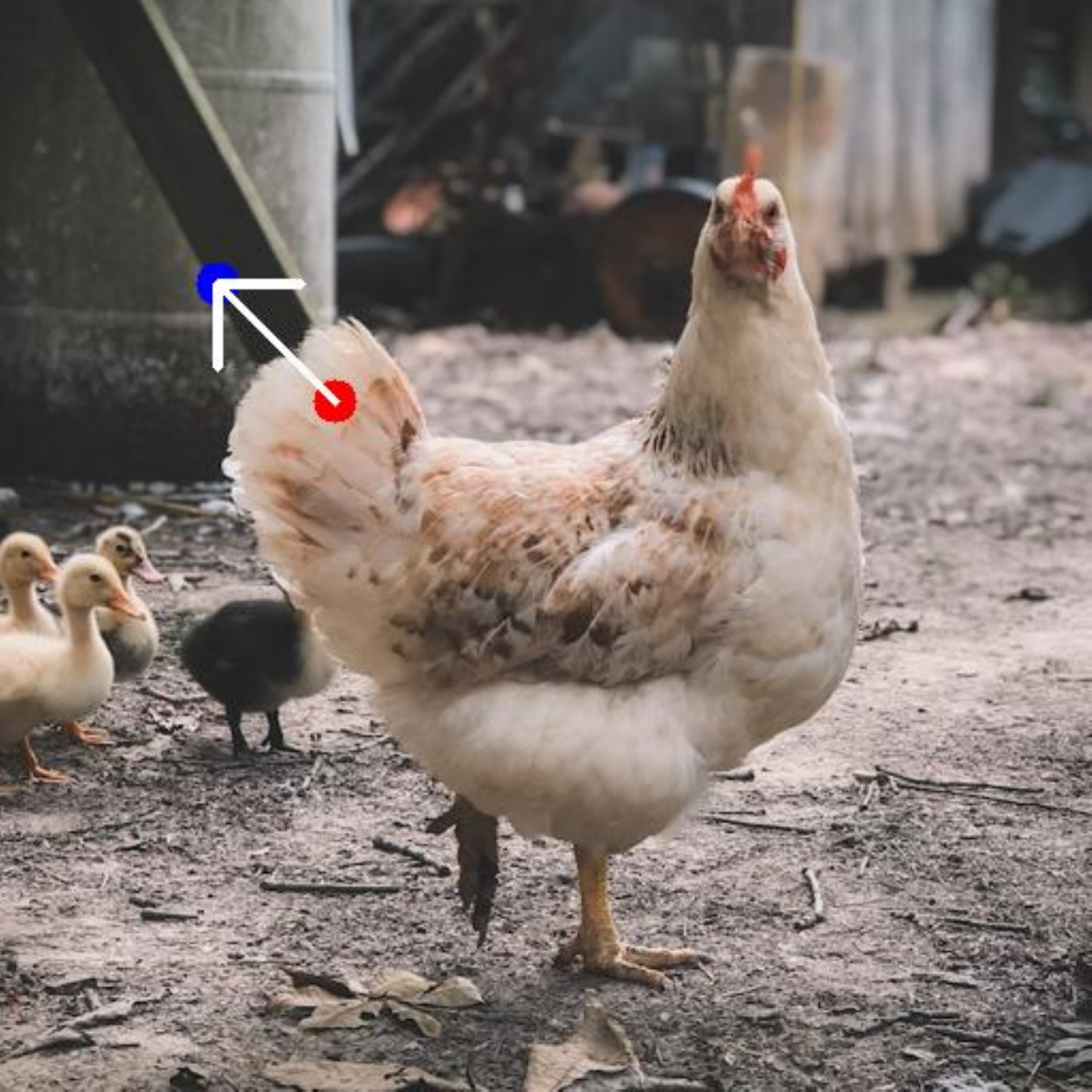}
    \end{subfigure}%
    \begin{subfigure}{.16\linewidth}
        \centering
        \includegraphics[width=\linewidth]{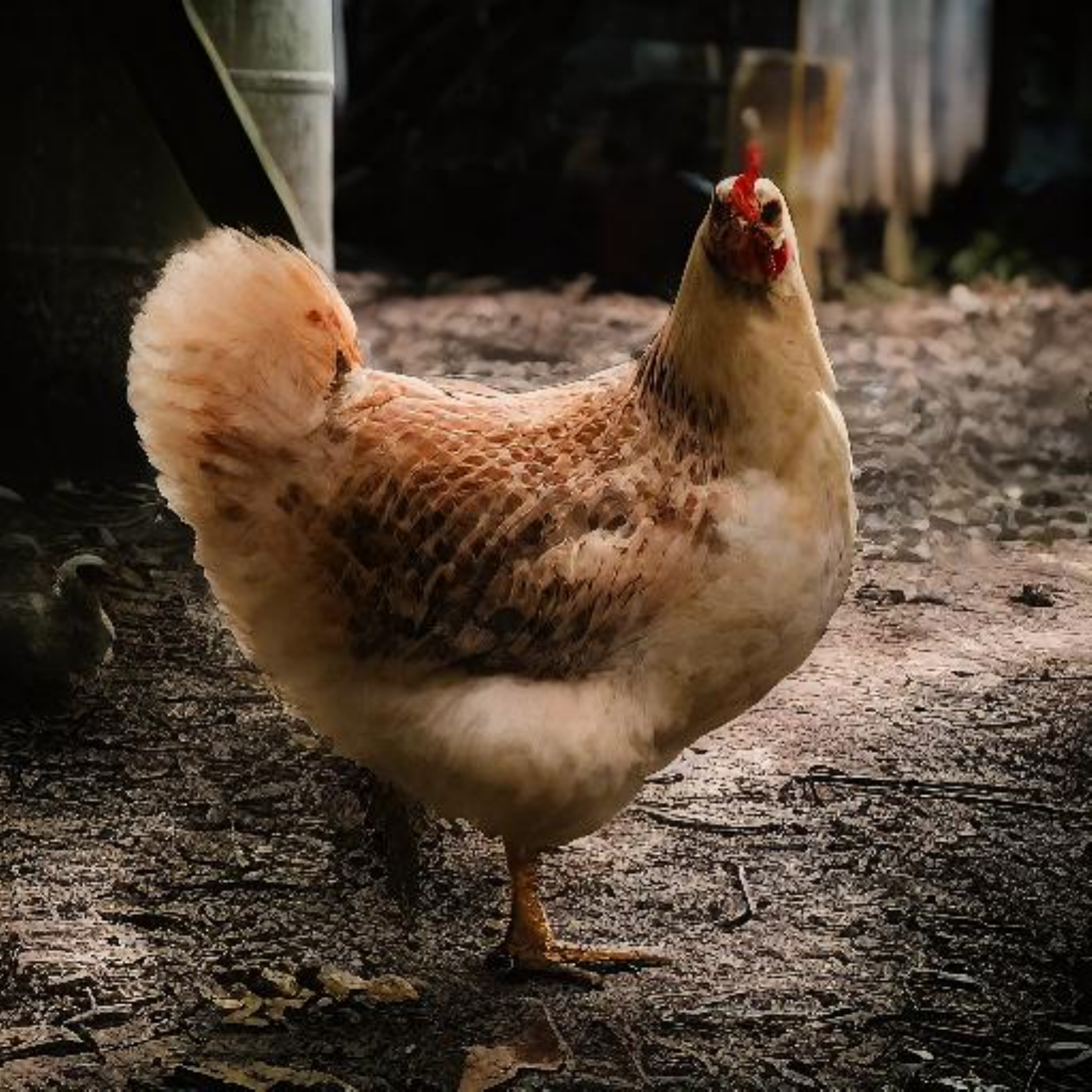}
    \end{subfigure}%
    \begin{subfigure}{.16\linewidth}
        \centering
        \includegraphics[width=\linewidth]{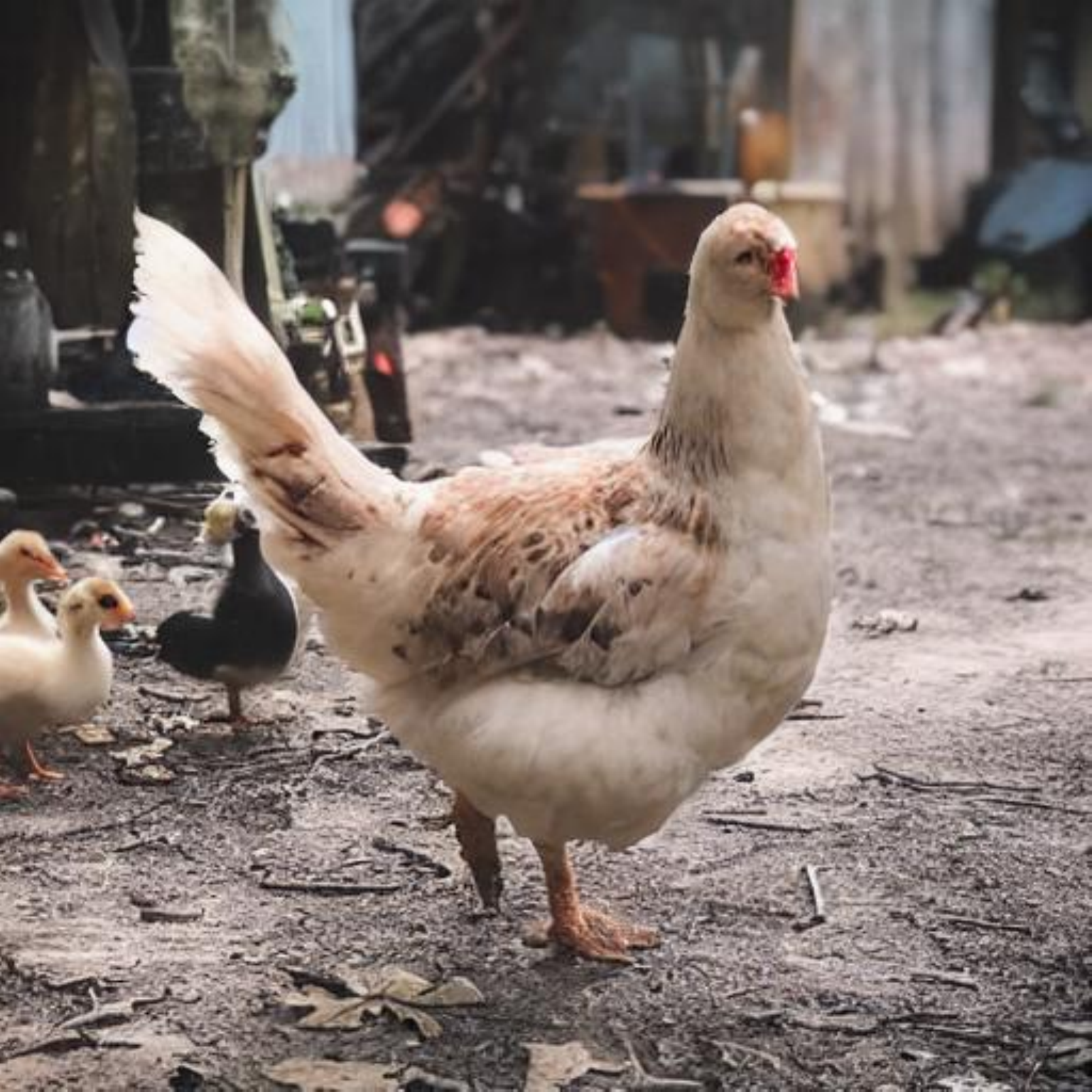}
    \end{subfigure}%
    \begin{subfigure}{.16\linewidth}
        \centering
        \includegraphics[width=\linewidth]{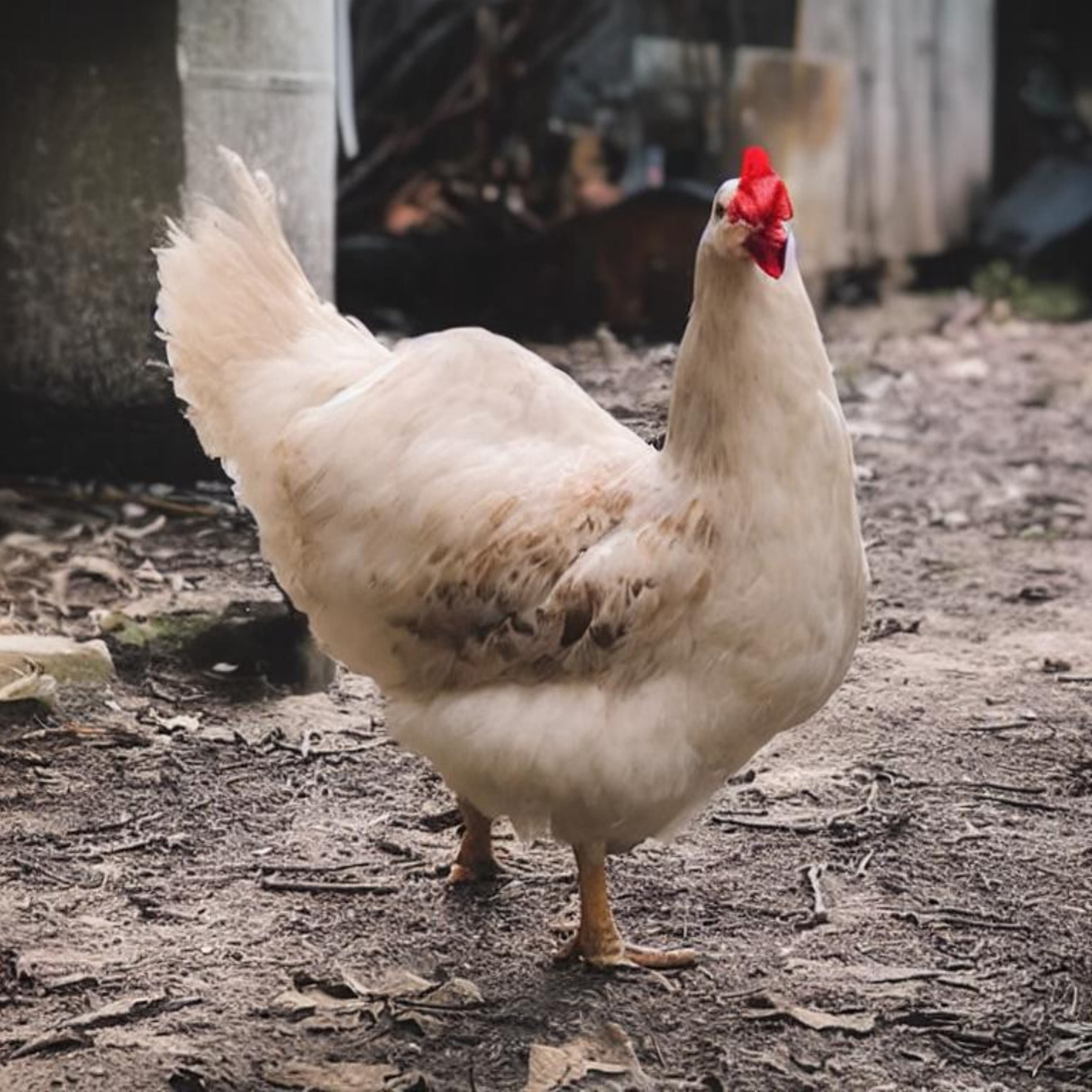}
    \end{subfigure}%
    \hspace{1mm}
    \begin{subfigure}{.16\linewidth}
        \centering
        \includegraphics[width=\linewidth]{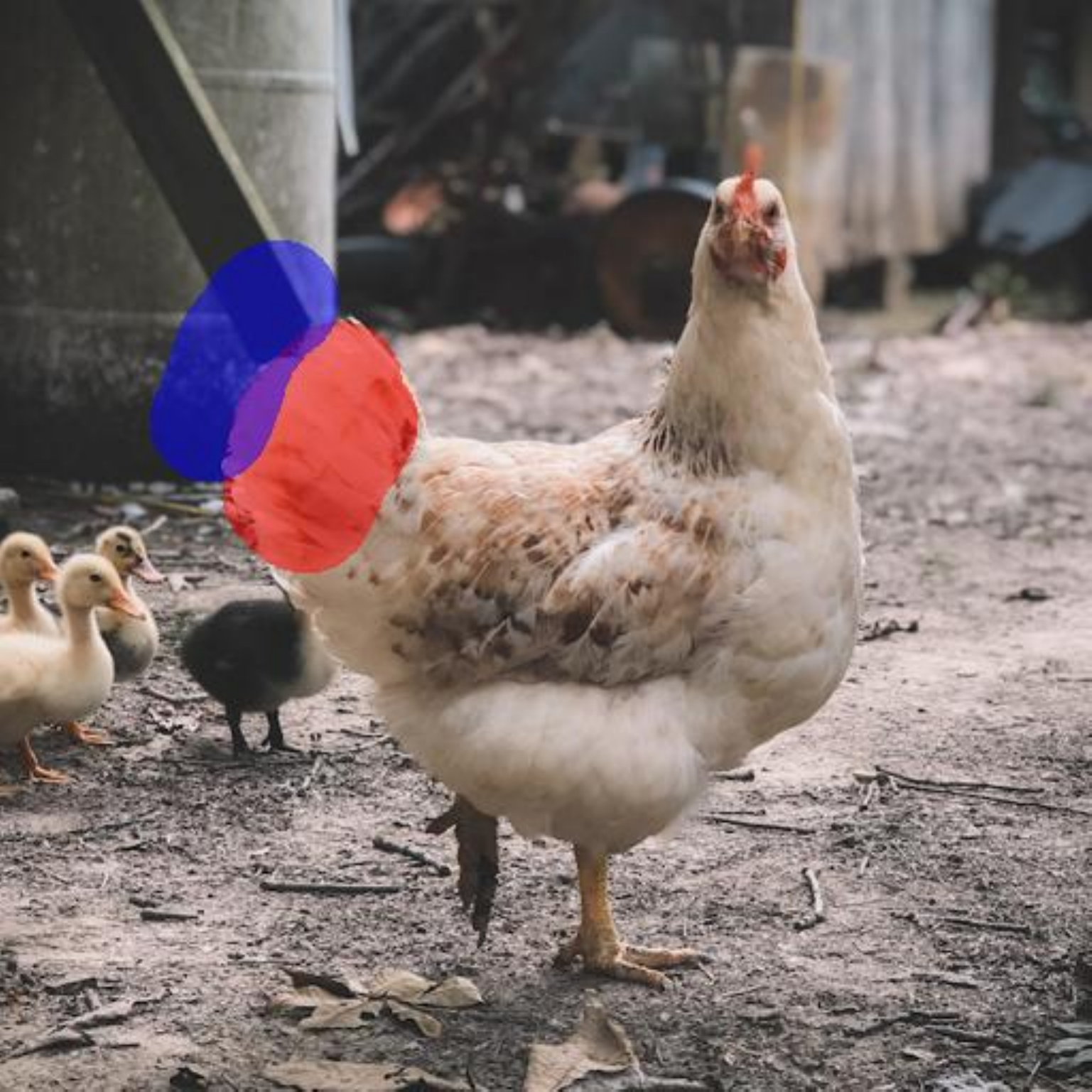}
    \end{subfigure}%
    \begin{subfigure}{.16\linewidth}
        \centering
        \includegraphics[width=\linewidth]{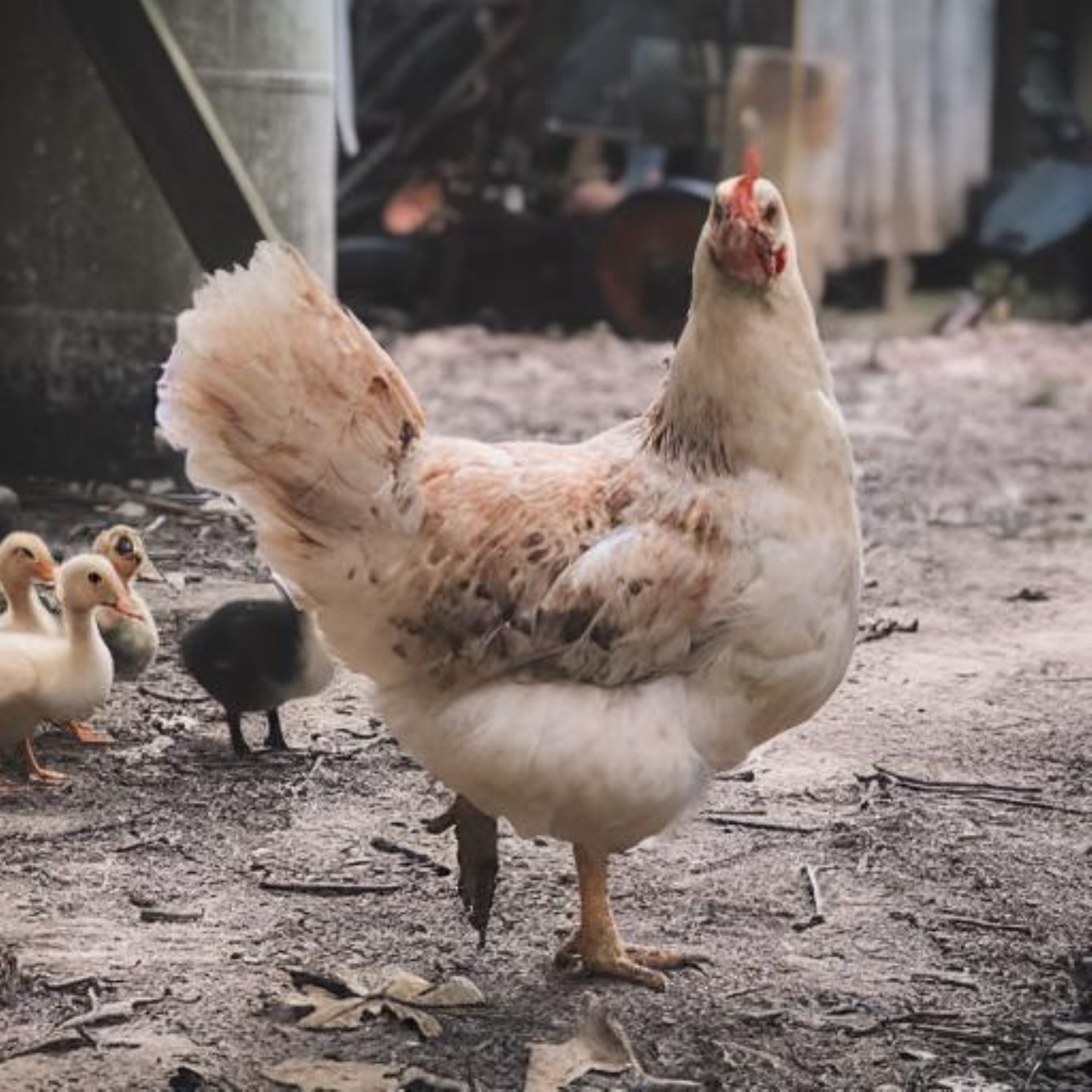}
    \end{subfigure}%
    
     \begin{subfigure}{.16\linewidth}
        \centering
        \includegraphics[width=\linewidth]{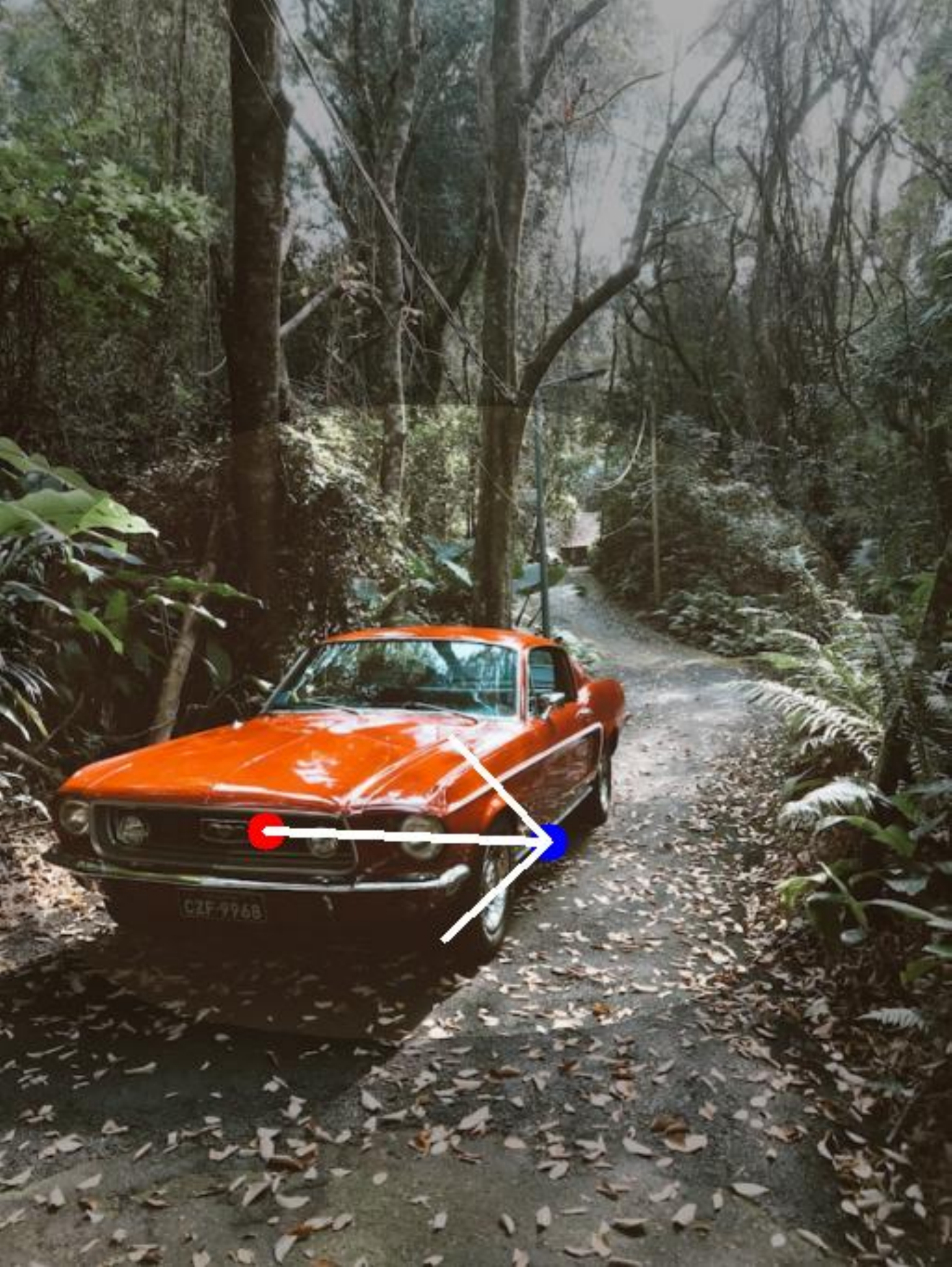}
    \end{subfigure}%
    \begin{subfigure}{.16\linewidth}
        \centering
        \includegraphics[width=\linewidth]{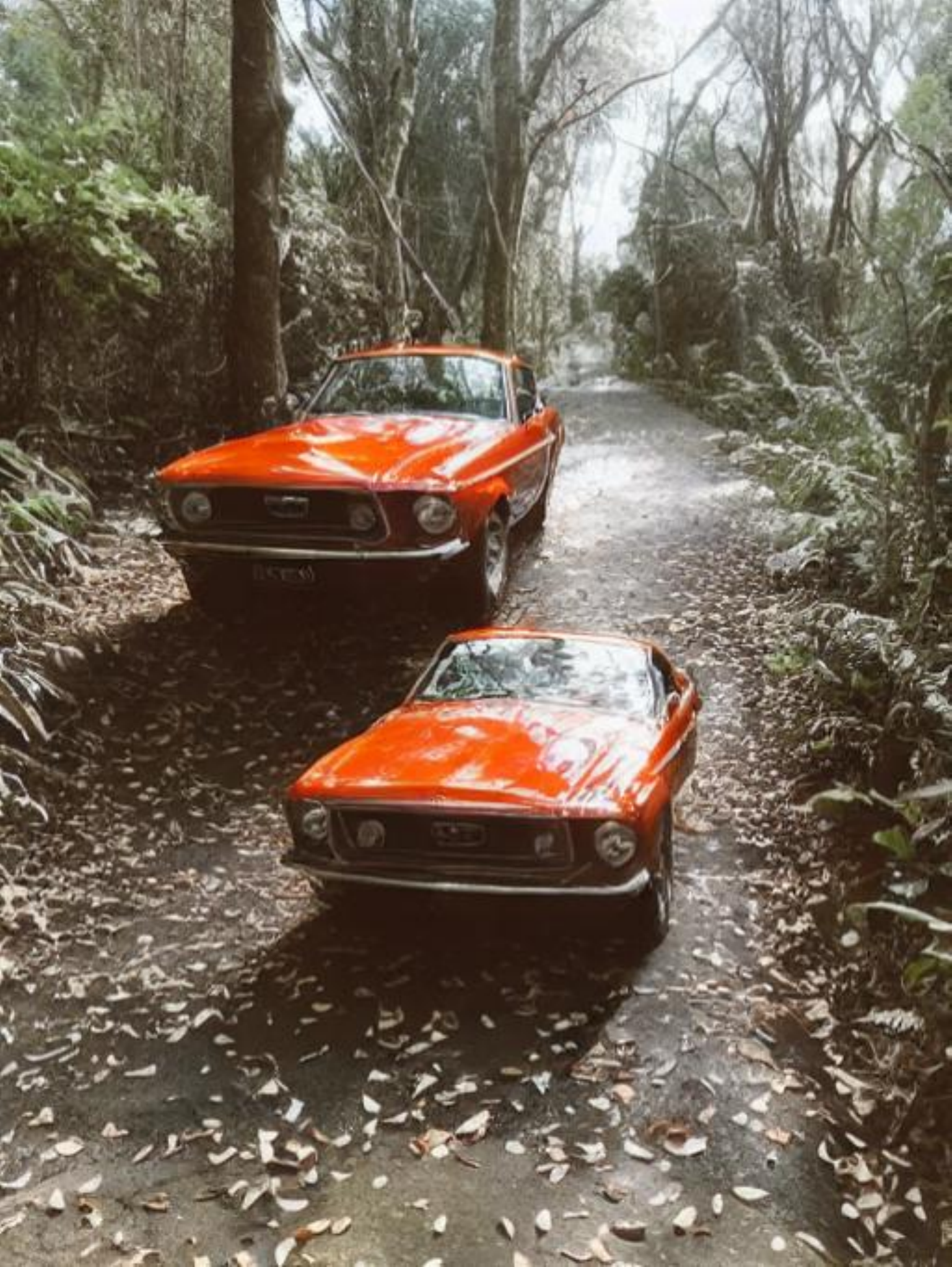}
    \end{subfigure}%
    \begin{subfigure}{.16\linewidth}
        \centering
        \includegraphics[width=\linewidth]{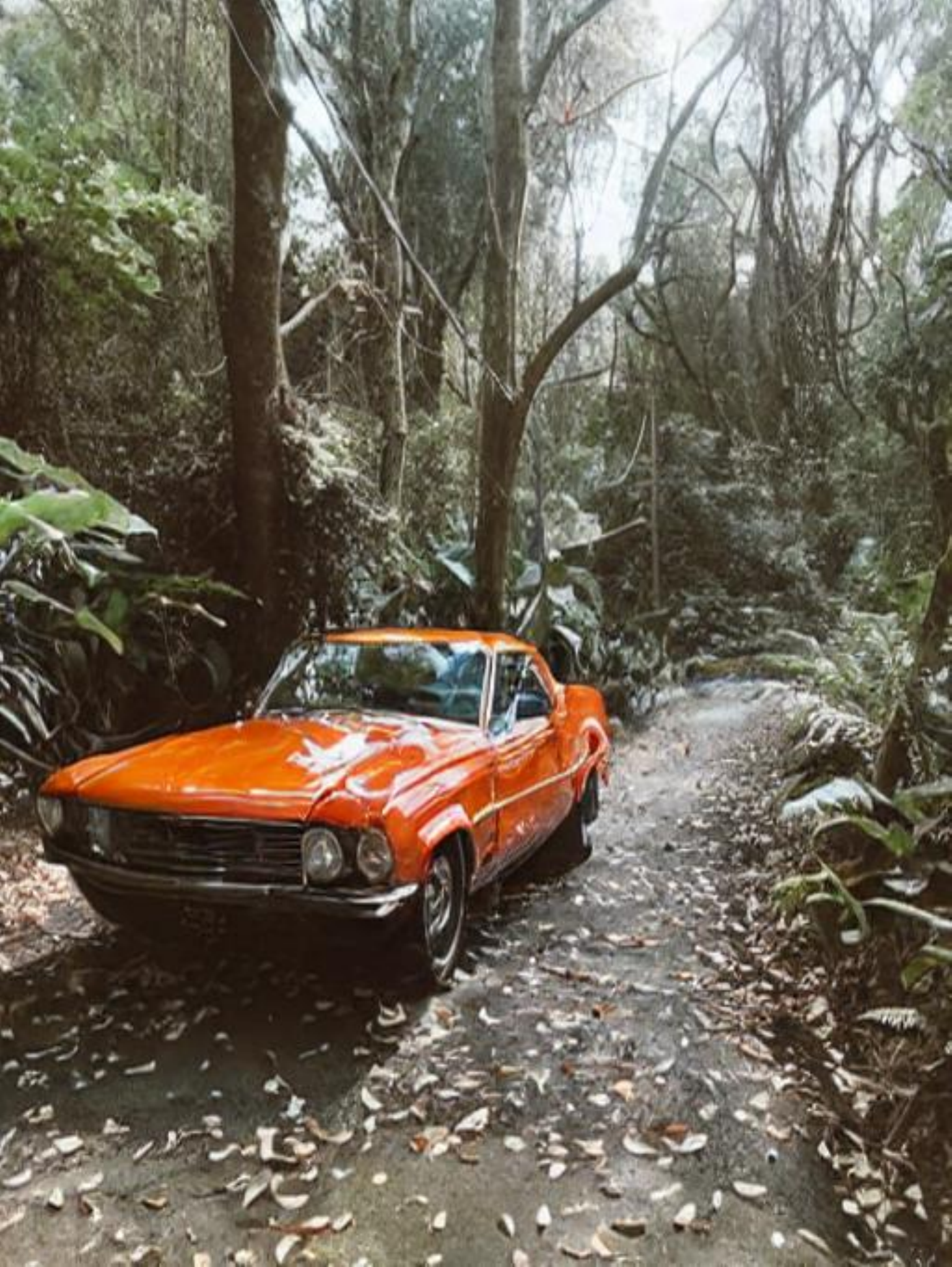}
    \end{subfigure}%
    \begin{subfigure}{.16\linewidth}
        \centering
        \includegraphics[width=\linewidth]{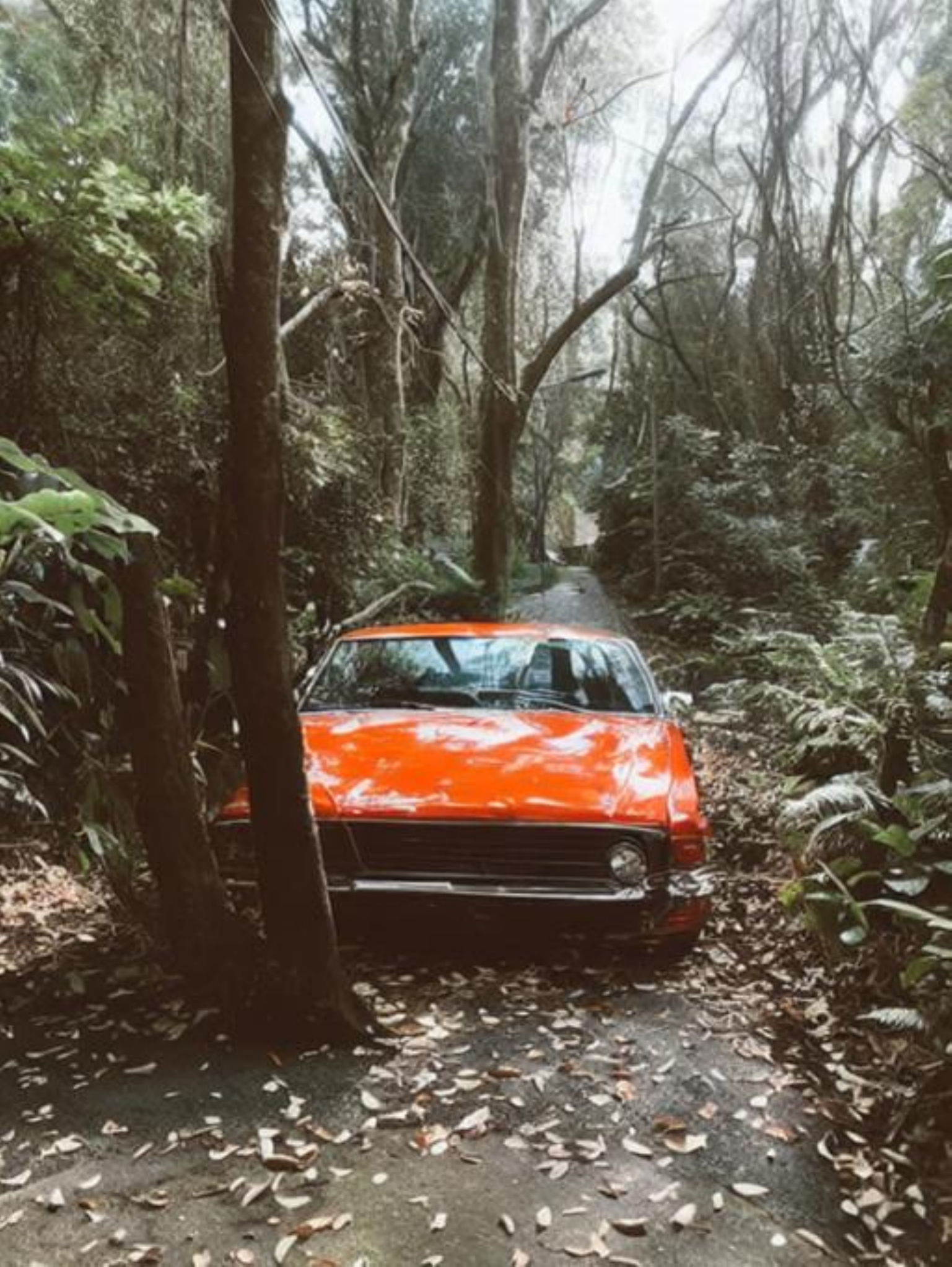}
    \end{subfigure}%
    \hspace{1mm}
    \begin{subfigure}{.16\linewidth}
        \centering
        \includegraphics[width=\linewidth]{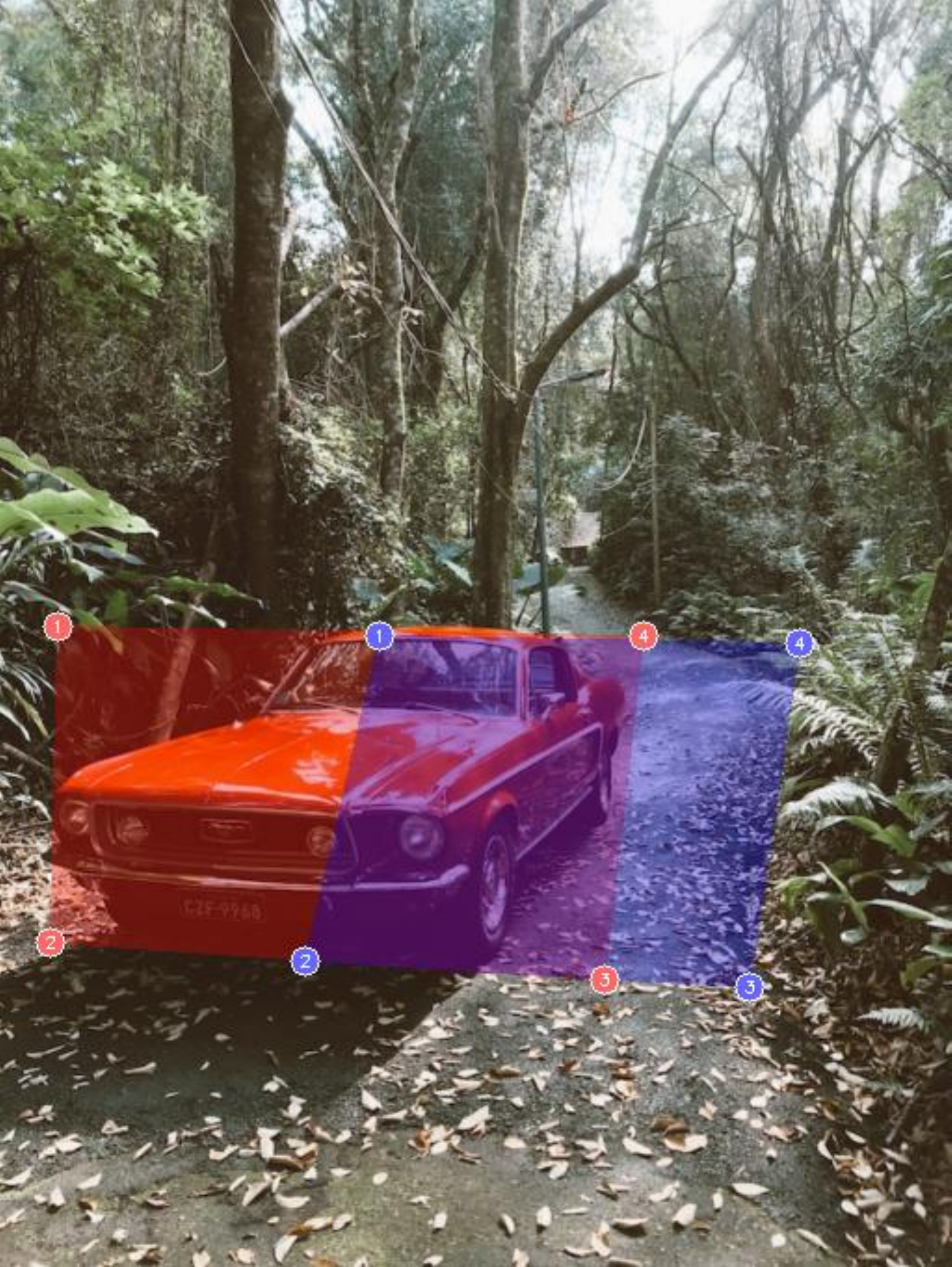}
    \end{subfigure}%
    \begin{subfigure}{.16\linewidth}
        \centering
        \includegraphics[width=\linewidth]{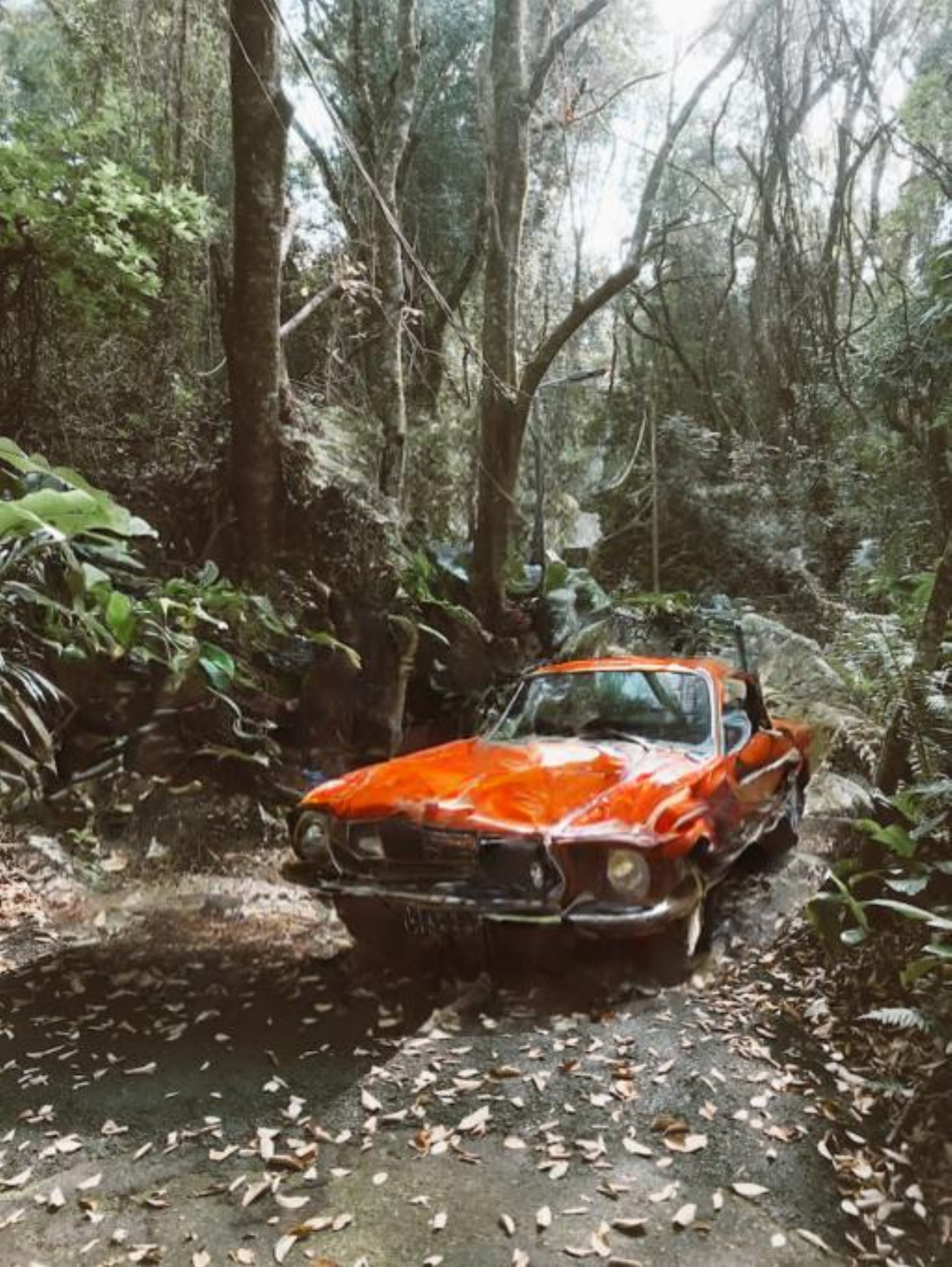}
    \end{subfigure}%

    \begin{subfigure}{.16\linewidth}
        \centering
        \includegraphics[width=\linewidth]{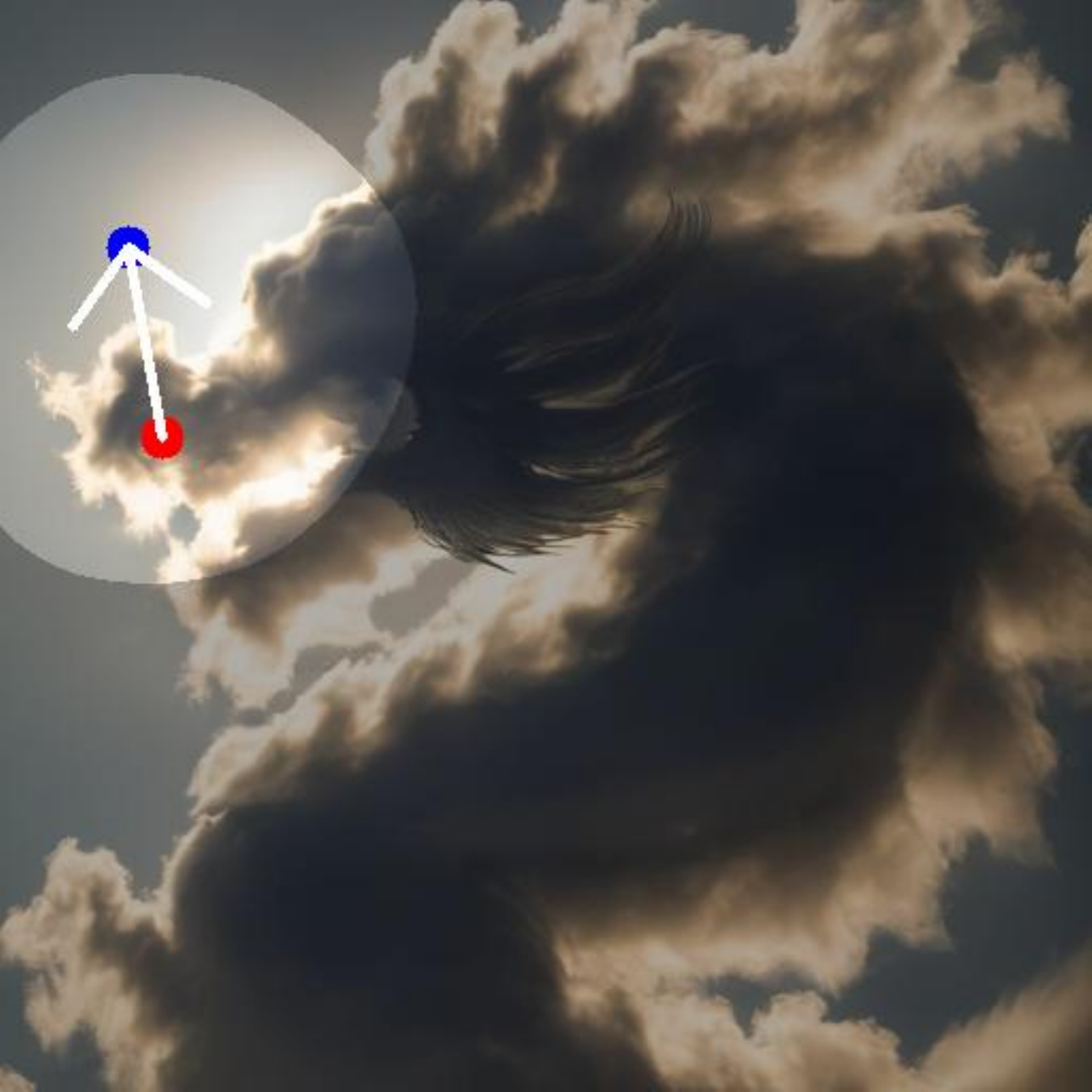}
    \end{subfigure}%
    \begin{subfigure}{.16\linewidth}
        \centering
        \includegraphics[width=\linewidth]{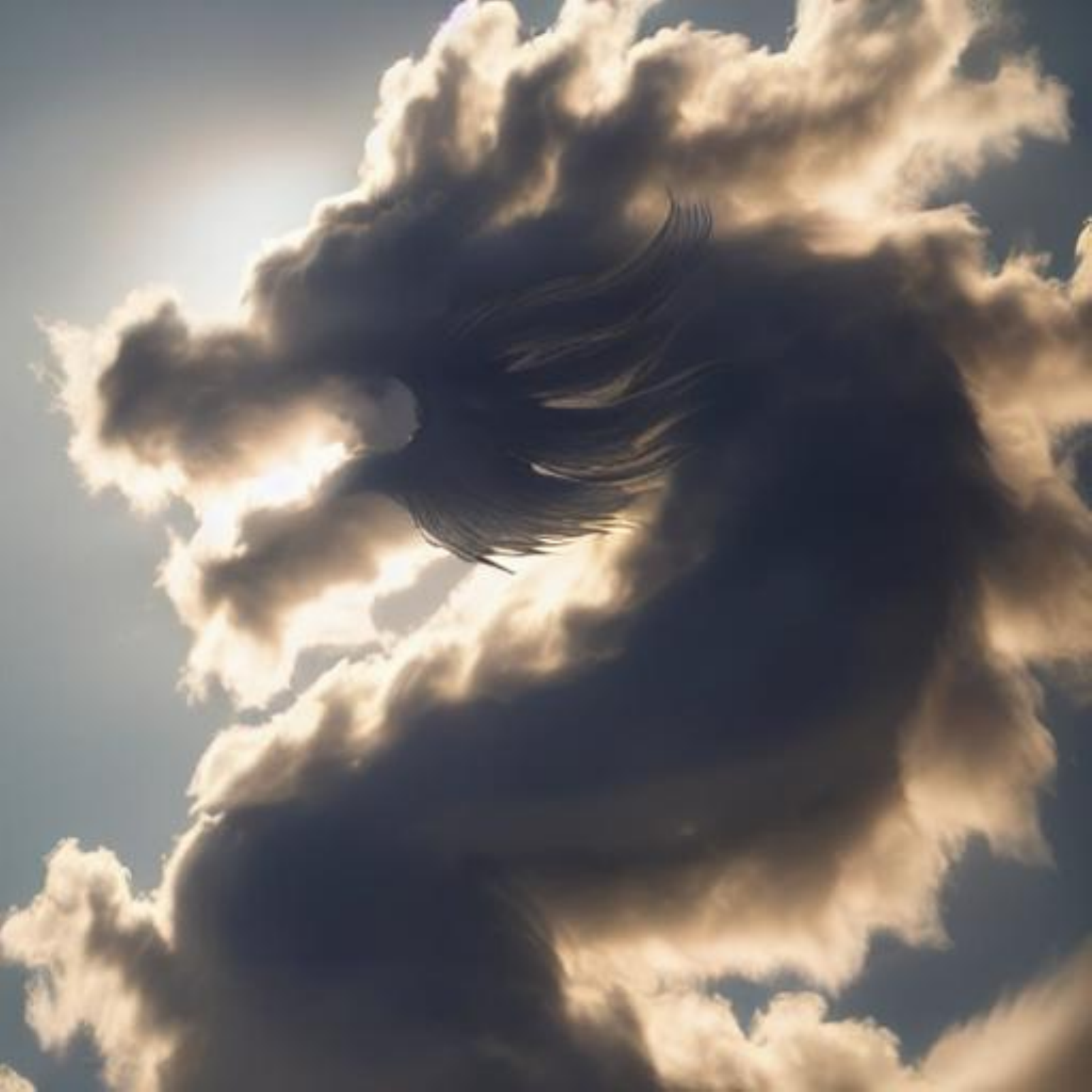}
    \end{subfigure}%
    \begin{subfigure}{.16\linewidth}
        \centering
        \includegraphics[width=\linewidth]{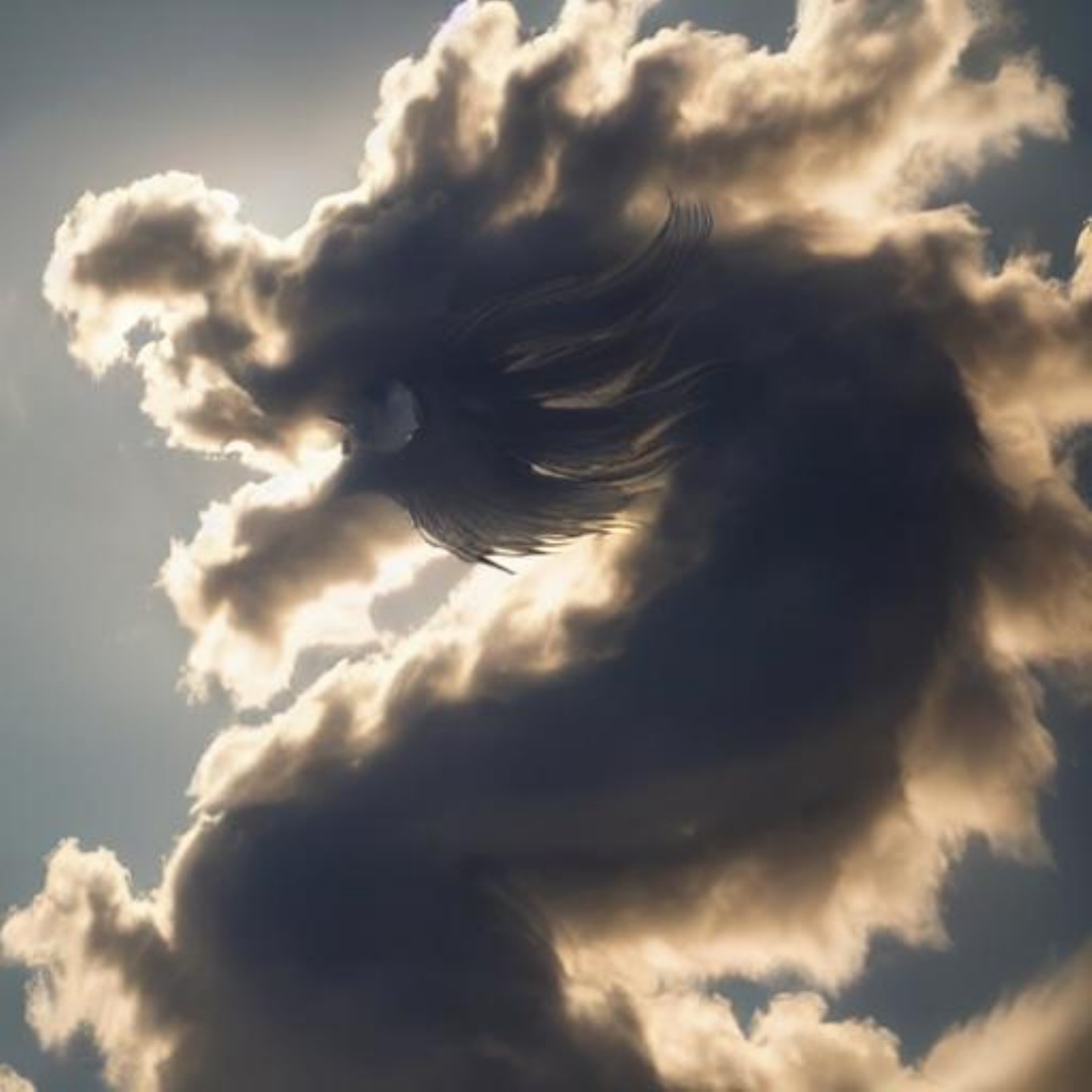}
    \end{subfigure}%
    \begin{subfigure}{.16\linewidth}
        \centering
        \includegraphics[width=\linewidth]{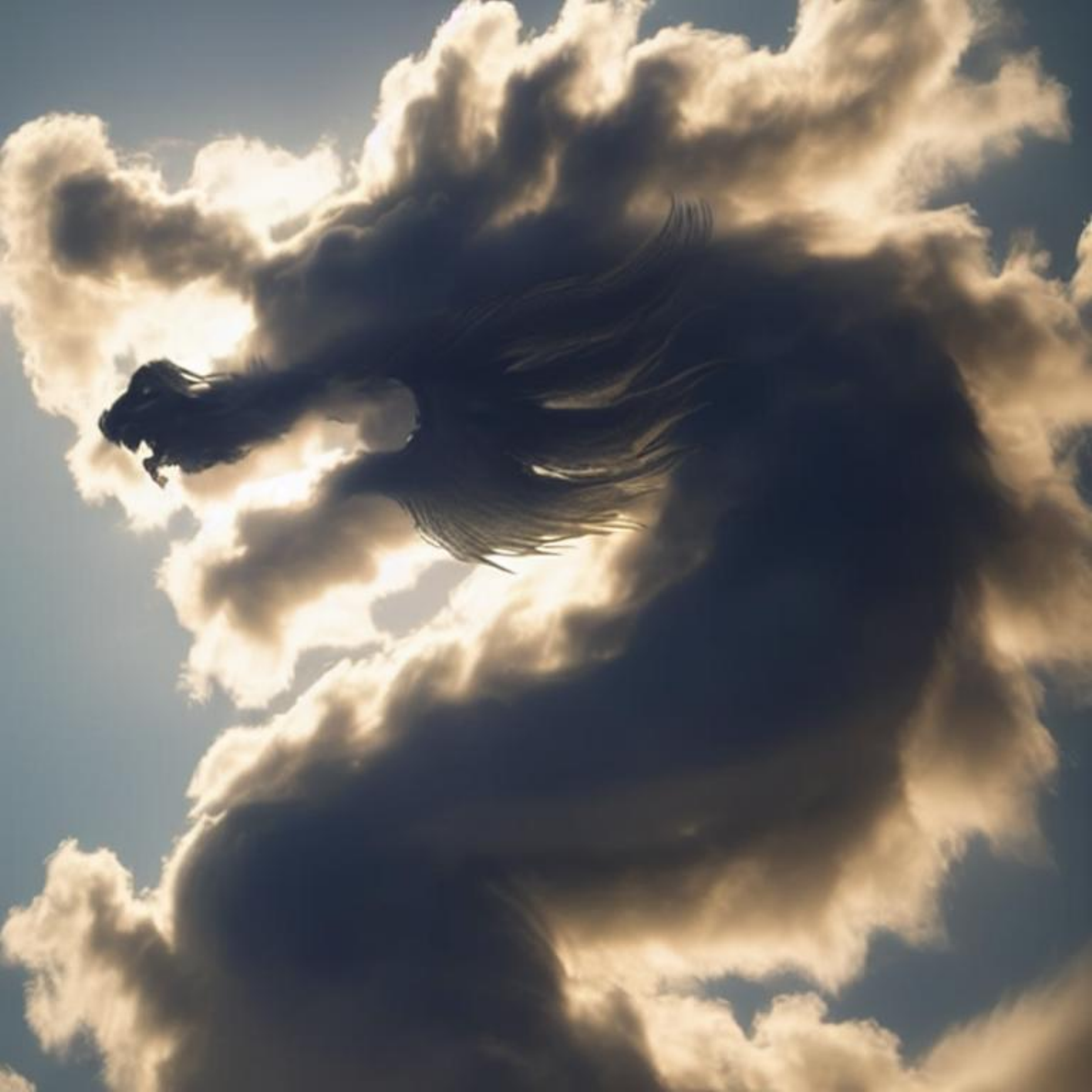}
    \end{subfigure}%
    \hspace{1mm}
    \begin{subfigure}{.16\linewidth}
        \centering
        \includegraphics[width=\linewidth]{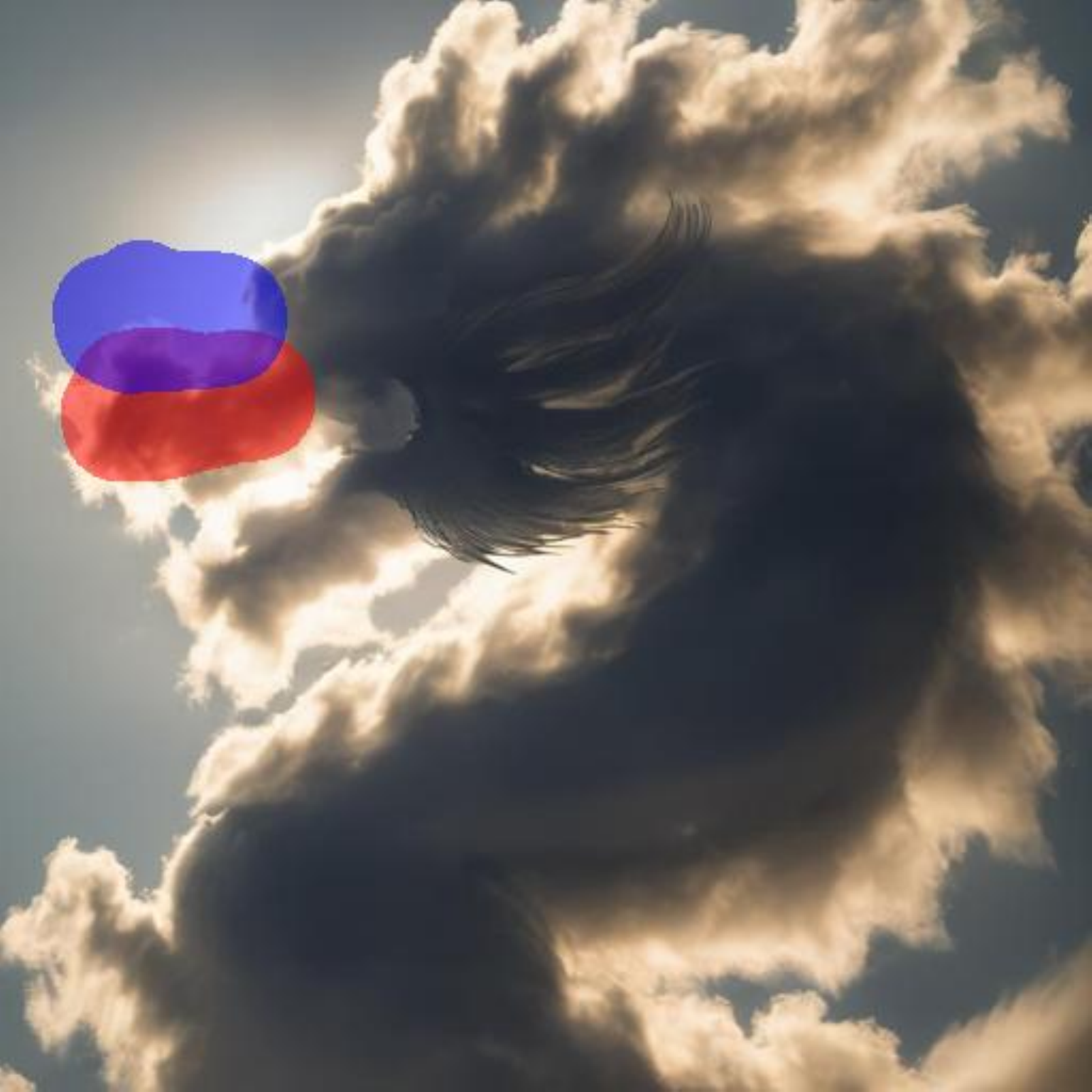}
    \end{subfigure}%
    \begin{subfigure}{.16\linewidth}
        \centering
        \includegraphics[width=\linewidth]{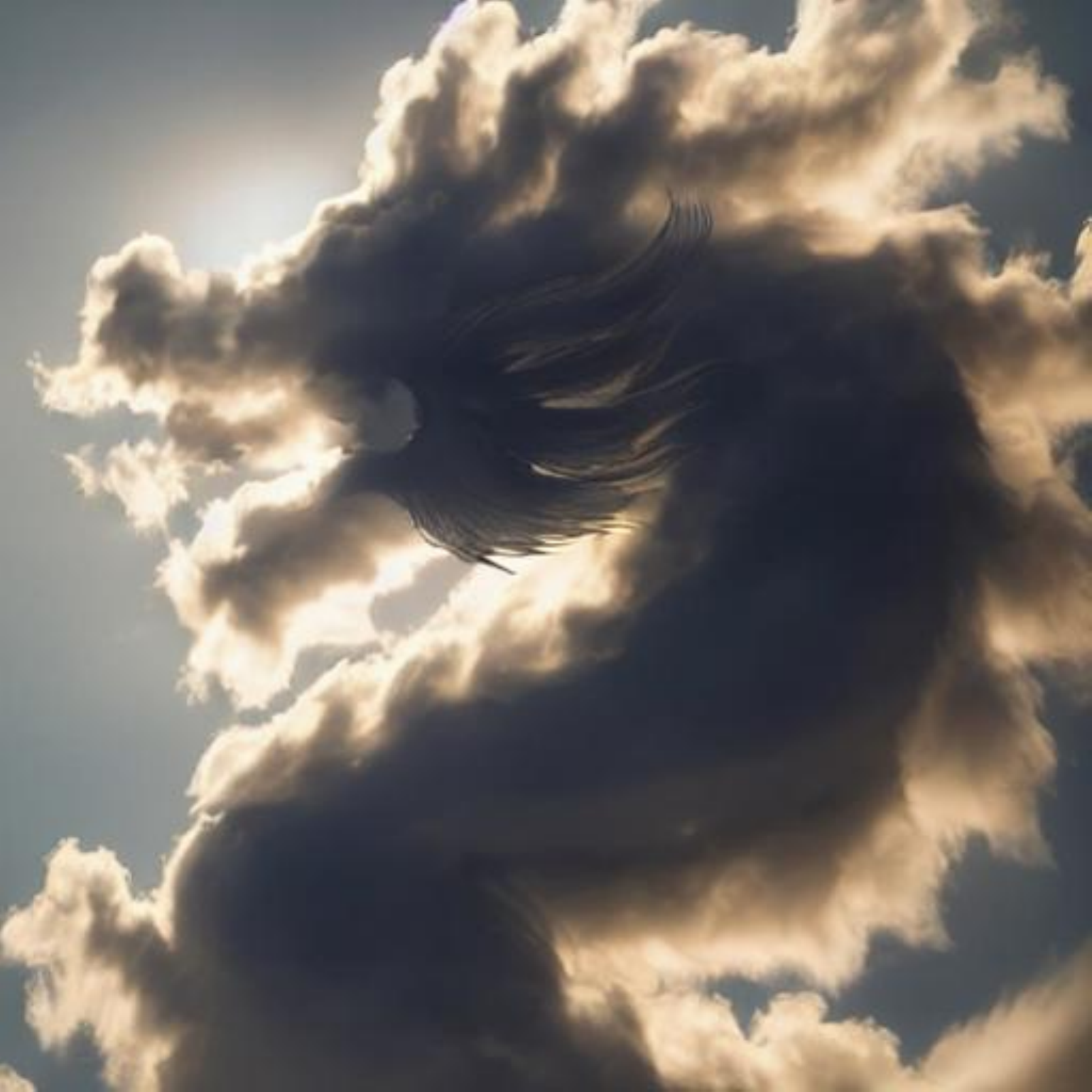}
    \end{subfigure}%

    \begin{subfigure}{.16\linewidth}
        \centering
        \includegraphics[width=\linewidth]{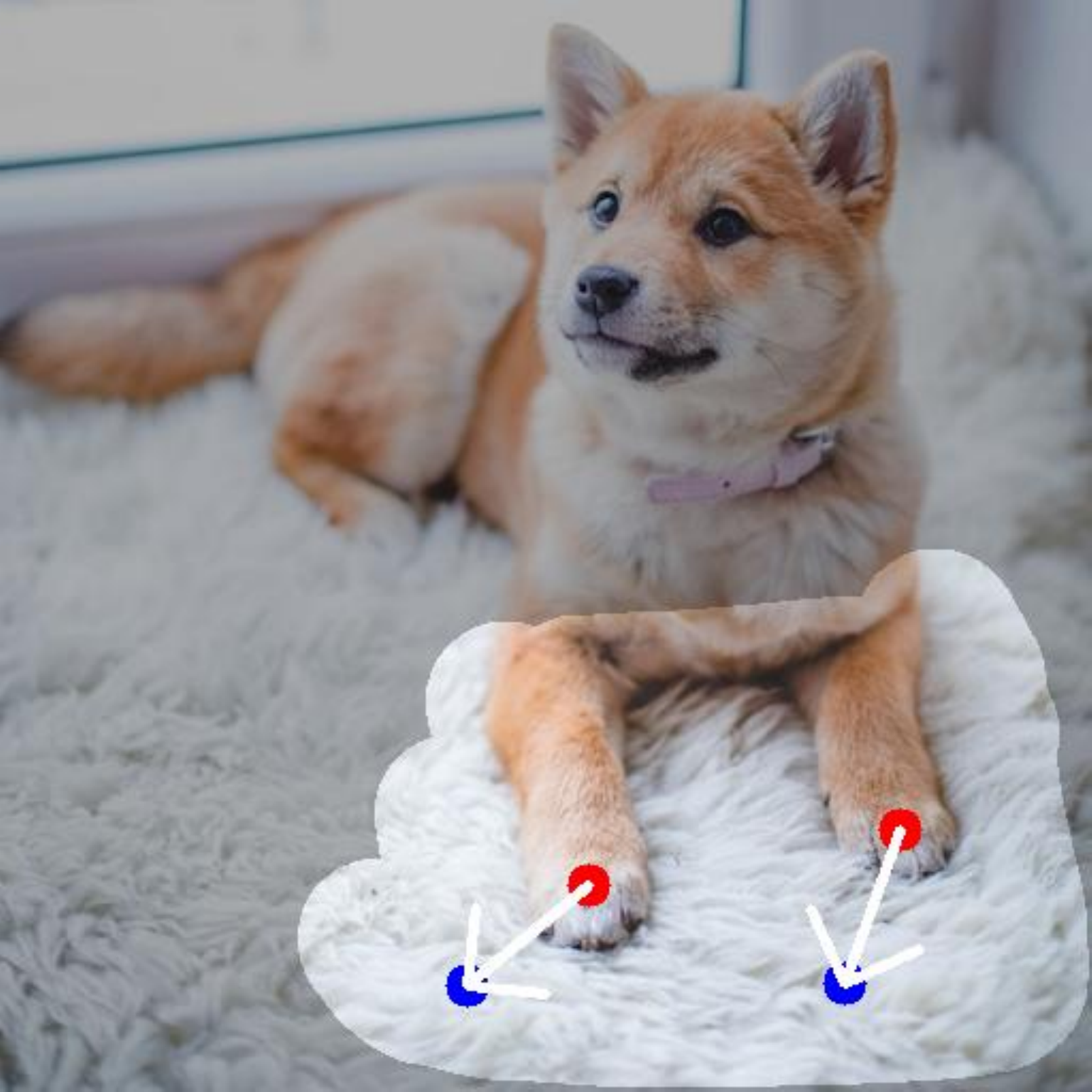}
    \end{subfigure}%
    \begin{subfigure}{.16\linewidth}
        \centering
        \includegraphics[width=\linewidth]{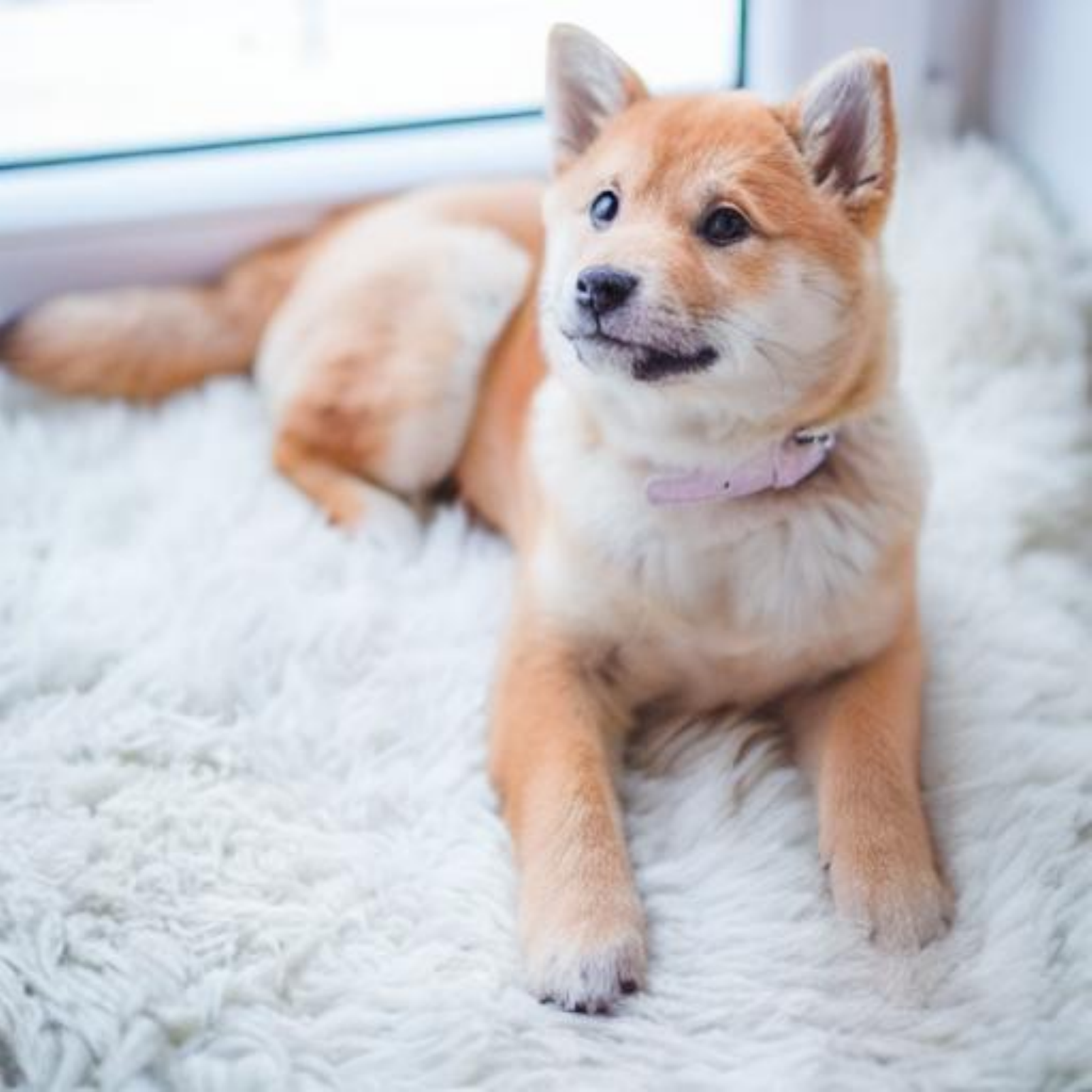}
    \end{subfigure}%
    \begin{subfigure}{.16\linewidth}
        \centering
        \includegraphics[width=\linewidth]{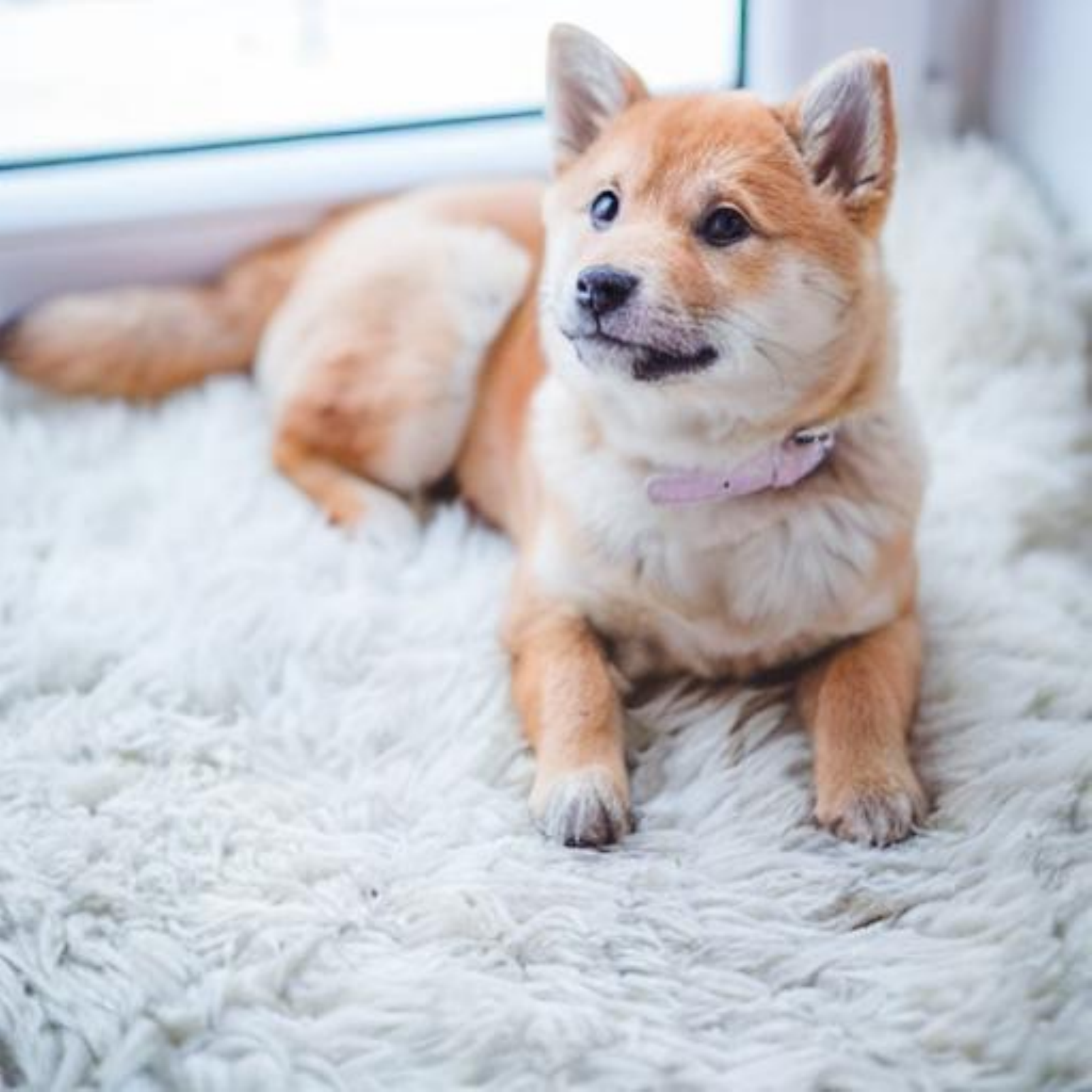}
    \end{subfigure}%
    \begin{subfigure}{.16\linewidth}
        \centering
        \includegraphics[width=\linewidth]{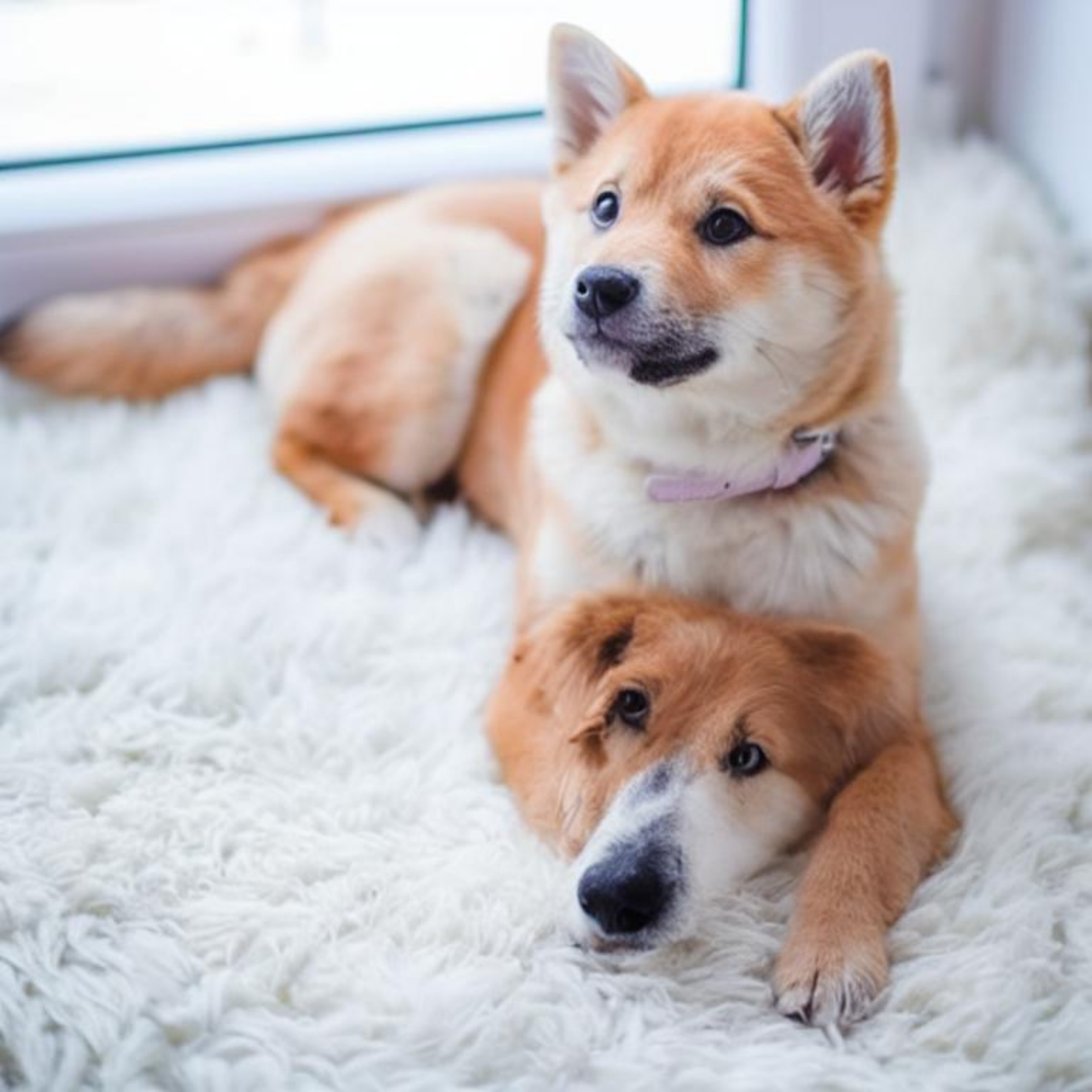}
    \end{subfigure}%
    \hspace{1mm}
    \begin{subfigure}{.16\linewidth}
        \centering
        \includegraphics[width=\linewidth]{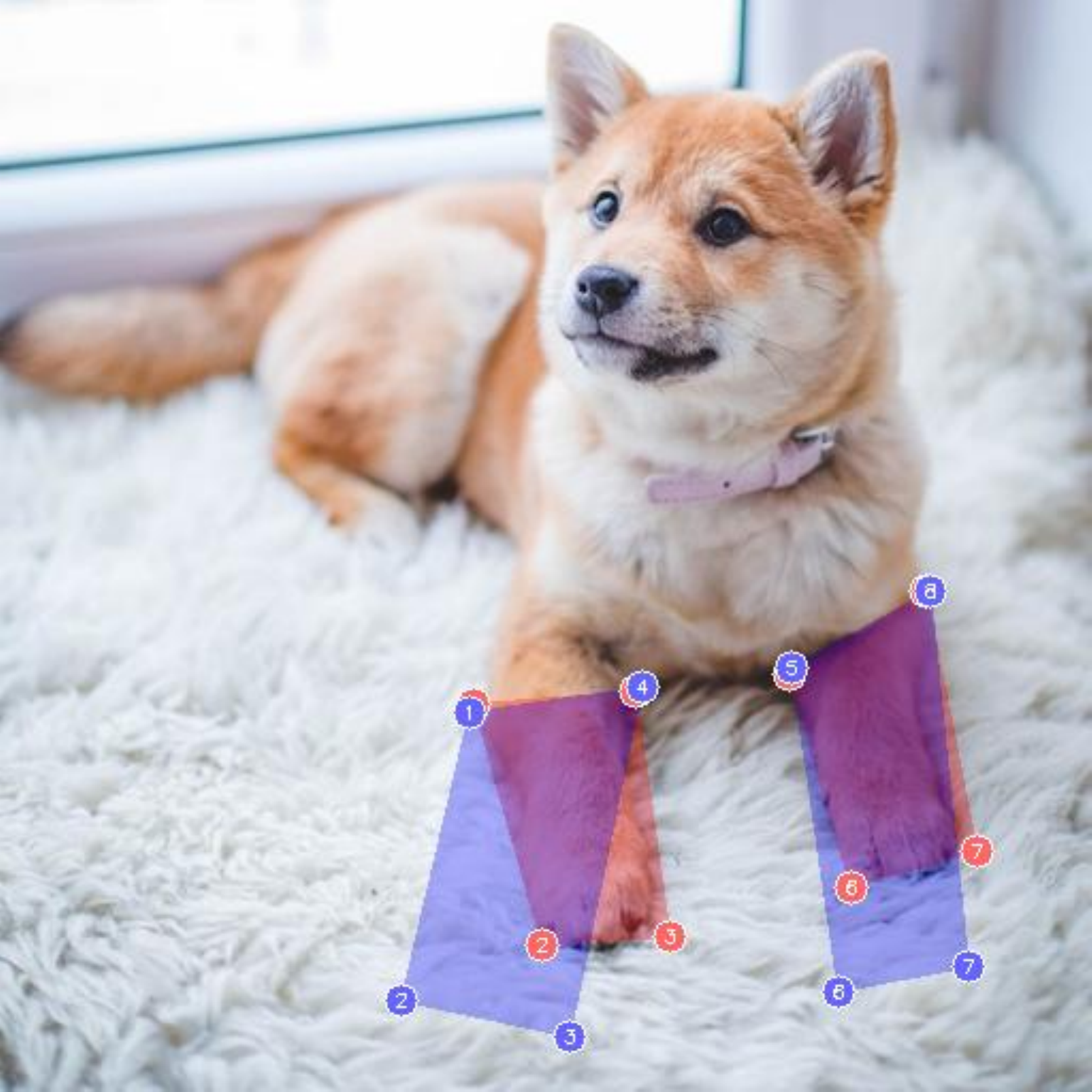}
    \end{subfigure}%
    \begin{subfigure}{.16\linewidth}
        \centering
        \includegraphics[width=\linewidth]{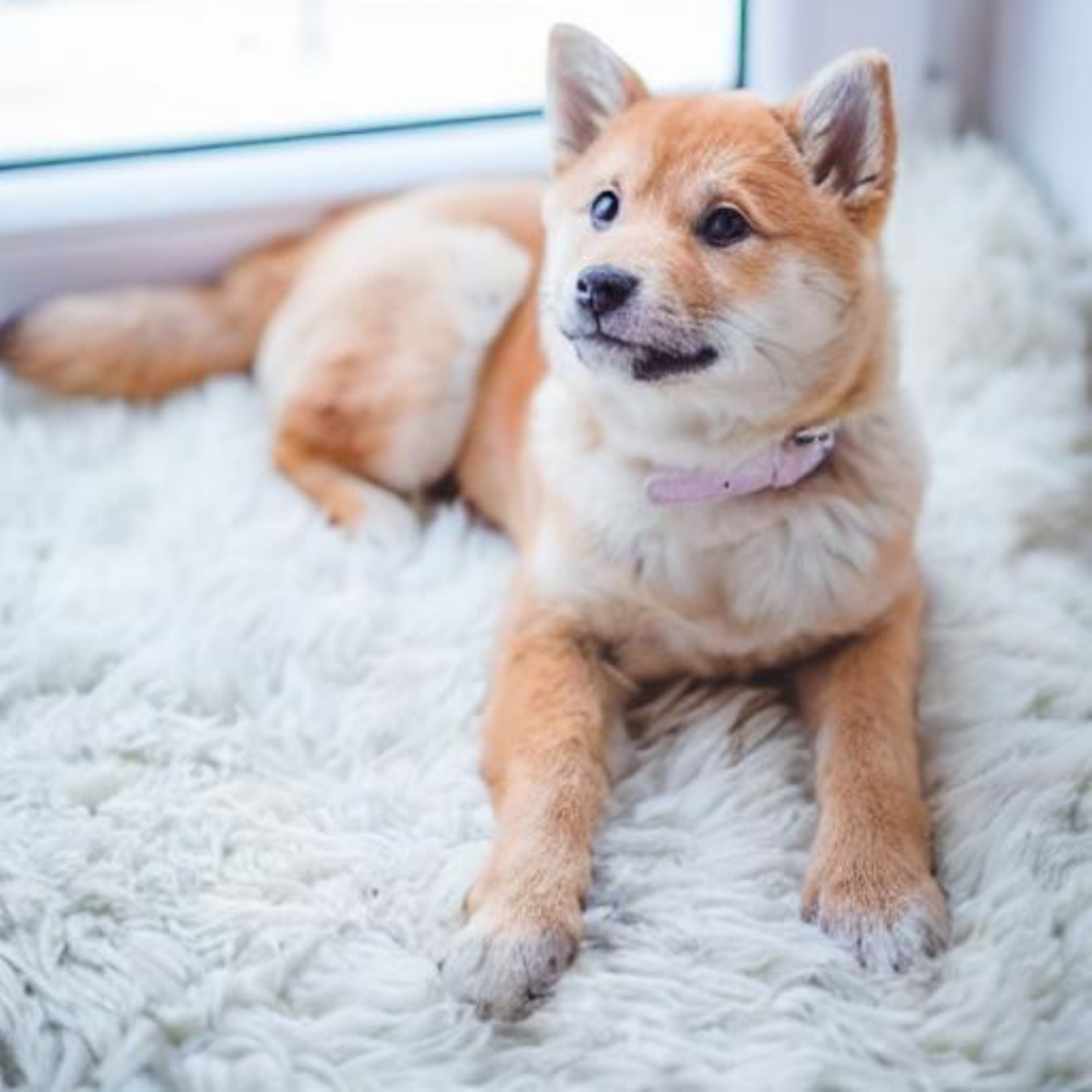}
    \end{subfigure}%

    \begin{subfigure}{.16\linewidth}
        \centering
        \includegraphics[width=\linewidth]{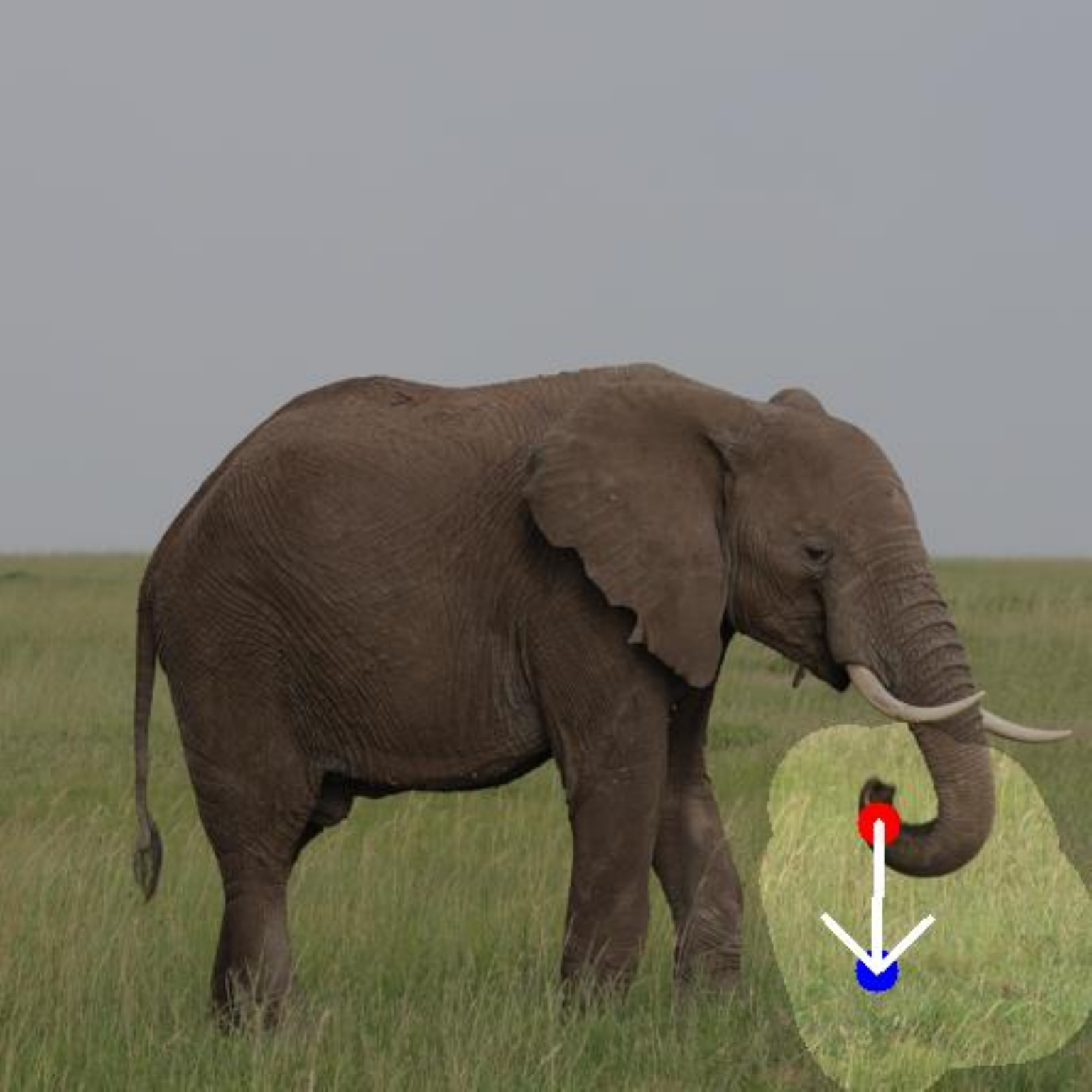}
    \end{subfigure}%
    \begin{subfigure}{.16\linewidth}
        \centering
        \includegraphics[width=\linewidth]{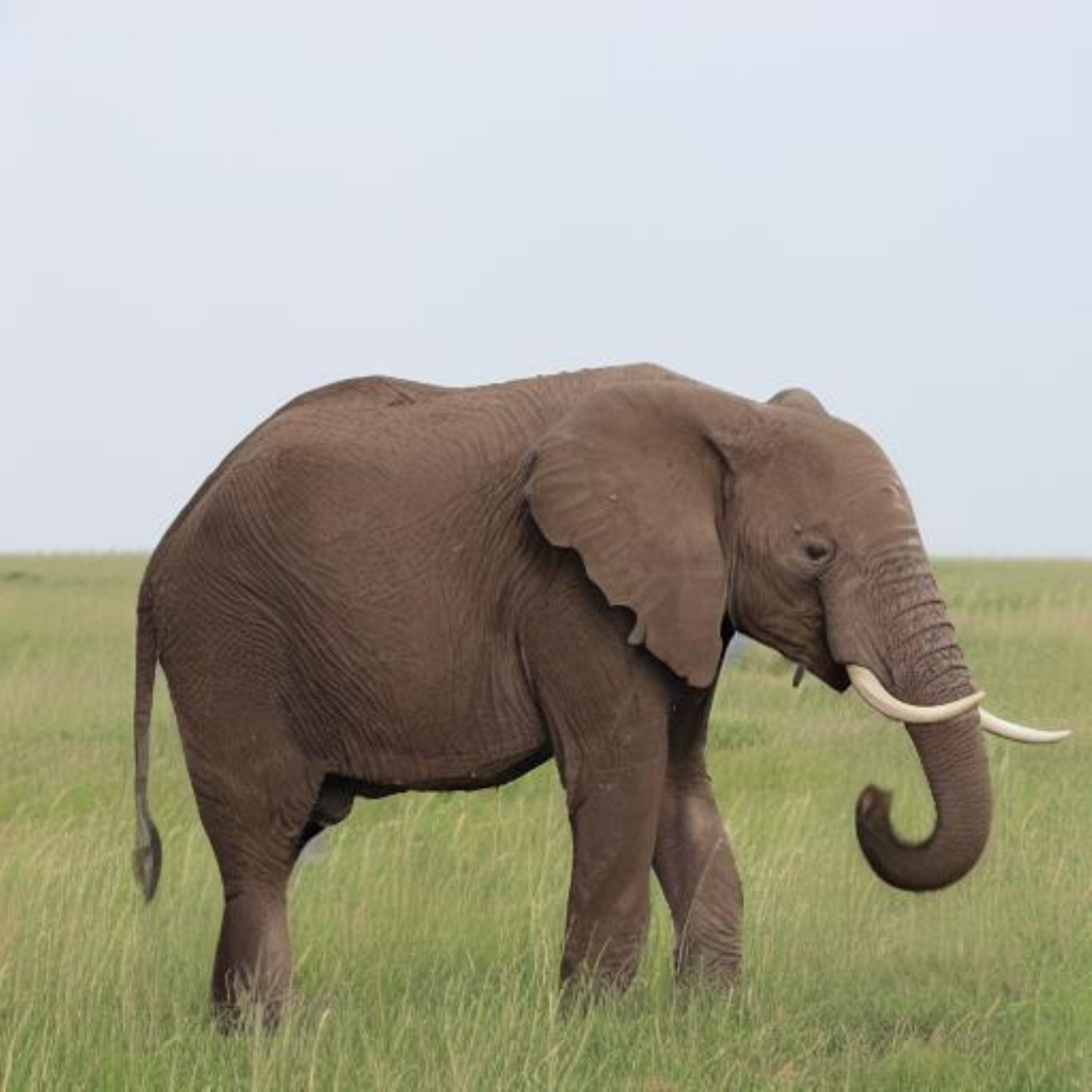}
    \end{subfigure}%
    \begin{subfigure}{.16\linewidth}
        \centering
        \includegraphics[width=\linewidth]{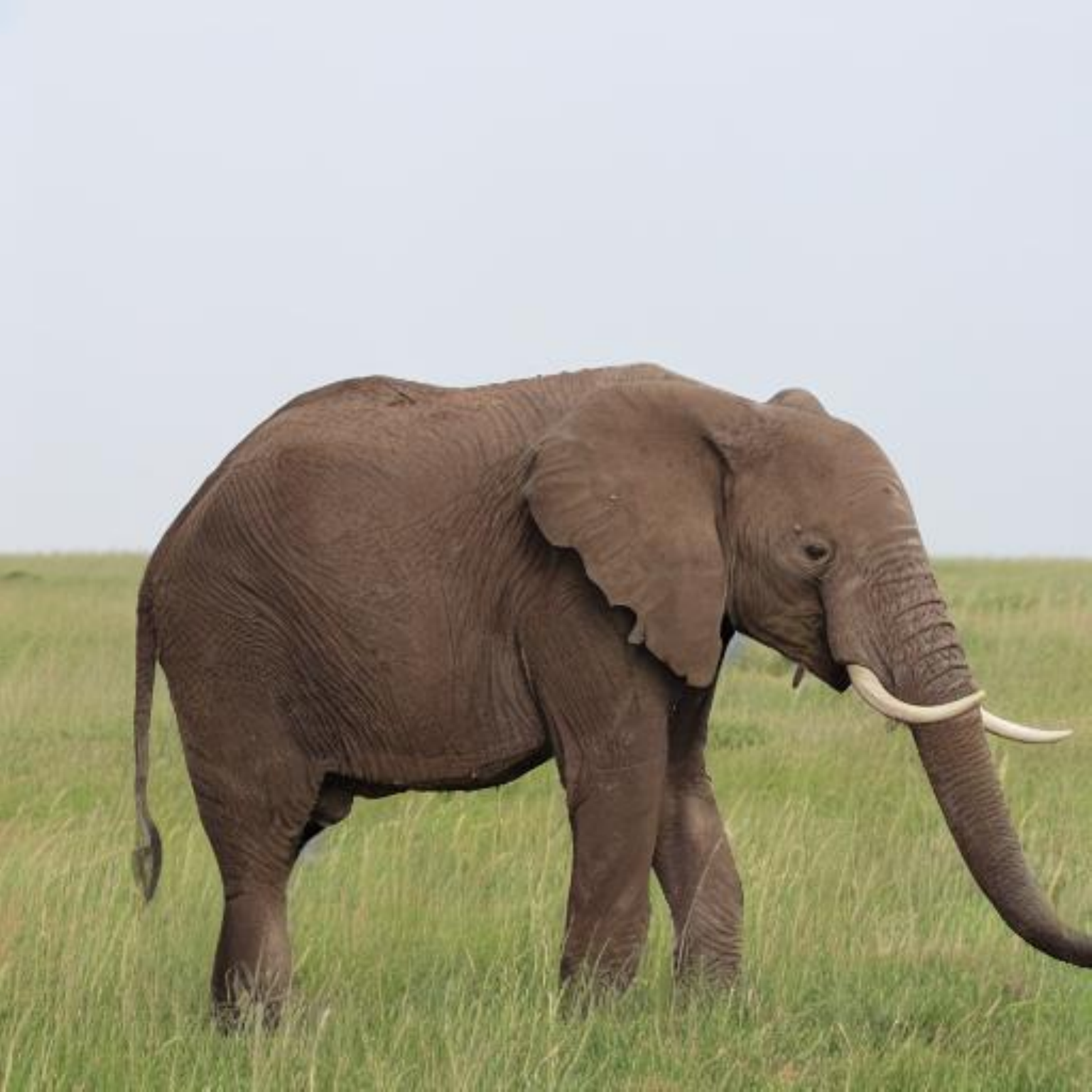}
    \end{subfigure}%
    \begin{subfigure}{.16\linewidth}
        \centering
        \includegraphics[width=\linewidth]{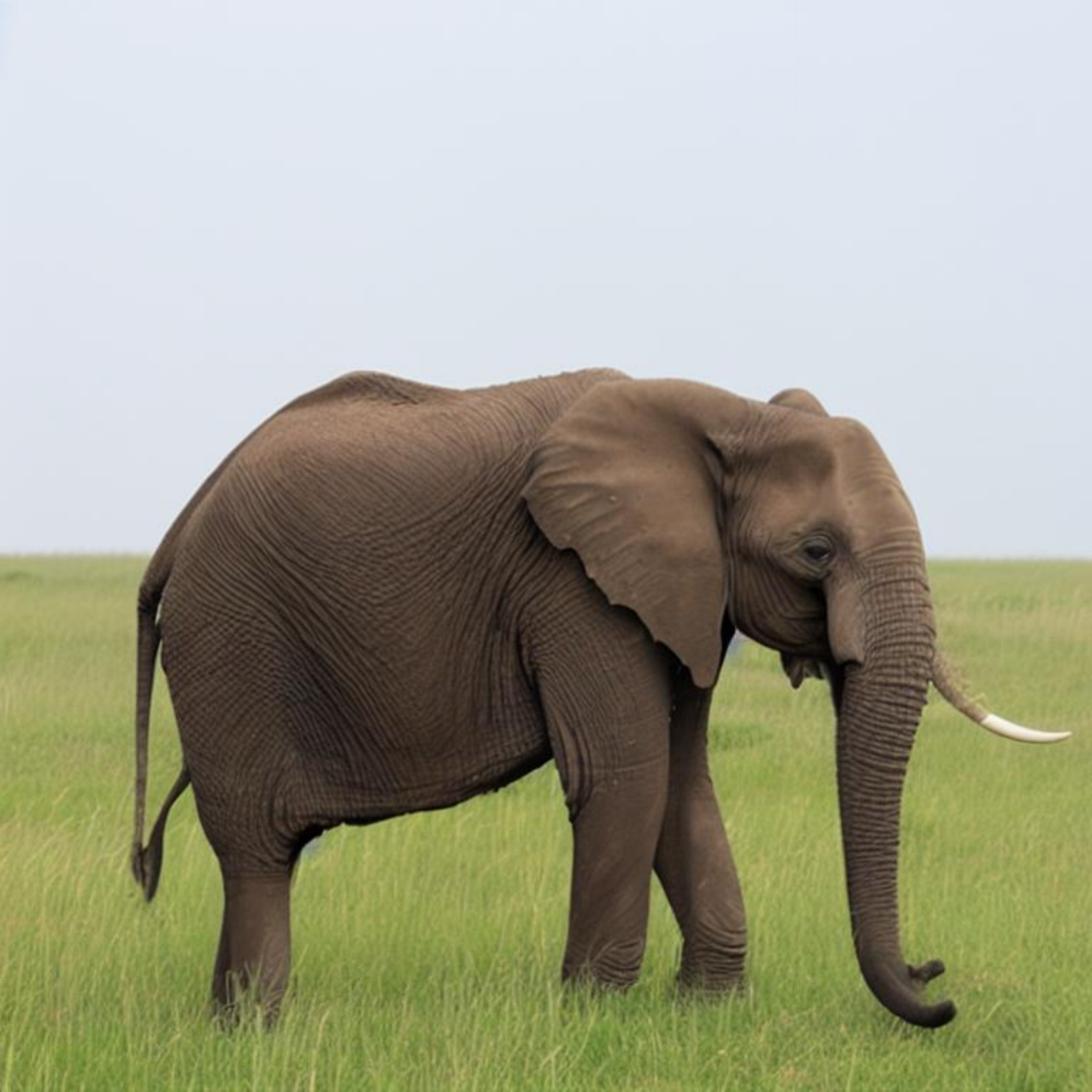}
    \end{subfigure}%
    \hspace{1mm}
    \begin{subfigure}{.16\linewidth}
        \centering
        \includegraphics[width=\linewidth]{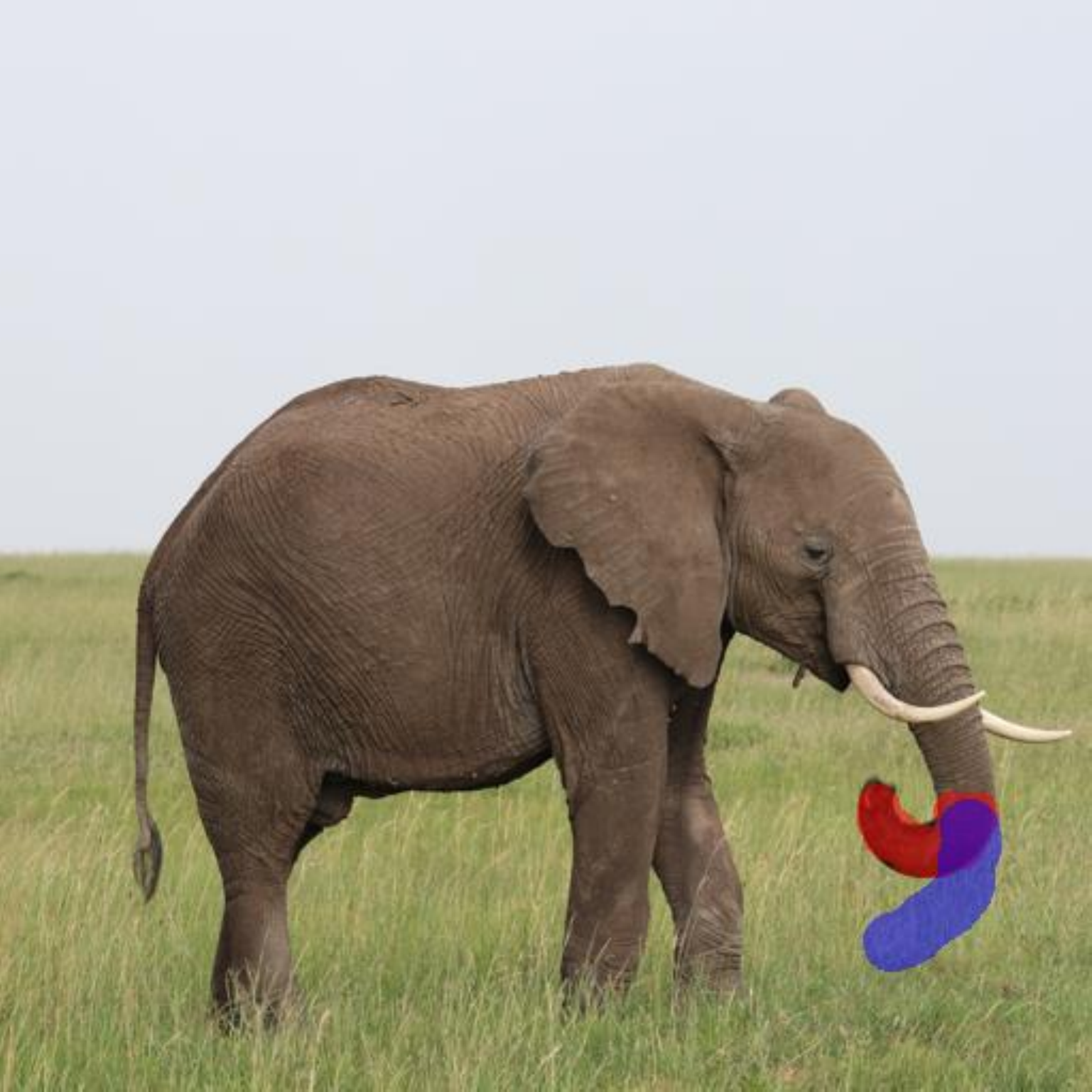}
    \end{subfigure}%
    \begin{subfigure}{.16\linewidth}
        \centering
        \includegraphics[width=\linewidth]{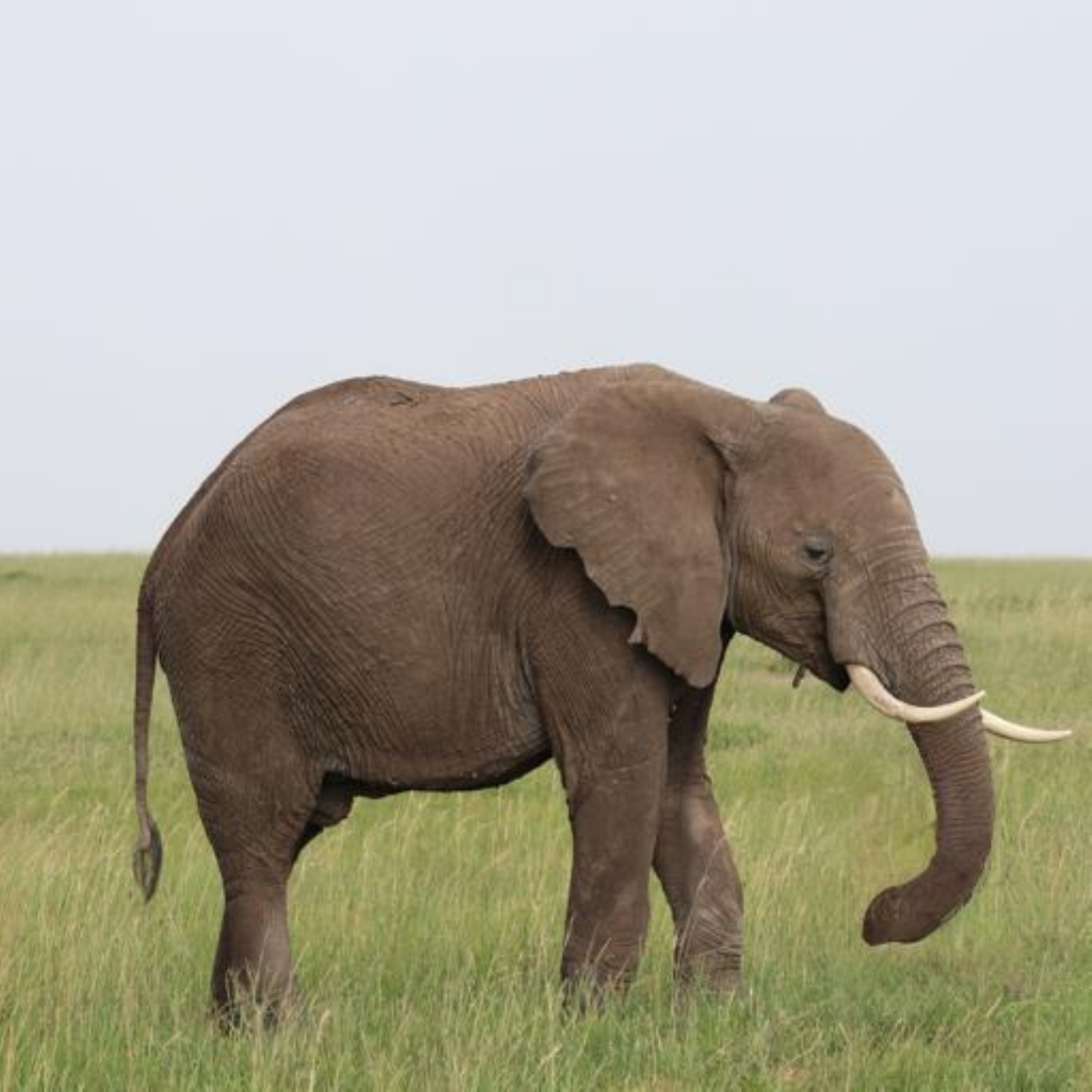}
    \end{subfigure}%

     \begin{subfigure}{.16\linewidth}
        \centering
        \includegraphics[width=\linewidth]{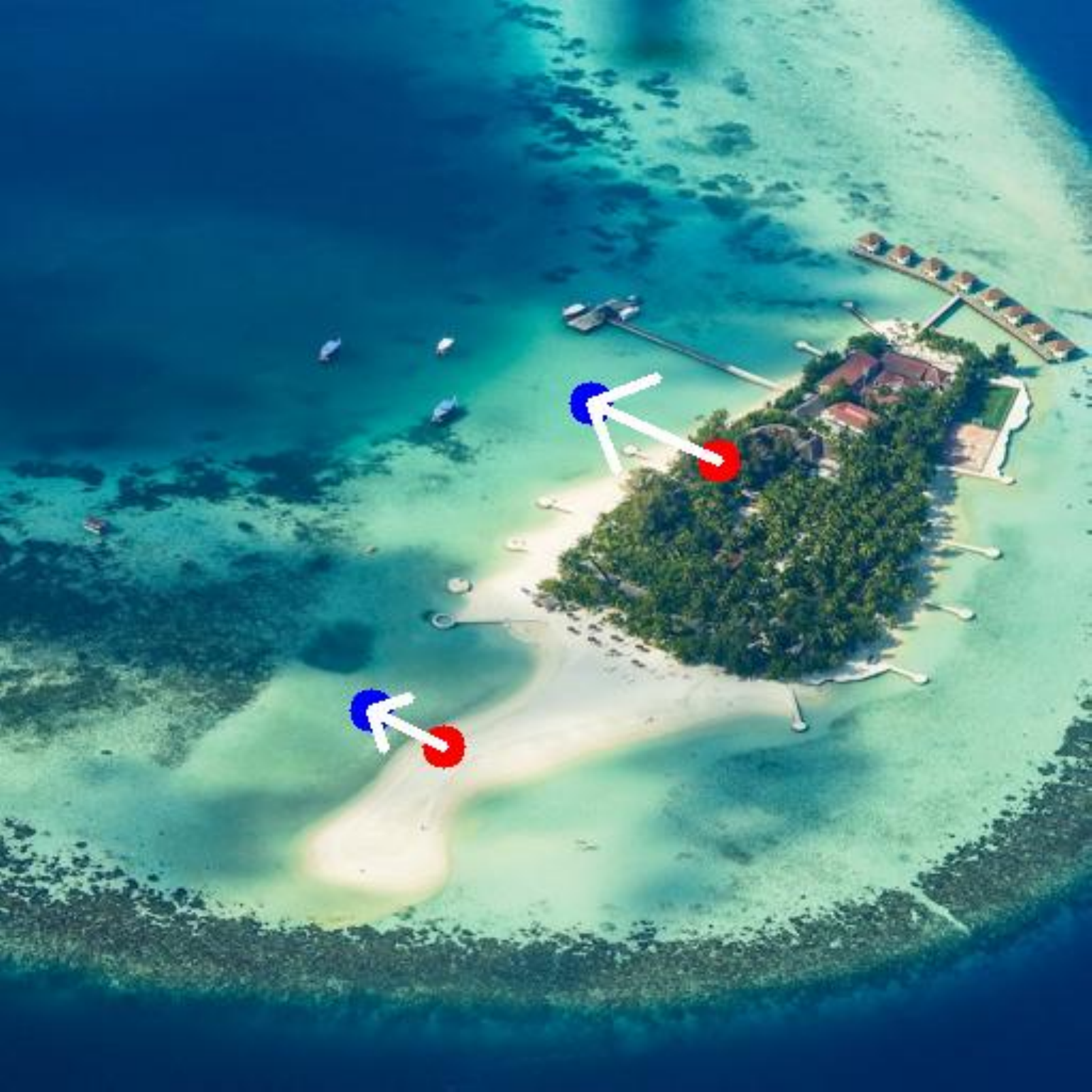}
    \end{subfigure}%
    \begin{subfigure}{.16\linewidth}
        \centering
        \includegraphics[width=\linewidth]{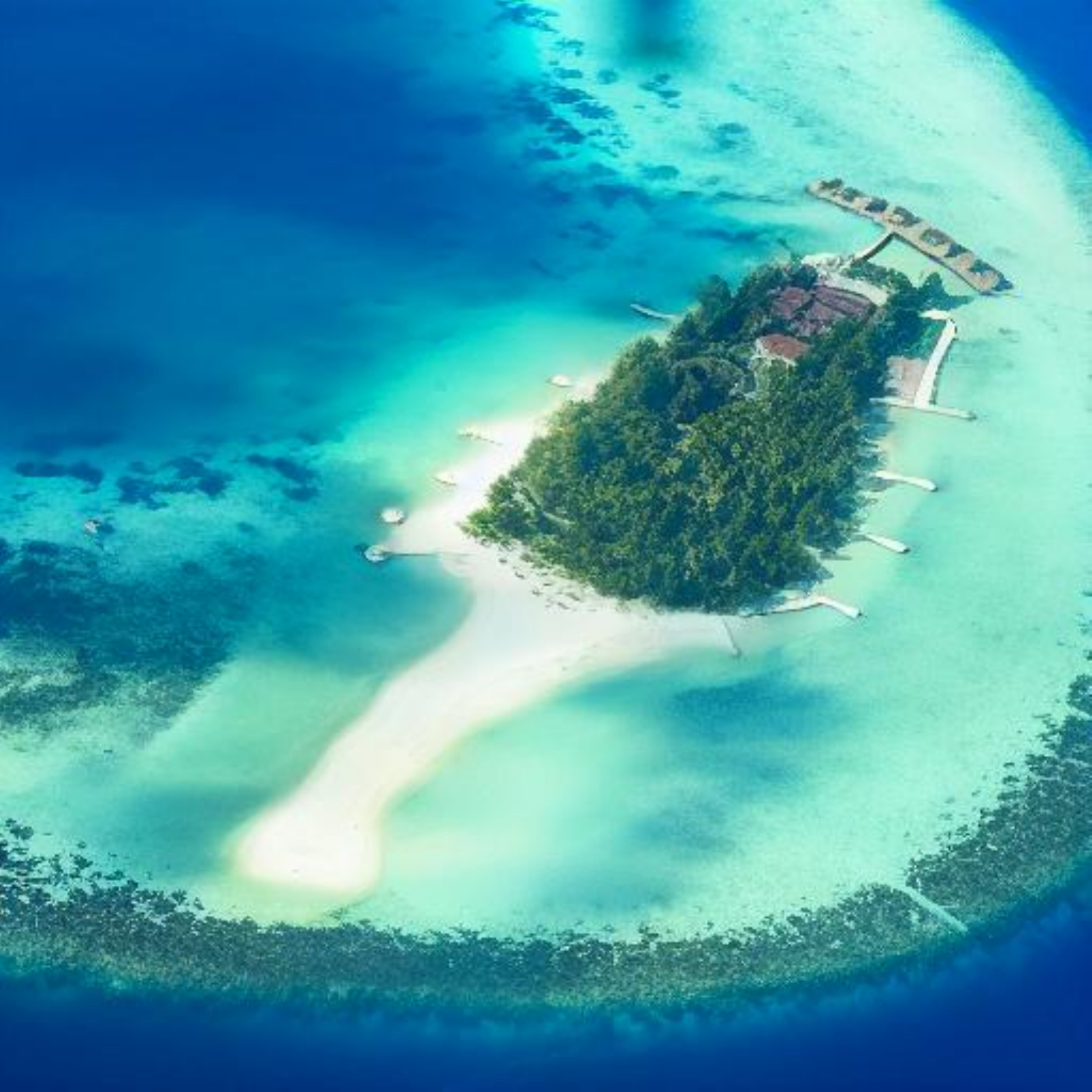}
    \end{subfigure}%
    \begin{subfigure}{.16\linewidth}
        \centering
        \includegraphics[width=\linewidth]{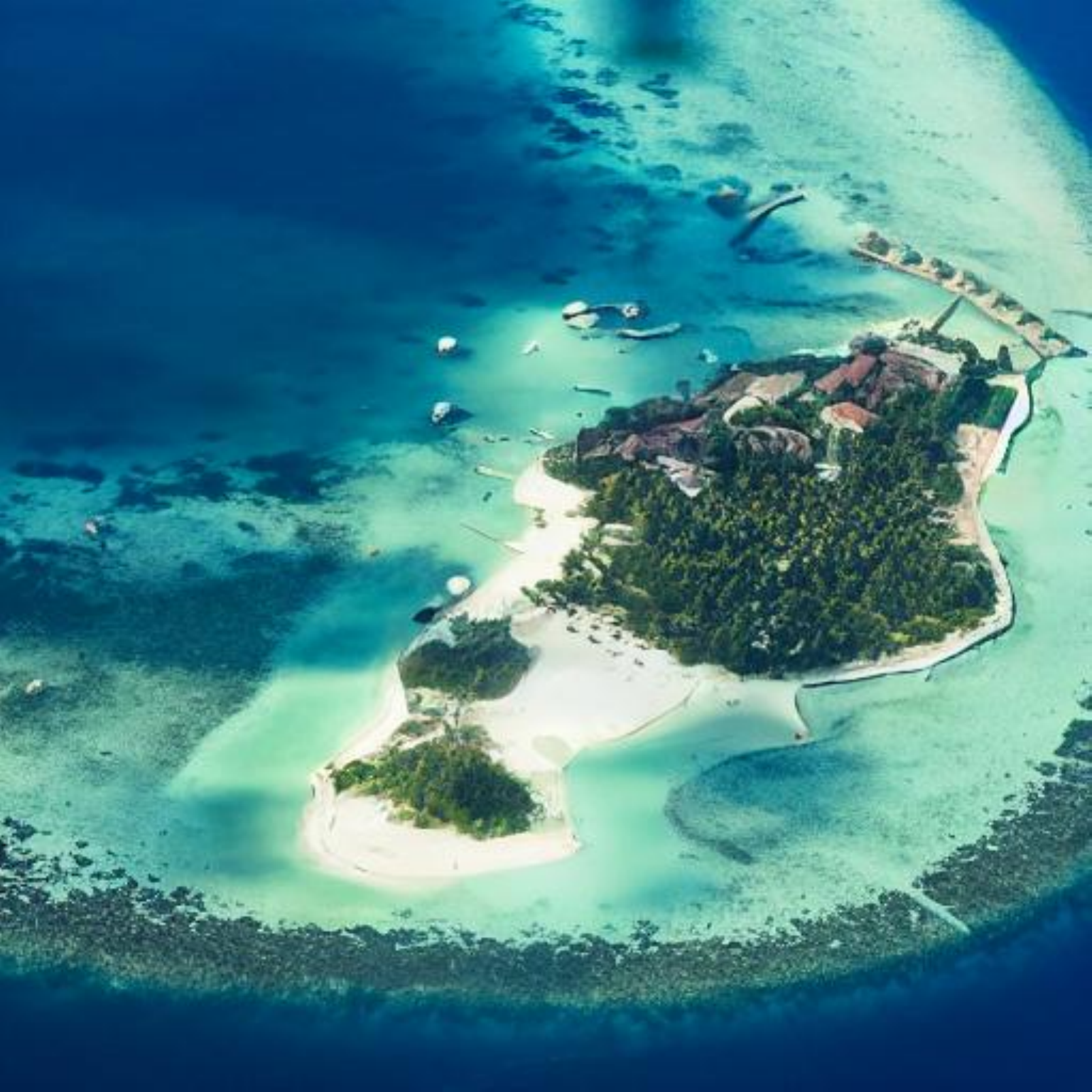}
    \end{subfigure}%
    \begin{subfigure}{.16\linewidth}
        \centering
        \includegraphics[width=\linewidth]{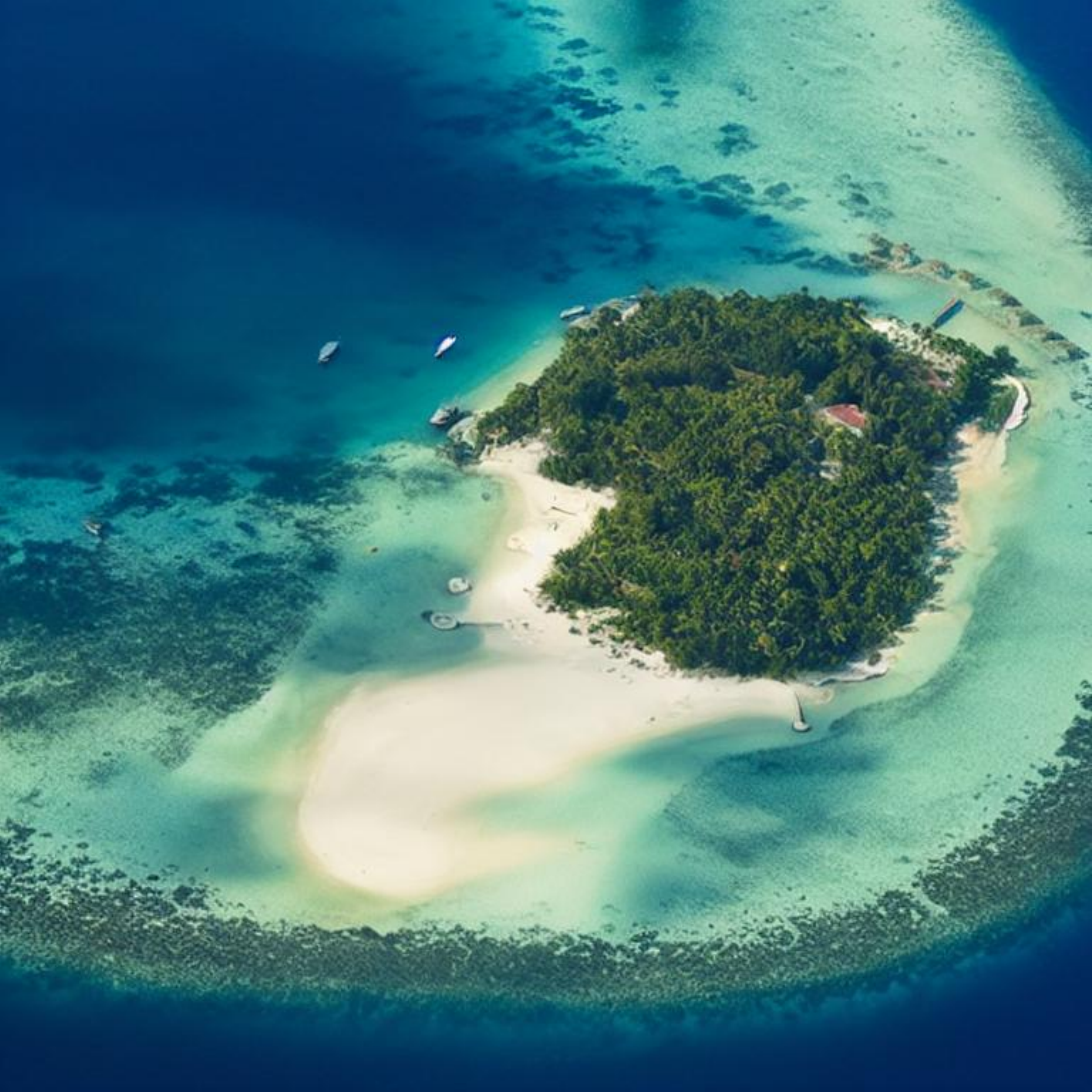}
    \end{subfigure}%
    \hspace{1mm}
    \begin{subfigure}{.16\linewidth}
        \centering
        \includegraphics[width=\linewidth]{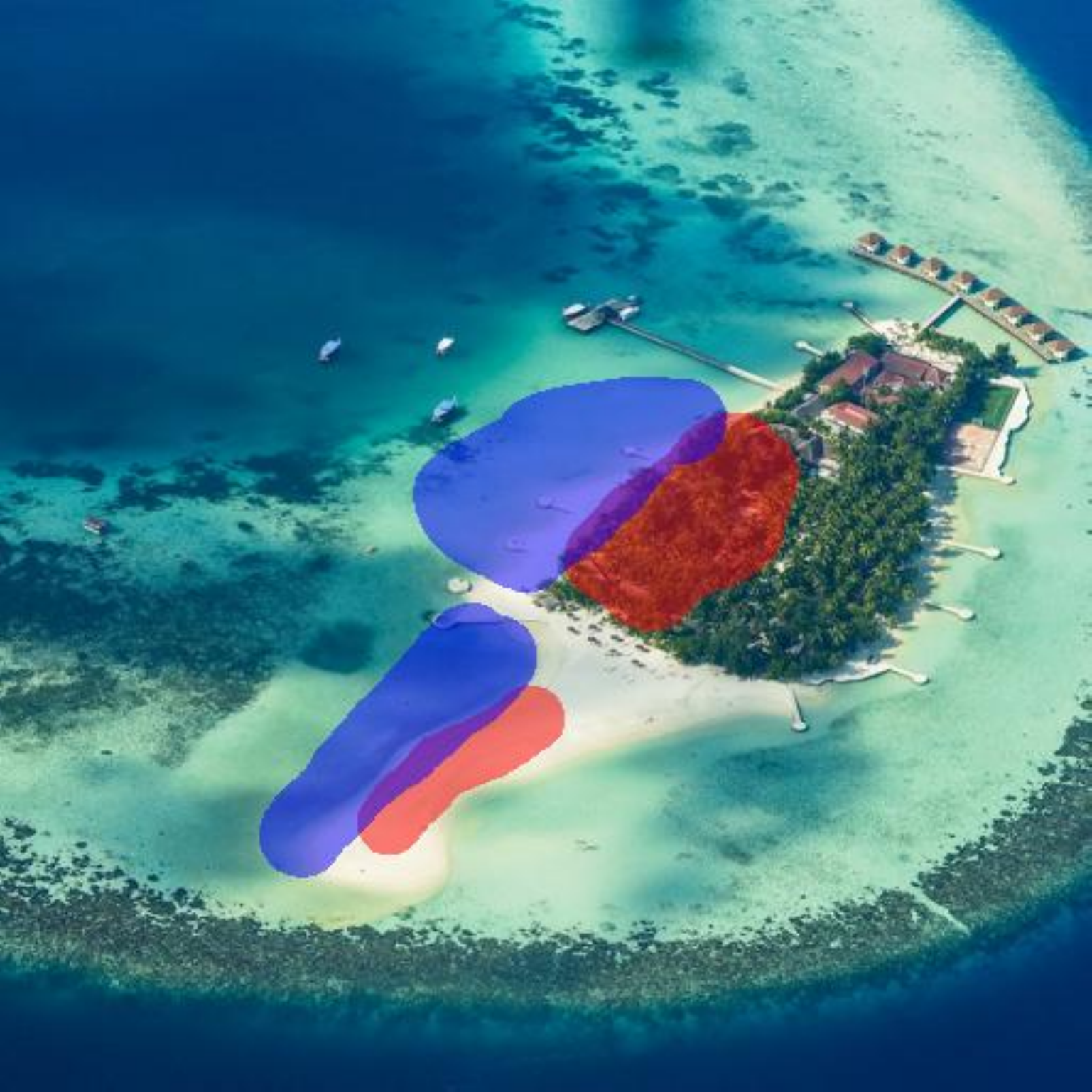}
    \end{subfigure}%
    \begin{subfigure}{.16\linewidth}
        \centering
        \includegraphics[width=\linewidth]{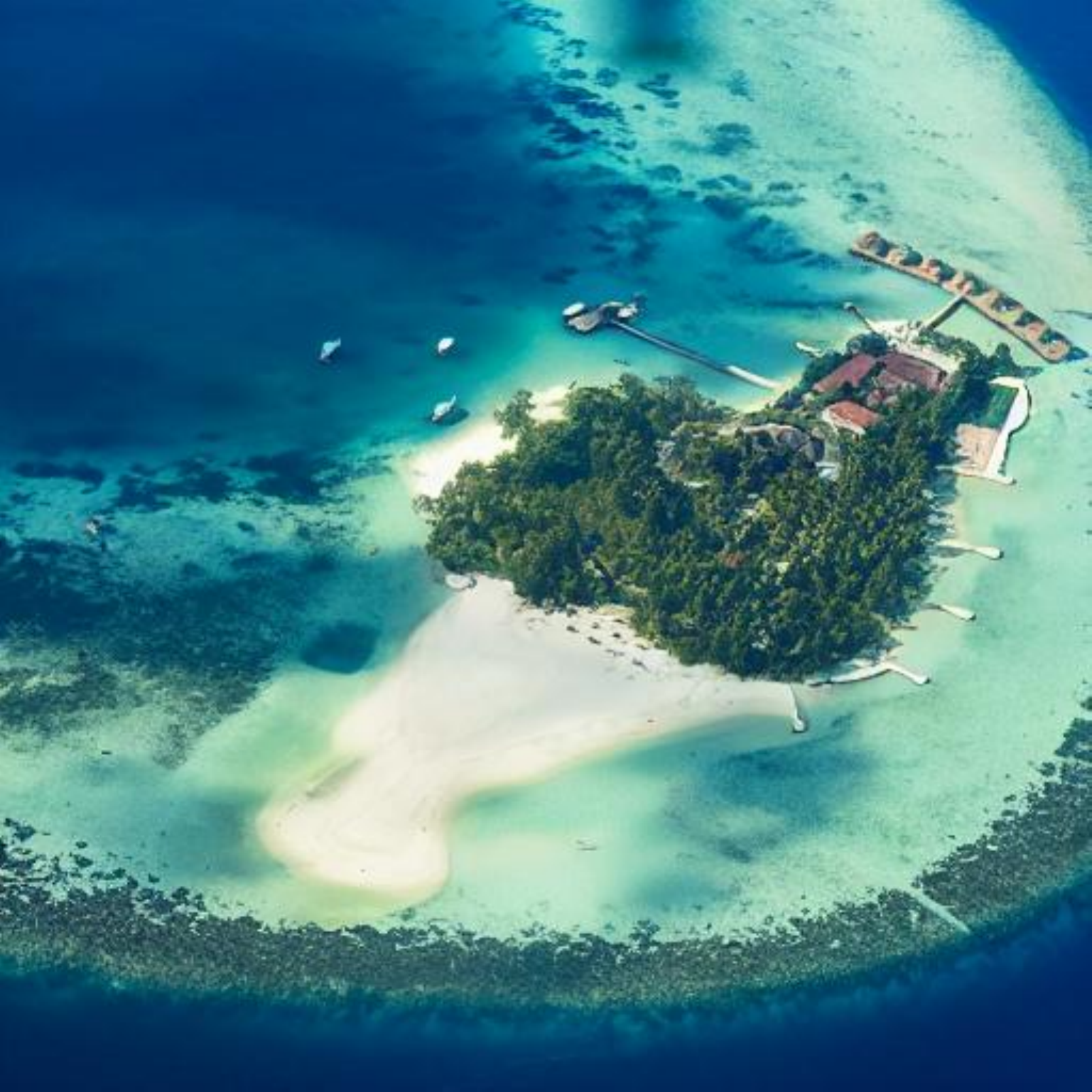}
    \end{subfigure}%

    \begin{subfigure}{.16\linewidth}
        \centering
        \includegraphics[width=\linewidth]{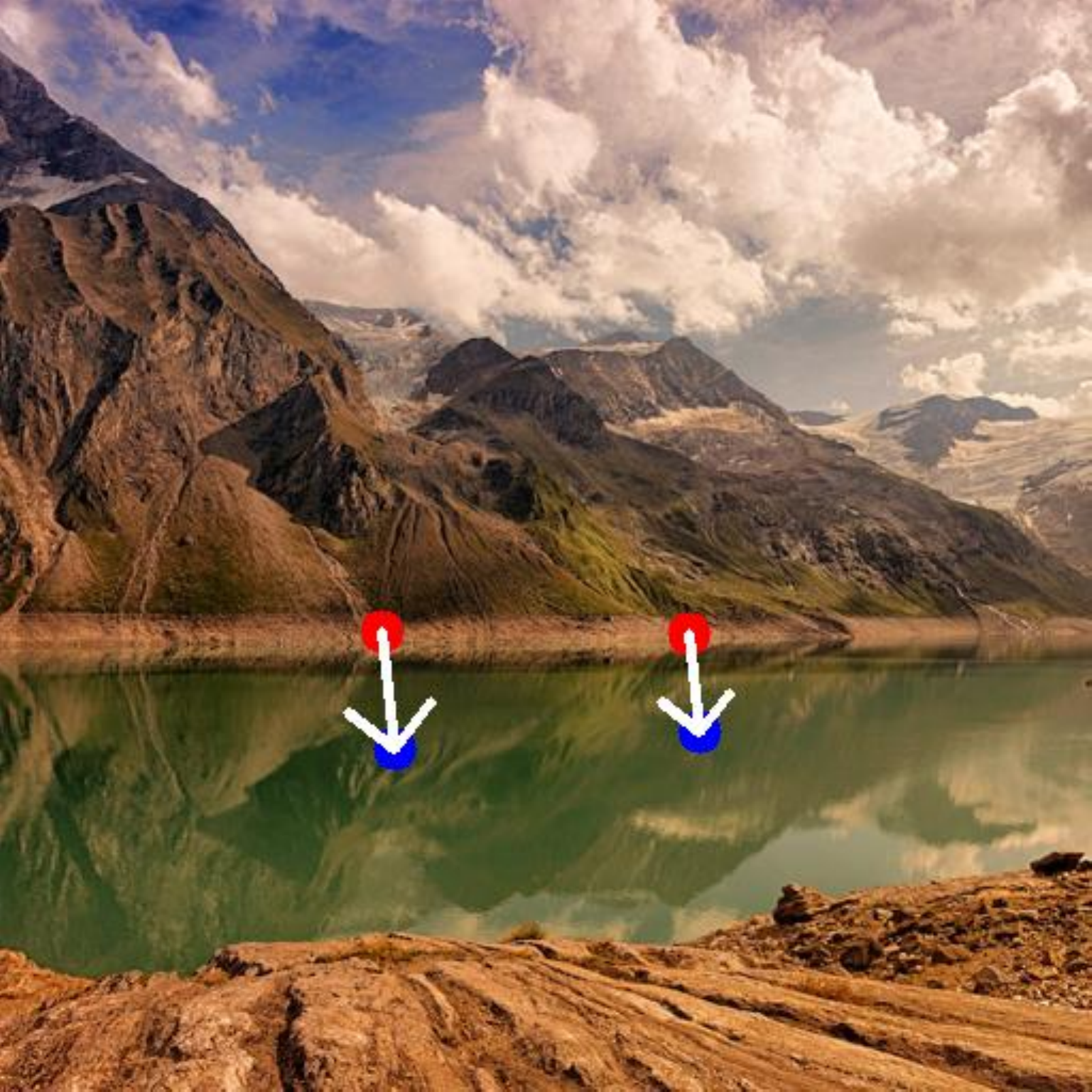}
    \end{subfigure}%
    \begin{subfigure}{.16\linewidth}
        \centering
        \includegraphics[width=\linewidth]{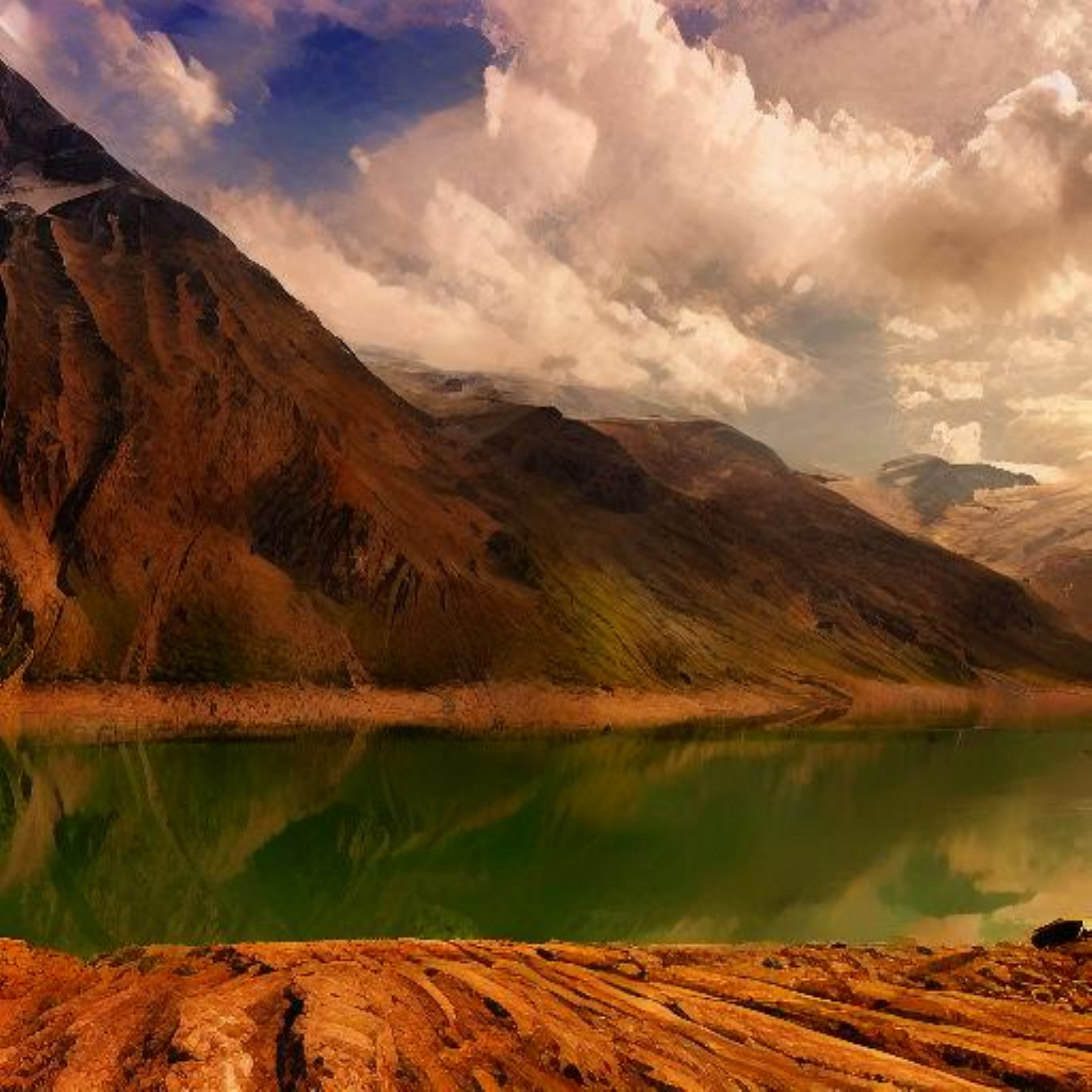}
    \end{subfigure}%
    \begin{subfigure}{.16\linewidth}
        \centering
        \includegraphics[width=\linewidth]{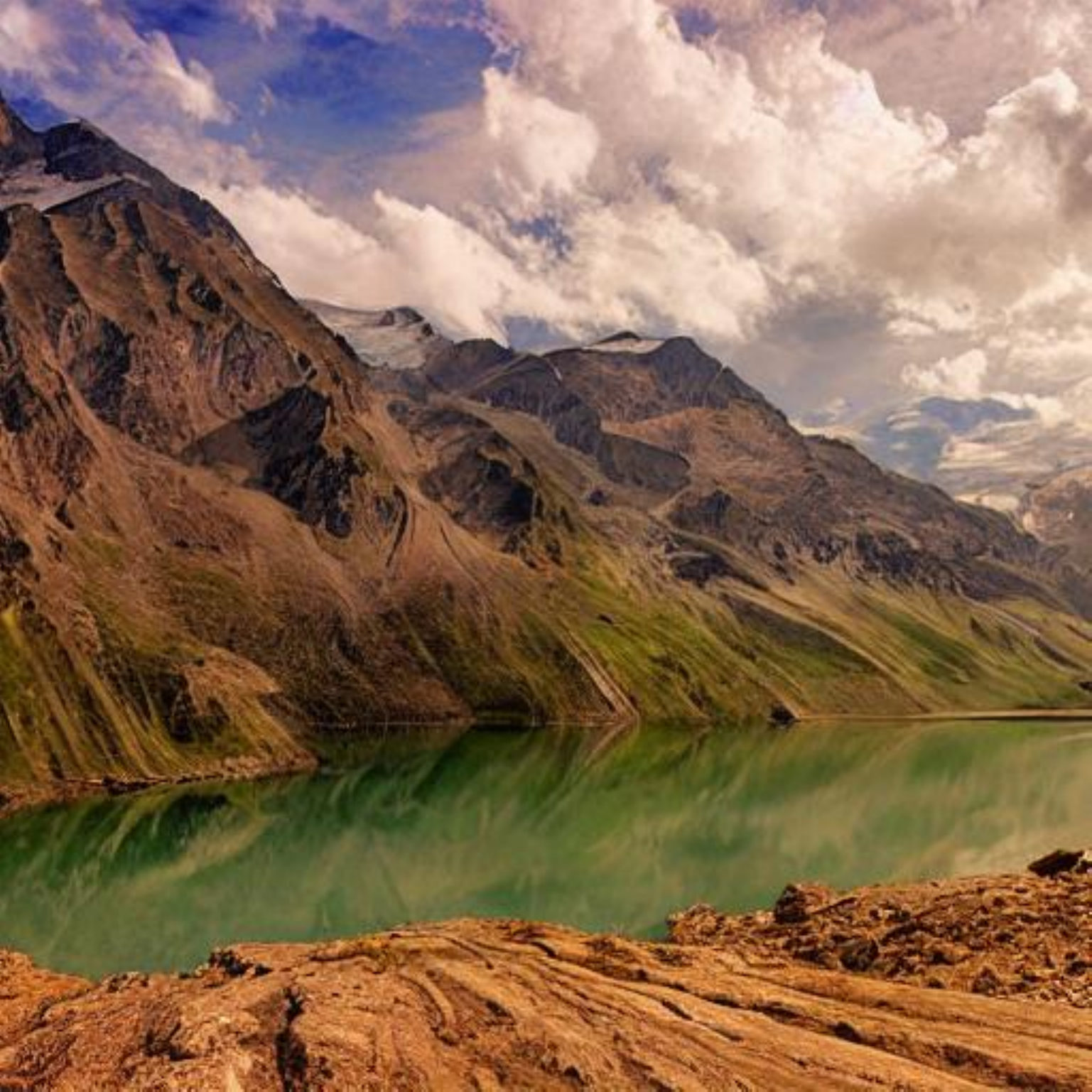}
    \end{subfigure}%
    \begin{subfigure}{.16\linewidth}
        \centering
        \includegraphics[width=\linewidth]{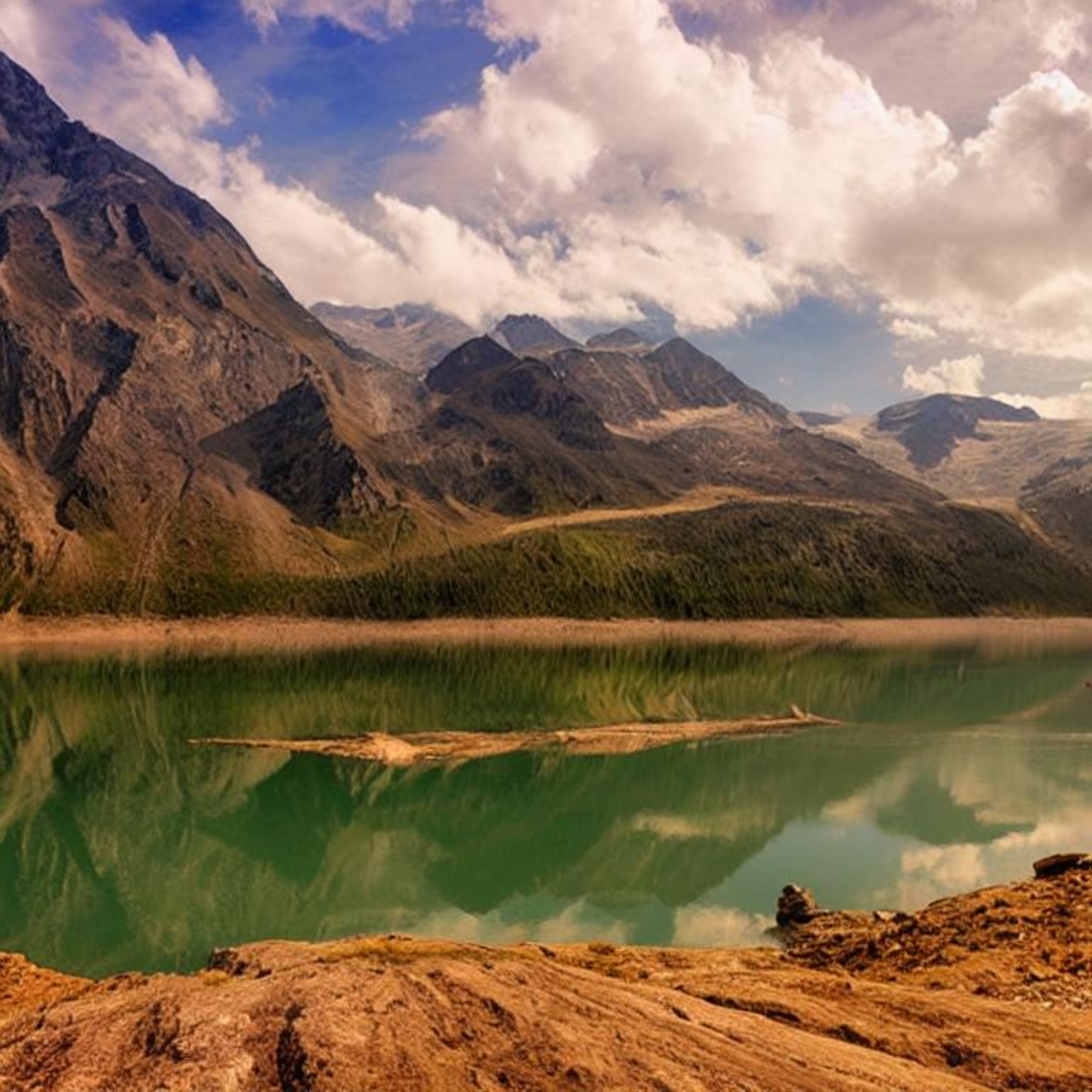}
    \end{subfigure}%
    \hspace{1mm}
    \begin{subfigure}{.16\linewidth}
        \centering
        \includegraphics[width=\linewidth]{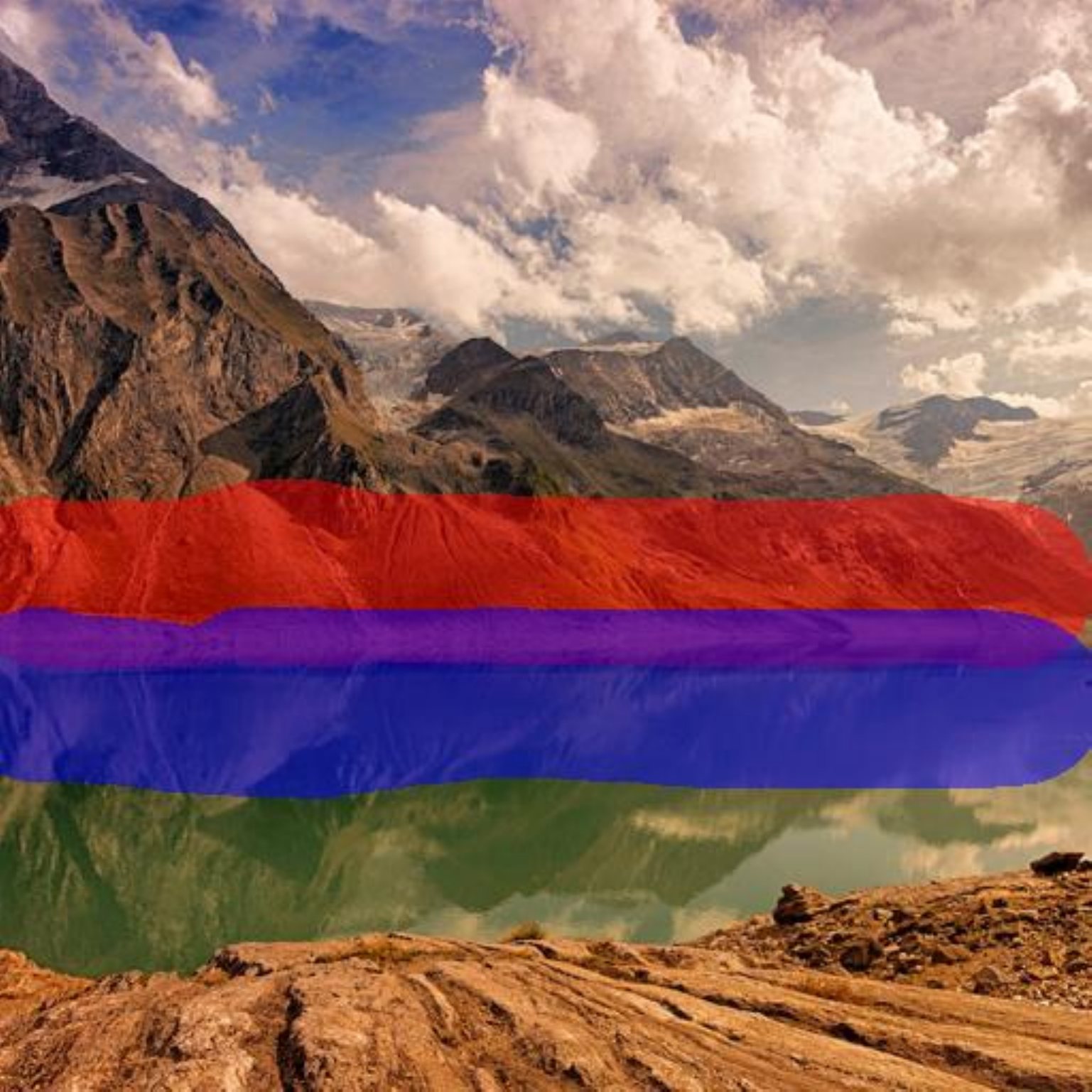}
    \end{subfigure}%
    \begin{subfigure}{.16\linewidth}
        \centering
        \includegraphics[width=\linewidth]{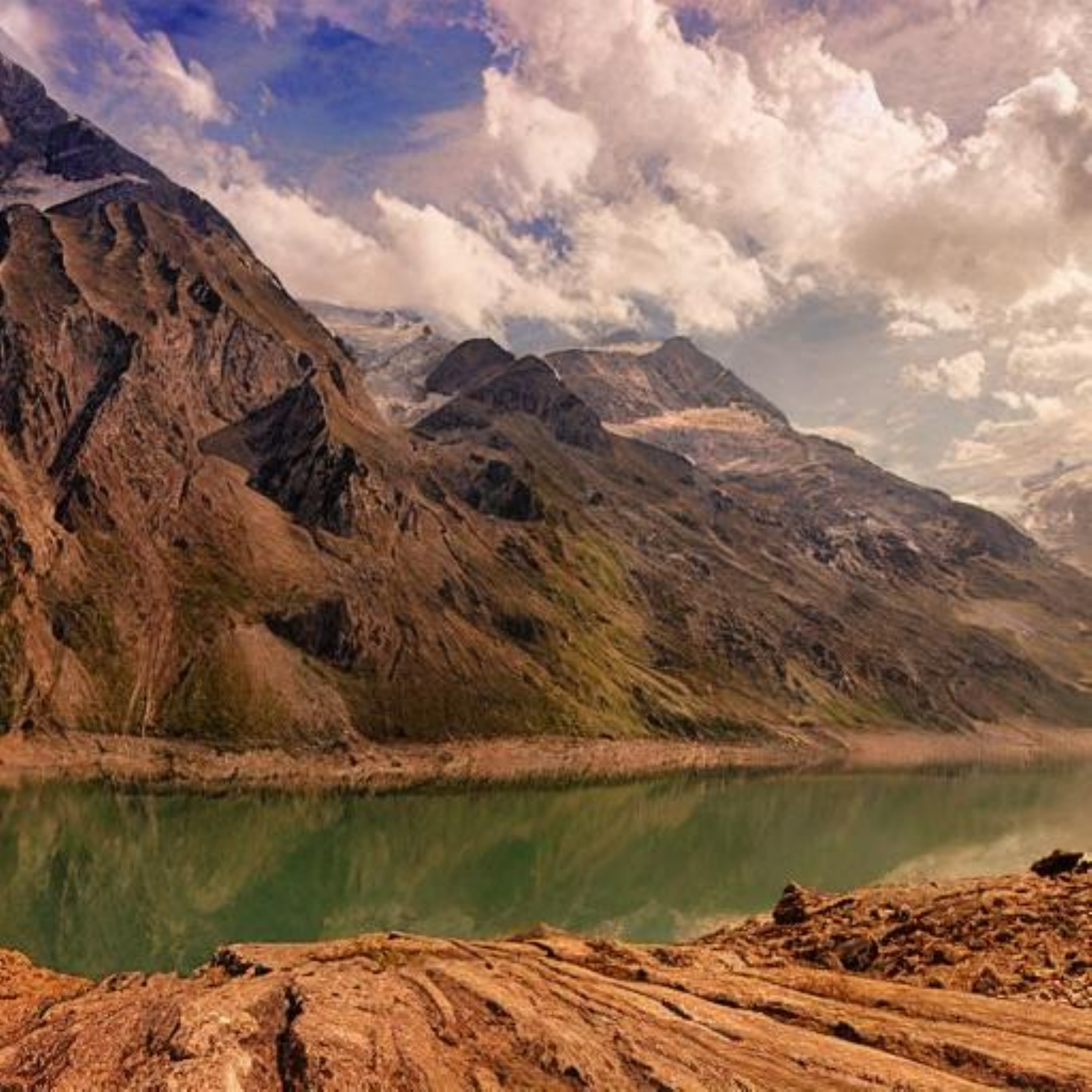}
    \end{subfigure}%

    \caption{More qualitative comparisons with baseline methods.}
    \label{fig:supp_results} 
\end{figure*}

\Cref{fig:supp_results} presents additional examples comparing point-based editing outcomes with our region-based editing results. The region-based edits maintain object identity, resulting in higher-quality modifications that closely adhere to the desired output.

\clearpage

\end{document}